\documentclass{article} 
\usepackage{iclr2025_conference,times}

\usepackage{amsmath,amsfonts,bm}

\def\eqref#1{equation~\ref{#1}}

\def\1{\bm{1}}

\DeclareMathAlphabet{\mathsfit}{\encodingdefault}{\sfdefault}{m}{sl}
\SetMathAlphabet{\mathsfit}{bold}{\encodingdefault}{\sfdefault}{bx}{n}

\usepackage{hyperref}
\usepackage{url}
\usepackage{amsfonts}       
\usepackage{nicefrac}       
\usepackage{microtype}      
\PassOptionsToPackage{prologue,dvipsnames}{xcolor}
\usepackage{colortbl}
\usepackage{graphicx}
\usepackage{amsmath}
\usepackage{multirow}
\usepackage{subcaption} 
\usepackage{wrapfig}
\usepackage{algorithmic}
\usepackage{algorithm}
\usepackage{tcolorbox}
\usepackage{booktabs}
\usepackage{bbm}
\usepackage{enumitem}
\usepackage{amssymb}
\usepackage{makecell}
\usepackage{pifont}
\usepackage{xcolor}

\newtheorem{assumption}{\bf Assumption}  

\definecolor{MyDarkRed}{rgb}{0.8,0.02,0.02}
\definecolor{royalpurple}{rgb}{0.47, 0.32, 0.66}
\colorlet{mylinkcolor}{royalpurple} 
\colorlet{mycitecolor}{royalpurple}
\colorlet{myurlcolor}{MyDarkRed}
\hypersetup{
  citecolor  = mycitecolor,
  linkcolor = mylinkcolor,
  urlcolor = myurlcolor,
  colorlinks = true
}
\usepackage{setspace}
\newcommand{\codesite}{\url{https://sites.google.com/view/rl-cip/}}
\newenvironment{compactitemize}{\begin{itemize}[nosep,leftmargin=*]}{\end{itemize}}

\newcommand{\cmark}{\textcolor{green}{\ding{51}}} 
\newcommand{\xmark}{\textcolor{red}{\ding{55}}}   

\title{Causal Information Prioritization for \\ Efficient Reinforcement Learning}

\author{Hongye Cao$^{1}$ \quad Fan Feng$^{2,3}$ \quad Tianpei Yang$^{1,4}$\thanks{Corresponding to Tianpei Yang (\texttt{tianpei.yang@nju.edu.cn}).} \quad Jing Huo$^{1}$ \quad \textbf{Yang Gao}$^{1,4}$\\
$^{1}$National Key Laboratory for Novel Software Technology, Nanjing University\\
$^{2}$University of California, San Diego \quad $^{3}$MBZUAI\\
$^{4}$School of Intelligence Science and Technology, Nanjing University\\
}

\iclrfinalcopy
\begin{document}

\maketitle

\begin{abstract}
Reinforcement Learning (RL) methods often suffer from sample inefficiency, one of the underlying reasons is that blind exploration strategies may neglect causal relationships among states, actions, and rewards. Although recent causal approaches aim to address this problem, they lack grounded modeling of reward-guided causal understanding of states and actions for goal orientation, thus impairing learning efficiency. To tackle this issue, we propose a novel method named Causal Information Prioritization ($\texttt{\textbf{CIP}}$) that improves sample efficiency by leveraging factored MDPs to infer causal relationships between different dimensions of states and actions with respect to rewards, enabling the prioritization of causal information. Specifically, $\texttt{\textbf{CIP}}$ identifies and leverages causal relationships between states and rewards to execute counterfactual data augmentation to prioritize high-impact state features under the causal understanding of the environments. Moreover, $\texttt{\textbf{CIP}}$ integrates a causality-aware empowerment learning objective, which significantly enhances the agent's execution of reward-guided actions for more efficient exploration in complex environments. 
To fully assess the effectiveness of $\texttt{\textbf{CIP}}$, we conduct extensive experiments across $39$ tasks in $5$ diverse continuous control environments, encompassing both locomotion and manipulation skills learning with pixel-based and sparse reward settings. Experimental results demonstrate that $\texttt{\textbf{CIP}}$ consistently outperforms existing RL methods across a wide range of scenarios. The project page is \codesite. 

\end{abstract}
\vspace{-1em}
\section{Introduction} 
\vspace{-1em}

Reinforcement Learning (RL) has emerged as a powerful paradigm for training intelligent decision-making agents to learn optimal behaviors by interacting with their environments, receiving reward feedback, and iteratively optimizing their decision-making policies~\citep{haarnoja2018soft,ze2024h,sutton2018reinforcement,silver2017mastering,cao2023enhancing}. 
Despite its notable successes, most RL approaches are faced with the sample-inefficiency problem, which means they typically necessitate an enormous number of interactions with the environment to learn policies, which can be impractical or costly in real-world scenarios~\citep{savva2019habitat, kroemer2021review}. 
Inefficient policy learning often results from blind exploration strategies that neglect causal relationships, leading to spurious correlations and suboptimal solutions with high exploration costs~\citep{zeng2023survey,liu2024learning}. 

Causal reasoning captures essential information by analyzing causal relationships between different factors, filtering out irrelevant information, and avoiding interference from spurious correlations~\citep{wang2022causal,pitis2022mocoda, zhang2024interpretable, huang2022action}. 
These approaches build internal causal structural models, enabling agents to strategically focus their exploration on the most pertinent aspects of the environment. They significantly reduce the number of samples required and demonstrate remarkable performance in single-task learning, generalization, and counterfactual reasoning~\citep{richens2024robust,urpicausal,deng2023causal,huangadarl,feng2023learning}. 
However, most of these works overlook the reward-relevant causal relationships among different factors, or only partially consider the causal connections between states, actions, and rewards~\citep{liu2024learning,ji2024ace}, thus hindering efficient exploration. 

\begin{figure}
    \centering
    \includegraphics[width=1\linewidth]{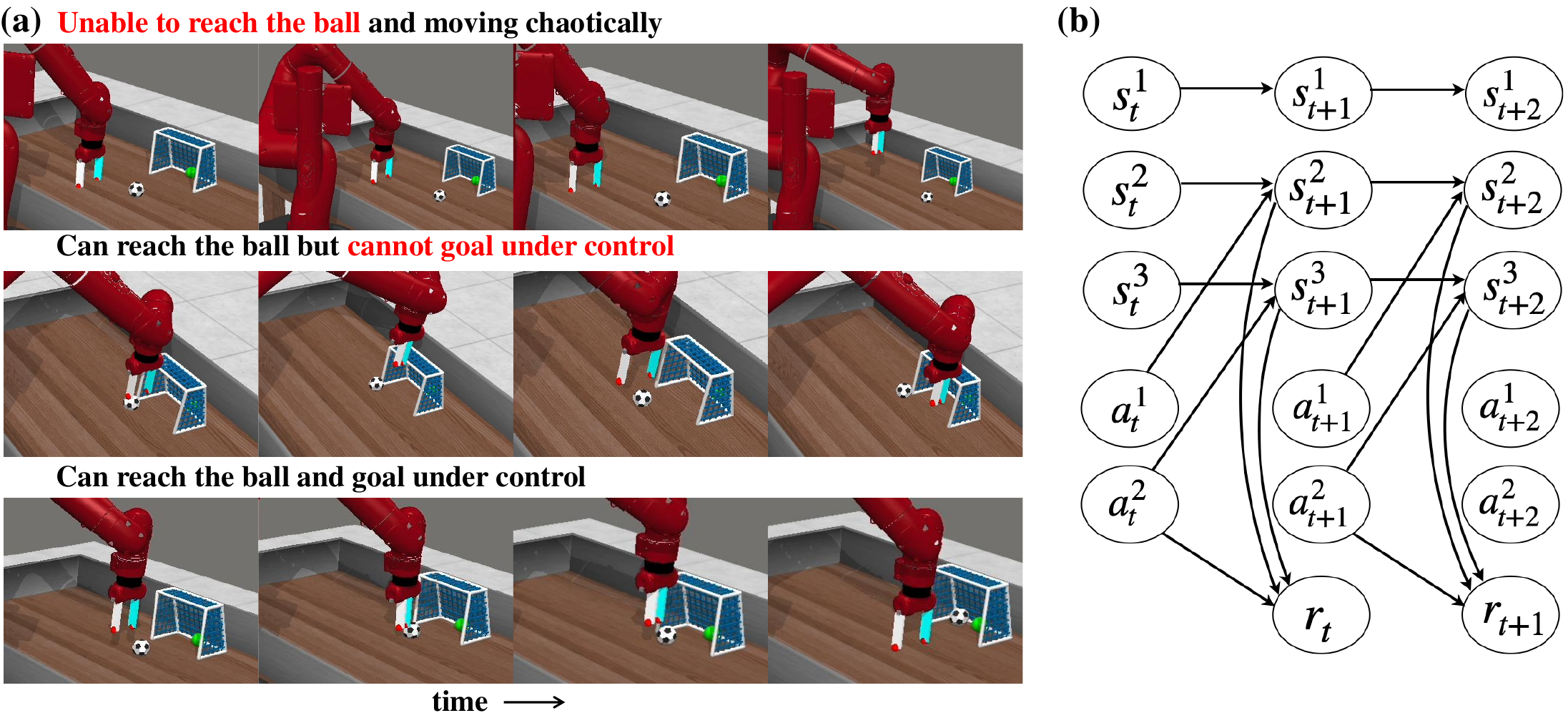}
    \caption{(a). An example of a robot manipulation soccer task with three trajectories, where the objective is to move the ball into the goal. (b). Underlying causal structure of this example in a factored MDP. Different nodes represent different dimensional states and actions. }
    \label{fig:example}
    \vspace{-8mm}
\end{figure}

In this work, we aim to identify and exploit task-specific causal relationships between states, actions, and rewards, enabling agents to discern relevant states and select actions that maximize rewards, ultimately facilitating precise and goal-oriented behaviors. Here we provide a motivating example in Figure~\ref{fig:example}, showing three trajectories for executing a manipulation soccer task, along with the underlying causal structure in a factored Markov Decision Process (MDP)~\citep{kearns1999efficient}. In the first trajectory (row 1), when the agent fails to distinguish states with more intricate causal relationships of the task, the robotic arm exhibits chaotic moving and receives no rewards. The second trajectory (row 2) shows that even without chaotic movements, uncontrollable actions unrelated to the reward lead to an inability to guide the ball towards the goal. Only by filtering out irrelevant state features and executing more controllable actions can we guarantee that the ball is kicked into the goal like row 3. 
Quantifying the contribution of different factors to the reward can effectively help analyzing important causal relationships. 

To address the limitation of sample-inefficiency and leverage the potential of causal reasoning, we propose a novel approach named Causal Information Prioritization ($\texttt{\textbf{CIP}}$) for efficient RL, improving learning efficiency from the perspective of rewards. 
Building upon the factored MDPs, $\texttt{\textbf{CIP}}$ infers causal relationships between states, actions, and rewards across different dimensions, respectively. $\texttt{\textbf{CIP}}$ employs counterfactual data augmentation based on the causality between states and rewards to generate transitions, prioritizing critical state transitions. Furthermore, $\texttt{\textbf{CIP}}$ leverages the causality between actions and rewards to reweight actions, while utilizing empowerment to maximize mutual information between causally informed actions and future states, thereby enabling better control. 

Specifically, $\texttt{\textbf{CIP}}$ leverages collected data to construct a reward-guided structural model that explicitly reasons about state-reward causal influences, enabling the swapping of causally independent state features across observed trajectories to generate synthetic transitions without additional environment interactions. By swapping independent state features across different transitions (i.e., irrelevant state dimensions of chaotic movements in the soccer task), $\texttt{\textbf{CIP}}$ accentuates causally dependent state information (i.e., relevant states to reach the ball), facilitating focused learning of critical state transitions. 
Subsequently, $\texttt{\textbf{CIP}}$ constructs another structural model that incorporates actions and rewards to reweight actions of dimensions.
To enhance the exploration efficiency, $\texttt{\textbf{CIP}}$ integrates a causality-aware empowerment term, quantifying the agent's capacity to exert controlled influence over its environment through the mutual information. 
This empowerment term, combined with causally weighted actions, is integrated into the learning objective, prioritizing actions with high causal influence. The synthesis of causal reasoning and action empowerment enables agents to focus on behaviors that are causally relevant to the task, leading to more efficient and effective policy learning. 
The main contributions of this work can be summarized as follows. 
\begin{compactitemize}
    \item 
    To address limitations of blind exploration and sample-inefficiency, we introduce $\texttt{\textbf{CIP}}$, a novel efficient RL framework that prioritizes causal information through the lens of reward. $\texttt{\textbf{CIP}}$ bridges the gap between causal reasoning and empowerment to facilitate efficient exploration. 
    \item  $\texttt{\textbf{CIP}}$ constructs reward-guided structural models to uncover causal relationships between states, actions, and rewards across dimensions. By leveraging state-reward causality, it performs counterfactual data augmentation, eliminating the need for additional environment interactions, and enabling learning on critical state transitions. Exploiting action-reward causality, it reweights actions to enhance exploration efficiency through empowerment. By prioritizing causal information, $\texttt{\textbf{CIP}}$ enables agents to focus on behaviors that have causally significant effects on their tasks. 
    \item  To validate the effectiveness of $\texttt{\textbf{CIP}}$, we conduct extensive experiments in $39$ tasks across $5$ diverse continuous control environments, including manipulation and locomotion. These comprehensive evaluations demonstrate the effectiveness of $\texttt{\textbf{CIP}}$ in pixel-based and sparse reward settings, underscoring its versatility and reliability. 
\end{compactitemize}
\vspace{-3mm}
\section{Related Work}
\vspace{-1mm}
\subsection{Causal RL} 
\vspace{-1mm}
The application of causal reasoning in RL has shown significant potential to improve sample efficiency and generalization by effectively excluding irrelevant environmental factors through causal analysis~\citep{huangadarl,feng2023learning,mutti2023provably, sun2024acamda, sun2022toward}. Wang~\citep{wang2021task} introduces a novel regularization-based method for causal dynamics learning, which explicitly identifies causal dependencies by regulating the number of variables used to predict each state variable. 
CDL~\citep{wang2022causal} takes an innovative approach by using conditional mutual information to compute causal relationships between different dimensions of states and actions. 
IFactor~\citep{liu2024learning} is a general framework to model four distinct categories of latent state variables, capturing various aspects of information. ACE~\citep{ji2024ace}, an off-policy actor-critic method, integrates causality-aware entropy regularization. Table~\ref{tab:causal_rl} provides a categorization of various causal RL methods, highlighting their focus on different reward-guided causal relationships. 
Existing approaches do not fully account for the causal relationships between both states and actions with rewards. Our goal is to explore these causal relationships from a reward-guided perspective to enhance sample efficiency across a broader range of tasks.
\vspace{-3mm}

\subsection{Empowerment in RL}
\vspace{-2mm}
Empowerment, an information theory-based concept of intrinsic motivation, has emerged as a powerful paradigm for enhancing an agent's environmental controllability \citep{mohamed2015variational,klyubin2005empowerment,cao2024towards}. This framework conceptualizes actions and future states as information transmission channels, offering a novel perspective on agent-environment interactions. 
In RL, empowerment has been applied to uncover more controllable associations between states and actions, as well as to develop robust skill~\citep{salge2014empowerment,bharadhwaj2022information,choi2021variational, eysenbach2018diversity,leibfried2019unified,seitzer2021causal}. Empowerment, expressed as maximizing mutual information $\max_{\pi} I$, serves as a learning objective in various RL frameworks, providing intrinsic motivation for exploration and potentially yielding more efficient and generalizable policies. Our approach extends empowerment in RL by examining the influence of state, actions, and rewards through a causal lens, integrating causal understanding with empowerment to enhance exploration strategy and learning efficiency.
\vspace{-3mm}

\vspace{-1mm}
\subsection{Object-centric RL and Object-Oriented RL}
\vspace{-2mm}
Recent advances in object-centric representation learning focus on acquiring and leveraging structured, object-wise representations from high-dimensional observations. Foundational works include Slot Attention~\citep{locatello2020object} and AIR~\citep{eslami2016attend, kosiorek2018sequential}, establishing basis for this field. Subsequent follow-ups have worked on these concepts by employing state-of-the-art architectures, including DINO-based approaches~\cite{zadaianchuk2023objectcentric}, transformer-based models~\citep{wu2022slotformer}, diffusion models~\citep{jiang2023object}, and state-space models~\citep{jiang2024slot}. Notably, learning object-centric representations can enable compositional generalization across various domains, such as video and scene generation~\citep{wu2023slotdiffusion, wu2024neural}. Moreover, several theoretical studies have explored the mechanisms underlying compositional generalization and the causal identifiability~\citep{kori2024identifiable, brady2023provably, lachapelle2024additive}.

Object-centric representations have been effectively employed in world models to capture multi-object dynamics, as demonstrated by works~\citep{jiang2019scalor,lin2020improving,kossen2019structured}. Building on these object-centric world models, various studies use them in RL by better modeling complex object-centric structures in partially observable MDPs~\citep{kossen2019structured, mambelli2022compositional, feng2023learning, choi2024unsupervised}, identifying critical objects ~\citep{zadaianchuk2022self, park2021object}, and learning object-centric policies~\citep{zadaianchuk2020self, yuan2022sornet} and applications in robotic manipulation tasks~\citep{li2020towards, mitash2024scaling, haramati2024entity, li2024manipllm}, as well as in learning intrinsic or curiosity-driven policies based on objects and their interactions~\citep{watters2019cobra, hu2023elden, wang2024skild}. 

Another research direction explores object-oriented MDPs, with the homomorphic object-oriented world model being a notable example that leverages MDP homomorphism to model object dynamics and enable efficient planning through symmetric equivalence in MDPs~\citep{diuk2008object, scholz2014physics, wandzel2019multi, van2020plannable, rezaei2022continuous, zhao2022toward}, provides a powerful foundation for learning object-oriented MDPs and facilitates efficient planning~\citep{wolfe2006defining}.   Our work, which focuses on uncovering general causal relationships among components in MDPs and empowerment optimization, is orthogonal to object-centric RL. However, object-centric RL could provide useful abstract object-based variables that could be useful for causal structure learning in complex environments~\footnote{We provide a detailed discussion in Appendix~\ref{ocrl}.}.

\vspace{-2mm}
\section{Preliminaries}
\vspace{-2mm}
\subsection{Markov Decision Process}
\vspace{-1mm}
In RL, the agent-environment interaction is formalized as an MDP. The standard MDP is defined by the tuple $ \mathcal{M} = \langle \mathcal{S}, \mathcal{A}, \mathcal{P}, \mu_0, r, \gamma \rangle $, where $\mathcal{S}$ denotes the state space, $\mathcal{A}$ represents the action space, $\mathcal{P}(s' | s, a)$ is the transition dynamics, $r(s, a)$ is the reward function, and $\mu_0$ is the distribution of the initial state $s_0$. The discount factor $\gamma \in [0, 1)$ is also included. The objective of RL is to learn a policy $\pi: \mathcal{S} \times \mathcal{A} \to [0, 1]$ that maximizes the expected discounted cumulative reward ${\eta _\mathcal{M}}(\pi) := \mathbb{E}_{s_0 \sim \mu_0, s_t \sim \mathcal{P}, a_t \sim \pi} \left[\sum\nolimits_{t = 0}^\infty {\gamma^t}r(s_t, a_t)\right]$. 
\vspace{-1mm}
\subsection{Structural Causal Model}
\vspace{-1mm}
A Structural Causal Model (SCM)~\citep{pearl2009causality} is defined by a distribution over random variables, defined as $\mathcal{V}=\{s_t^1, \cdots, s_t^d, a_t^1, \cdots, a_t^n, r_t, s_{t+1}^1, \cdots, s_{t+1}^d \}$ and a Directed Acyclic Graph (DAG) $\mathcal{G}=(\mathcal{V}, \mathcal{E})$ with a conditional distribution $\mathcal{P}(v_i|\mathrm{PA}(v_i))$ for node $v_i \in \mathcal{V}$. Then the distribution can be specified as: 
\begin{equation}
    p(v_1, \dots, v_{|\mathcal{V}|})= \prod_{i=1}^{|\mathcal{V}|}p(v_i|\mathrm{PA}(v_i) ) ,
\end{equation}
where $\mathrm{PA}(v_i)$ is the set of parents of the node $v_i$ in the graph $\mathcal{G}$. 
\paragraph{Causal Structures in MDP}
We use a factored MDP~\citep{kearns1999efficient,guestrin2003efficient, guestrin2001multiagent} to model the MDP and the underlying causal structures between states, actions, and rewards. In the factored MDP, nodes represent system variables (rewards and different dimensions of the states and actions), while the edges denote their relationships within the MDP. We employ causal discovery methods to learn the structures of $\mathcal{G}$. 

We can identify the graph structure in $\mathcal{G}$, which can be represented as the adjacency matrix $M$. To integrate such relationships in MDP, we explicitly encode the causal mask over variables into the reward function. Hence, the reward function in MDP with the causal structure is defined as follows: 
\begin{equation}
r_t = R(M^{s \to r} \odot s_t, M^{a \to r} \odot a_t, \epsilon_{r,t})
\label{eq:gen}
\end{equation}
where \( \odot \) denotes the element-wise product. $ M^{s \to r} \in \mathbb{R}^{|s|\times 1}$ and $ M^{a \to r} \in \mathbb{R}^{|a|\times 1}$ are the adjacency matrices indicating the influence of current states and actions on the reward, respectively, and \( \epsilon_{r,t} \) represents i.i.d. Gaussian noise. Under the Markov condition and faithfulness assumption~\citep{pearl2009causality,spirtes2001causation}, the structural vectors are identifiable. The detailed assumptions and propositions can be found in Appendix~\ref{sec:app_ass}. 
In this work, our objective is to discover and leverage these two causal matrices to prioritize causal information for efficient RL. 
\vspace{-2mm}
\subsection{Empowerment in RL}
\vspace{-2mm}
Empowerment quantifies an agent's capacity to influence its environment and perceive the consequences of its actions~\citep{klyubin2005empowerment,bharadhwaj2022information,jung2011empowerment}. In our framework, the empowerment is defined as the mutual information between the agent state ${s}_{t+1}$ and action ${a}_t$, conditioned on the present state $s_t$ and causal mask $M$, as shown follows: 
\begin{equation}
    \mathcal{E} := \max_{\pi} \mathcal{I}(a_{t}; s_{t+1} \mid s_t, M),
\end{equation}
where $\mathcal{E}$ denotes the channel capacity from actions to states. Unlike \citep{cao2024towards}, which focuses on action-to-state empowerment effects, we leverage causal understanding and more accurate entropy calculation to analyze state-to-action influences, facilitating the development of more controllable behavioral policies. 

\vspace{-2mm}
\section{Causal Information Prioritization}
\label{sec:app}
\begin{figure}[t]
    \centering
    \includegraphics[width=1\linewidth]{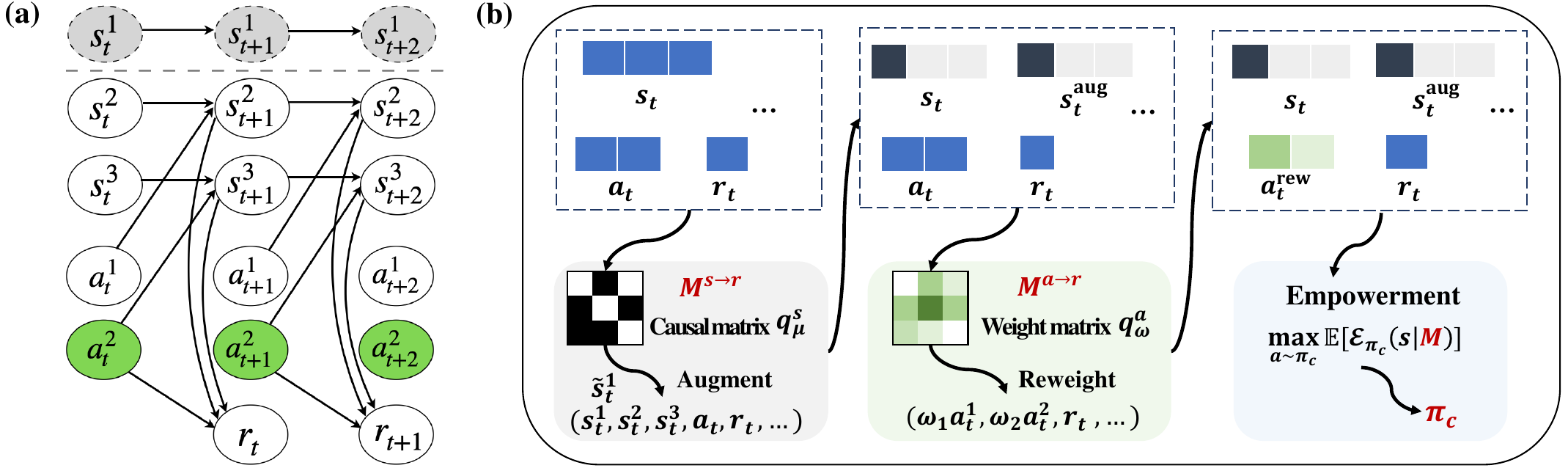}
    \caption{(a) Underlying causal 
 structure of $\texttt{\textbf{CIP}}$. (b) The whole learning process of $\texttt{\textbf{CIP}}$ includes counterfactual data augmentation, causal action reweight and causal action empowerment.}
    \label{fig:framework}
    \vspace{-3mm}
\end{figure}

\vspace{-2mm}
In this section, we introduce the proposed framework $\texttt{\textbf{CIP}}$, which implements causal information prioritization based on the causal relationships between states, actions, and rewards (as shown in Figure~\ref{fig:framework}). First, we train a structural model based on the causal discovery method, DirectLiNGAM~\citep{shimizu2011directlingam} using collected trajectories to obtain a causal matrix $M^{s \to r}$. Utilizing this matrix, $\texttt{\textbf{CIP}}$ executes the swapping of causally independent state features, generating synthetic transitions (Section~\ref{sec:aug}). This process of swapping independent state information accentuates causally dependent state information, enabling focused learning on critical state transitions. 
Subsequently, $\texttt{\textbf{CIP}}$ constructs another structural model to get a weight matrix $M^{a \to r}$ that incorporates actions and rewards to reweight actions (Section~\ref{sec:emp}). Furthermore, $\texttt{\textbf{CIP}}$ integrates a causality-aware empowerment term $\mathcal{E}_{\pi_c}(s)$ combined with causally weighted actions into the learning objective to promote efficient exploration. This integration encourages the agent's policy $\pi_c$ to prioritize actions with high causal influence, thereby enhancing its goal-achievement capabilities.

\vspace{-1mm}
\subsection{Counterfactual Data Augmentation}
\label{sec:aug}
\vspace{-1mm}
To discover the causal relationships between states and rewards, we initially collect trajectories to train a structural model by the DirectLiNGAM method, denoted as $q_\mu^s$, to obtain the causal matrix $M^{s \to r}$. Subsequently, we infer the local factorization, which is utilized to generate counterfactual transitions. 
For each state $s$ in the trajectories, we compute the uncontrollable set, defined as the set of variables in $s$ for which the agent has no causal influence on rewards: 
\begin{equation}
\mathcal{U}_s = \{ s^i \mid M^{s \to r} \cdot(s^i_t, r_t) < \theta; i \in [1,N] \},
\label{cda}
\end{equation}
where $\theta$ is a fixed threshold and $N$ is the dimension of the state space. The set $\mathcal{U}_s$ encompasses all dimensional state variables for which the causal relationship $s^i_t \to r_t$ does not exist in the causal matrix of states and rewards.
Utilizing the learned causal matrix $M^{s \to r}$, we partition all state variables in the factored MDP into controllable and uncontrollable sets. These uncontrollable sets are then leveraged for counterfactual data augmentation, thereby prioritizing the causally-informed state information to improve learning efficiency. 

To generate counterfactual samples, we perform a swap of variables that fall under the uncontrollable category (i.e., in set $\mathcal{U}_s$) sampled from the collect trajectories. Specifically, given two transitions $(s_t, a_t, s_{t+1}, r_t)$ and $(\hat{s}_j, \hat{a}_j, \hat{s}_{j+1}, \hat{r_j})$ sampled from trajectories, which share at least one uncontrollable sub-graph structure (i.e., $\mathcal{U}_s \cap \mathcal{U}_{\hat{s}} \neq \emptyset$), we construct a counterfactual transition $(\tilde{s_t}, \tilde{a_t}, \tilde{s}_{t+1}, \tilde{r_t})$ by swapping the irrelevant state variables $(s^i_t, s^i_{t+1})$ with $(\hat{s}^i_j, \hat{s}^{i}_{j+1})$ for each $i \in \mathcal{U}_s \cap \mathcal{U}_{\hat{s}}$. 
The augmented transitions will be added to the training data for causal reasoning during subsequent action empowerment, thus eliminating the need for additional environment interactions to prioritize causal information. Furthermore, we also consider directly using controllable state sets combined with causal action empowerment to replace counterfactual data augmentation for policy learning. The comparative experimental results validating this approach are presented in Appendix~\ref{sec:appendix_cda_replace}. 

\vspace{-1mm}
\subsection{Causal Action Prioritization Through Empowerment}
\label{sec:emp}
\vspace{-1mm}

\paragraph{Causal action reweight} 
Having analyzed the causal relationships between states and rewards to achieve efficient data augmentation, in this section, we further discover the causal relationships between actions and rewards to prioritize causally-informed decision-making behaviors. 
$\texttt{\textbf{CIP}}$ constructs a reward-guided structural model, incorporating states (including augmented states), actions, and rewards. This model forms the foundation for action prioritization in policy learning, enabling action reweighting based on causality. Leveraging this structural model to delineate relationships between policy decisions and rewards, we evaluate the causal impact of different actions on reward outcomes. In this way, the agent focuses on pivotal actions with demonstrable causal links to desired reward outcomes, potentially accelerating learning and optimizing performance in complex environments. 

Specifically, in $\texttt{\textbf{CIP}}$, we employ DirectLiNGAM method to train a causal structural model $q^a_{\omega}$, which yields a weight matrix $M^{a \to r}$, delineating the relationships between actions and rewards, conditioned on the states. 
For a given set of actions $(a^1_t, a^2_t, a^3_t, \dots)$, we utilize the weight matrix $M^{a \to r}$ to reweight them as $(\omega_1 a^1_t, \omega_2 a^2_t, \omega_3 a^3_t, \dots)$, where $\omega$ represents the causal weights derived from the matrix $M^{a \to r}$. 
By leveraging this causal structure, we can prioritize the most pivotal actions, potentially leading to more efficient policy exploration and targeted policy improvements. 

\paragraph{Causal action empowerment} 
Based on the learned causal structure, we propose the causal action empowerment to incorporate the reweighted actions into the learning objective for efficient exploration in a controllable manner. 
To this end, we design a causality-aware empowerment term $\mathcal{E}_{\pi_c}(s)$ for policy optimization. We maximize the empowerment gain of the policy $\pi_c$, where $\pi_c$ incorporates the learned causal structure. This approach allows us to quantify and maximize the empowerment that can be achieved by explicitly considering causal relationships, thereby bridging the gap between causal reasoning and empowerment. 

We denote the empowerment of the causal policy as $\mathcal{E}_{\pi_c}(s) = \max_a  \mathcal{I}\left(a_t; s_{t+1} \mid s_t; M\right)$. 
We then formulate the following objective empowerment function: 
\begin{equation}
\begin{aligned}
   \mathcal{E}_{\pi_c}(s) & = \max_a  \mathcal{I}\left(a_t; s_{t+1} \mid s_t; M\right) \\
   & = \max_{a_t \sim \pi_c(\cdot|s)} \mathcal{H}(\pi_c(a_t|s_t)) - \mathcal{H}(\pi_c(a_t|s_t; s_{t+1})), 
    \end{aligned}
    \label{eq:emp_comp}
\end{equation}
where $\pi_c$ is the policy under the causal weighted matrix $M^{a \to r}$. 
The first entropy term $\mathcal{H}(\pi_c(a_t|s_t))$ promotes action diversity within the constraints of the causal structure. It encourages the agent to explore a wide range of actions that are causally informed, while the second entropy term $- \mathcal{H}(\pi_c(a_t|s_t; s_{t+1}))$ enhances the action predictability in state transitions. It encourages the selection of actions that lead to predictable outcomes, given the current and subsequent states, thereby promoting controlled and goal-oriented behaviors. 
We train an inverse dynamics model to represent the policy $\pi_c(\cdot|s_t;s_{t+1})$. 
The detailed derivation proceeds as follows: 
\begin{equation}
    \mathcal{H}(\pi_c(\cdot|s_t)) = -\mathbb{E}_{a_t \in  \mathcal{A}} \left[
    \sum_{i=1}^{d_\mathcal{A}} {M^{a^i \to r}} \odot \pi_c(a_t^i|s_t)\log \pi(a_t^i|s_t) 
    \right], 
\end{equation}
    and 
\begin{equation}
    \mathcal{H}(\pi_c(\cdot|s_t;s_{t+1})) = -\mathbb{E}_{a_t \in  \mathcal{A}} \left[
    \sum_{i=1}^{d_\mathcal{A}} {M^{a^i \to r}} \odot \pi_c(a_t^i|s_t;s_{t+1})\log \pi(a_t^i|s_t;s_{t+1}) 
    \right],
\end{equation}
where $d_\mathcal{A}$ is the dimension of the action space. Hence, the learning objective of the causal action empowerment can be defined as follows: 
\begin{equation}
    \begin{aligned}
   \mathcal{E}_{\pi_c}(s)  &  = \max_{a_t \sim \pi_c(\cdot|s)} \mathcal{H}(\pi_c(a_t|s_t)) - \mathcal{H}(\pi_c(a_t|s_t; s_{t+1}))
    \\ & = \max_{a_t \sim \pi_c(\cdot|s)} \mathbb{E}_{\pi_c(a_t|s_t) p_{\pi_c}( a_t|s_t,s_{t+1})} \left[\log \mathcal{P}_{\phi_c}(a_t \mid s_t, s_{t+1}; M) - \log \mathcal{P}_{\pi_c}(a_{t}|s_t;M) \right],
    \end{aligned}
    \label{eq:emp_comp}
\end{equation}
where $ \mathcal{P}_{\pi_c}(a_t|s_t; M)$ is the action distribution given current state of policy $\pi_c$ with the causal structure, which can be denoted as $\pi_c(a_t|s_t)$. $\mathcal{P}_{\phi_c}(a_t|s_{t+1},s_t; M)$ represents an inverse dynamics model trained on the collected transitions of state variables. 
Hence, we update the target policy $\pi_c$ by maximizing the empowerment objective function derived in Eq.~\ref{eq:emp_comp}.

Adhering to the maximum entropy paradigm~\citep{haarnoja2018soft},
we calculate $\mathcal{E}_{\pi_c}(s)$ for maximization instead of standard entropy, thus prioritizing exploration of pivotal actions that are more likely to have significant causal effects on the reward. This targeted exploration strategy has the potential to accelerate learning by focusing on the most influential actions in current controllable states. 
Based on the causality-aware empowerment, the Q-value for policy $\pi_c$ could be computed iteratively by applying a modified Bellman operator $\mathcal{T}^{\pi}_c$ with $\mathcal{E}_{\pi_c}(s)$ term as stated below: 
\begin{equation}
\fontsize{8.5}{1}
\begin{aligned}
    \mathcal{T}^{\pi}_c Q(s_t,a_t) & =r(s_t,a_t)+\gamma \mathbb{E}_{s_{t+1}\sim \mathcal{P}} \left[
    \mathbb{E}_{a_t \sim \pi_c}\left[Q(s_{t+1},a_{t+1})
    +\alpha \mathcal{E}_{\phi_c}(s)
    \right]
    \right] \\
    & =r(s_t,a_t)+\gamma \mathbb{E}_{s_{t+1}\sim \mathcal{P}} \left[
    \mathbb{E}_{a_t \sim \pi_c}\left[Q(s_{t+1},a_{t+1})
    +\alpha (\mathcal{H}(\pi_c(a_t|s_t)) - \mathcal{H}(\pi_c(a_t|s_t; s_{t+1})))
    \right]
    \right]. \\
\end{aligned}
\label{eq:alpha}
\end{equation}Hence, we integrate the causality-aware empowerment term into the policy optimization objective function, $\hat{\eta}_\mathcal{M}(\pi_c)= 
\mathbb{E}_{s_0 \sim \mu_0, s_t \sim \mathcal{P}, a_t \sim \pi_c} \left[\sum\nolimits_{t = 0}^\infty {\gamma^t} (r(s_t, a_t) +\alpha \mathcal{E}_{\pi_c}(s))\right]
$. 

In summary, $\texttt{\textbf{CIP}}$ harnesses empowerment to integrate the causal understanding into decision-making. By maximizing the empowerment gain of the causally-informed policy, we guide the agent to prioritize actions that align with the environment's underlying causal relationships. This approach enhances the agent's exploration efficiency, focusing on actions with meaningful causal impacts and correlated with desired outcomes. Algorithm~\ref{alg:algorithm1} illustrates the complete $\texttt{\textbf{CIP}}$ pipeline. 

\begin{figure}[t]
    \centering
    \includegraphics[width=1\textwidth]{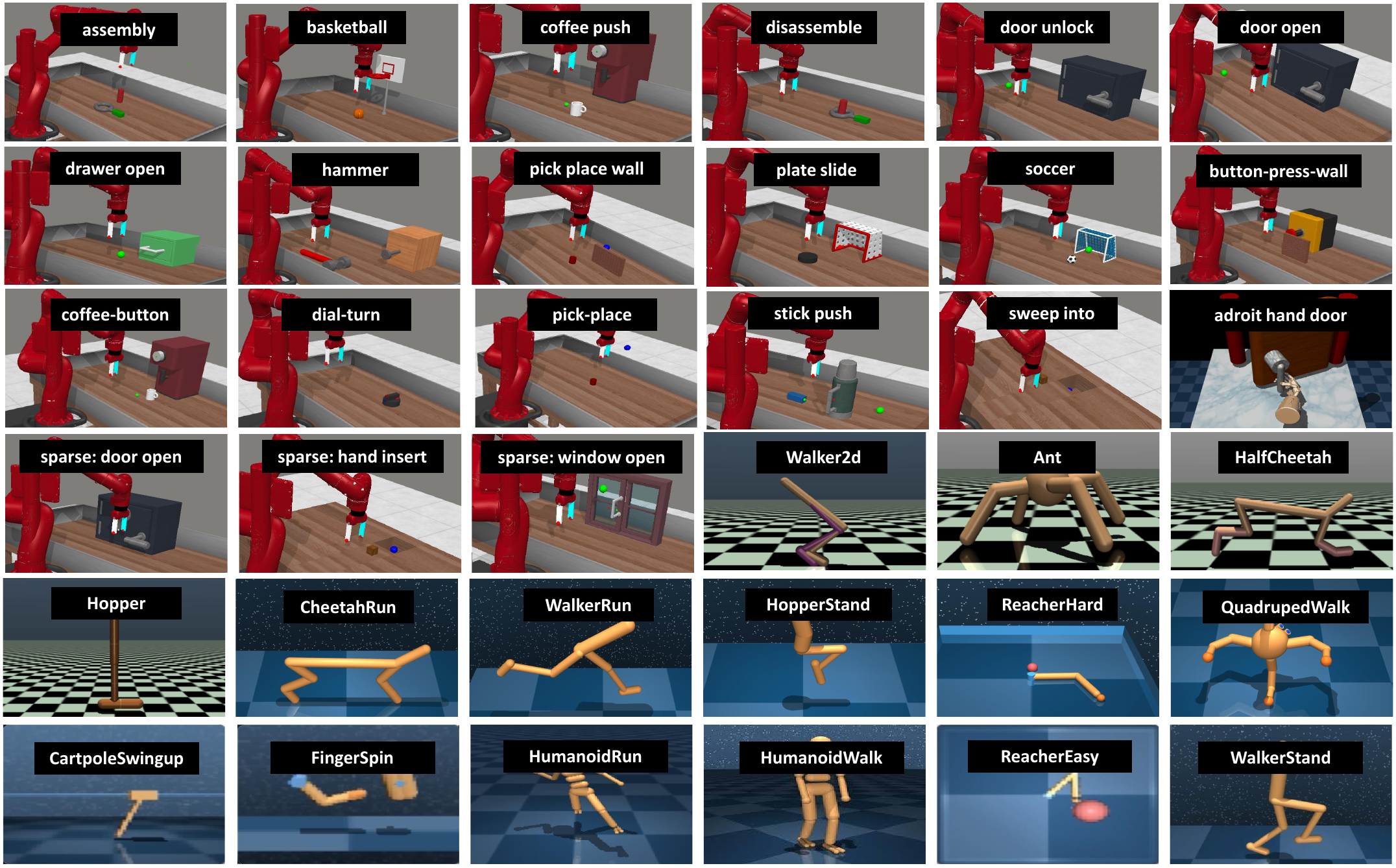}
    \caption{The $36$ experimental tasks in $5$ continuous control environments} 
    \label{fig:environment}
\end{figure}

\begin{figure}[t]
    \centering
    \includegraphics[width=1\linewidth]{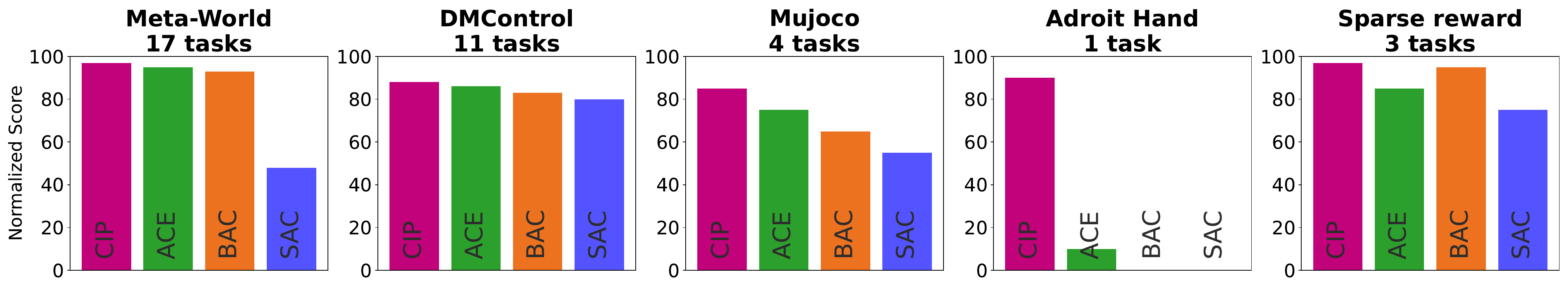}
    \caption{Experimental results with normalized score across all $36$ tasks in $5$ environments.}
    \label{fig:main_results}
    \vspace{-3mm}
\end{figure}
\vspace{-3mm}
\section{Experiments}
\vspace{-3mm}
Our experiments aim to address the following questions:
(i) How does the performance of $\texttt{\textbf{CIP}}$ compare to other RL approaches in diverse continuous control tasks, including manipulation and locomotion with sparse rewards, high-dimensional action spaces, and pixel-based challenges?
(ii) Can $\texttt{\textbf{CIP}}$, through data augmentation and empowerment, improve sample efficiency and learn reliable policies?
(iii) What are the effects of the components and hyperparameters in $\texttt{\textbf{CIP}}$?

\vspace{-3mm}
\subsection{Experimental setup} 
\vspace{-1mm}
\textbf{Environments.} We evaluate $\texttt{\textbf{CIP}}$ on $5$ continuous control environments, including MuJoCo~\citep{todorov2012mujoco}, DMControl~\citep{tassa2018deepmind}, Meta-World~\citep{yu2020meta}, Adroit Hand~\citep{rajeswaran2017learning}, and sparse reward setting environments in Meta-World. 
This comprehensive evaluation encompasses 36 tasks, spanning both locomotion and manipulation skill learning, as illustrated in Figure~\ref{fig:environment}. We also conduct experiments in $4$ pixel-based tasks of the DMControl and Cartpole environment as shown in Figure~\ref{fig:appendix_pixel_env}.
Our experimental tasks incorporate a wide range of challenges, including high-dimensional state and action spaces, sparse reward settings, pixel-based scenarios, and locomotion. 
For extensive experimental settings, please refer to Appendix~\ref{sec:experimental_setup_appendix}. 

\textbf{Baselines.} 
We compare $\texttt{\textbf{CIP}}$ with three popular RL baselines across all $36$ tasks and against IFactor~\citep{liu2024learning} in $3$ pixel-based tasks: 
(1) SAC~\citep{haarnoja2018soft}, an off-policy actor-critic algorithm featuring maximum entropy regularization. 
(2) ACE~\citep{ji2024ace}, a method employing causality-aware entropy regularization. 
(3) BAC~\citep{ji2023seizing}. a method that balances sufficient exploitation of past successes with exploration optimism. 
(4) IFactor~\citep{liu2024learning}. a causal framework modeling four distinct categories of latent state variables for pixel-based tasks. 
To ensure robustness and statistical significance, we conduct each experiment using 4 random seeds. 

\begin{figure}[t]
    \centering
    \includegraphics[width=1\linewidth]{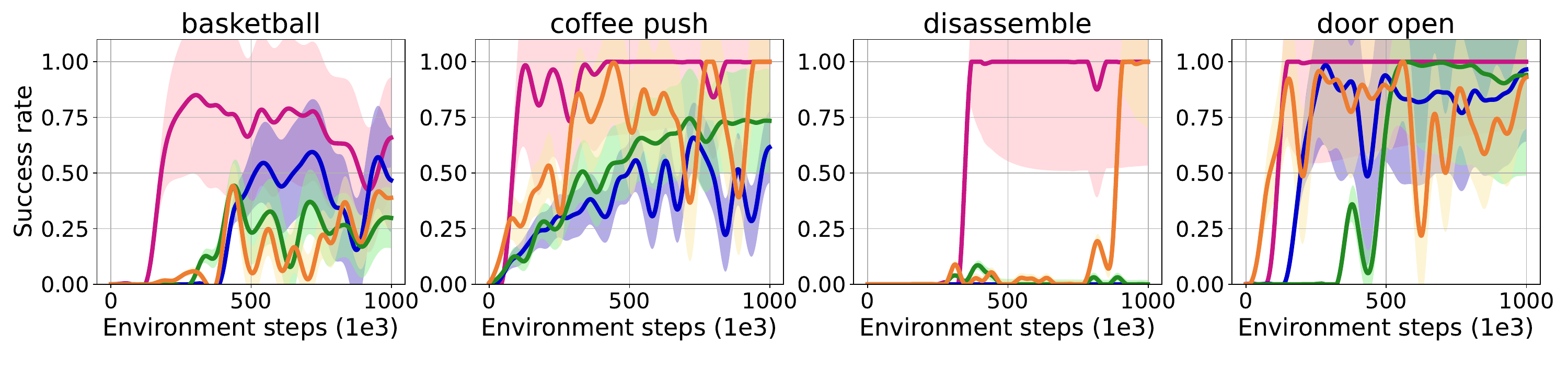}
    \includegraphics[width=1\linewidth]{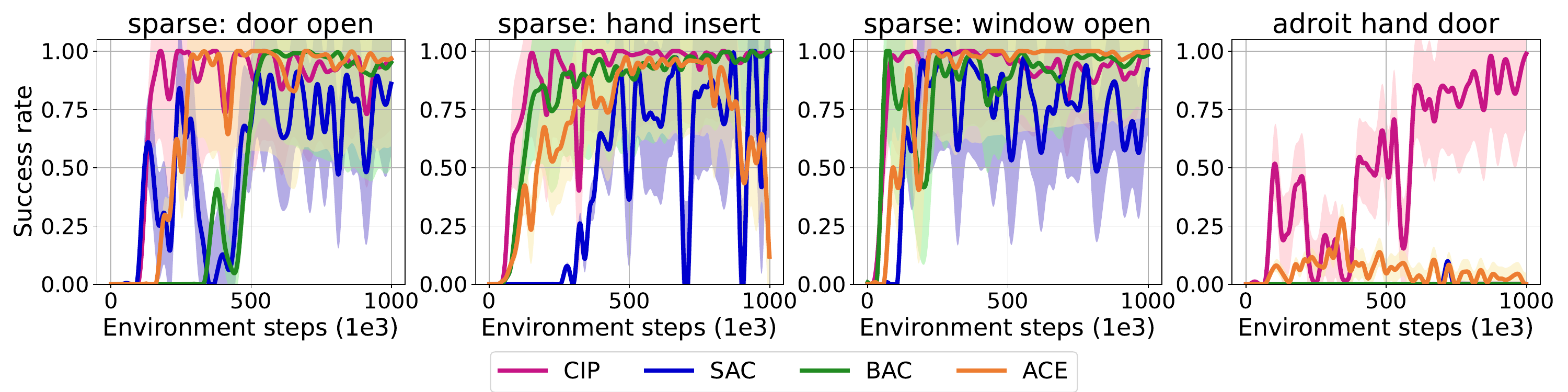}
    \caption{Experimental results of $8$ manipulation skill learning tasks in Meta-World and adroit hand environments including sparse reward settings. For all tasks results, please refer to Appendix~\ref{sec:full_results_appendix}.}
    \label{fig:manipulation}
    \vspace{-2mm}
\end{figure}

\begin{figure}[t]
    \centering
    \includegraphics[width=0.24\textwidth]{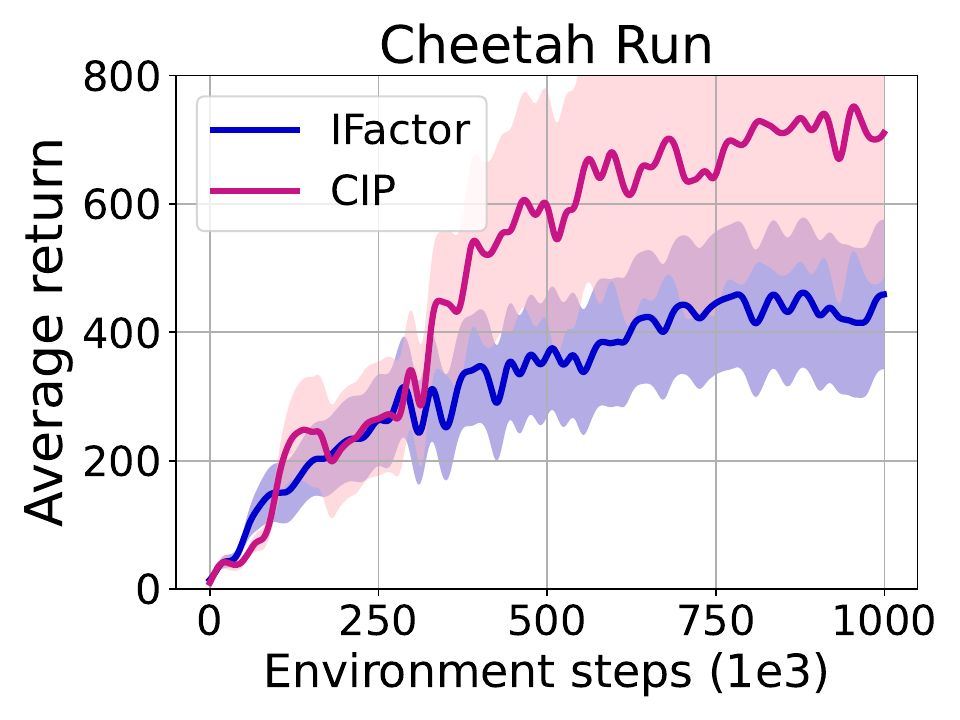}
    \includegraphics[width=0.24\textwidth]{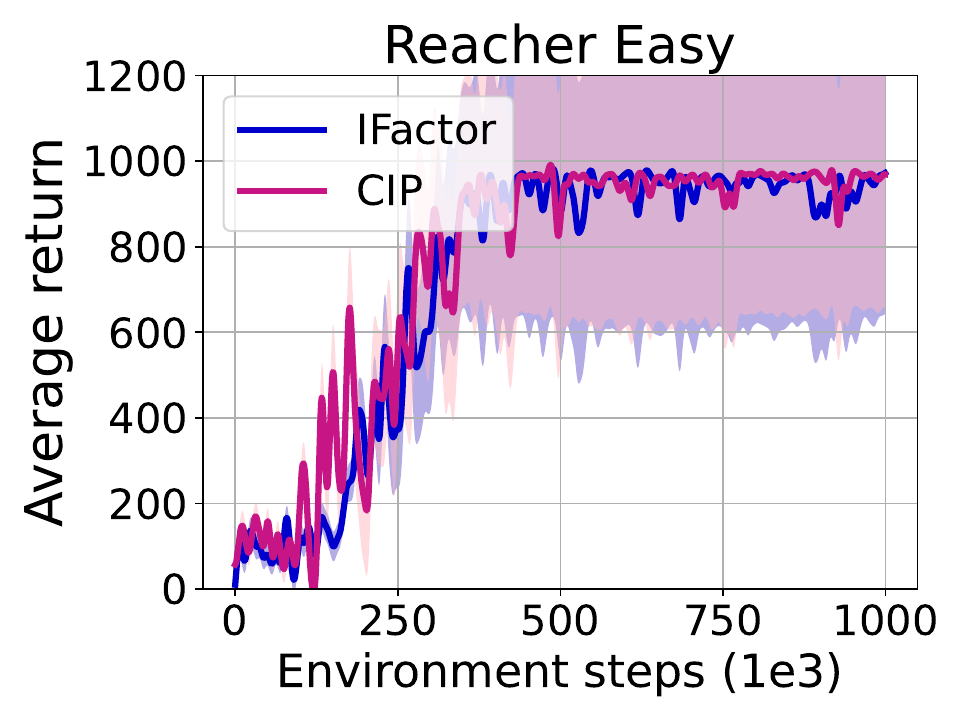}
    \includegraphics[width=0.24\textwidth]{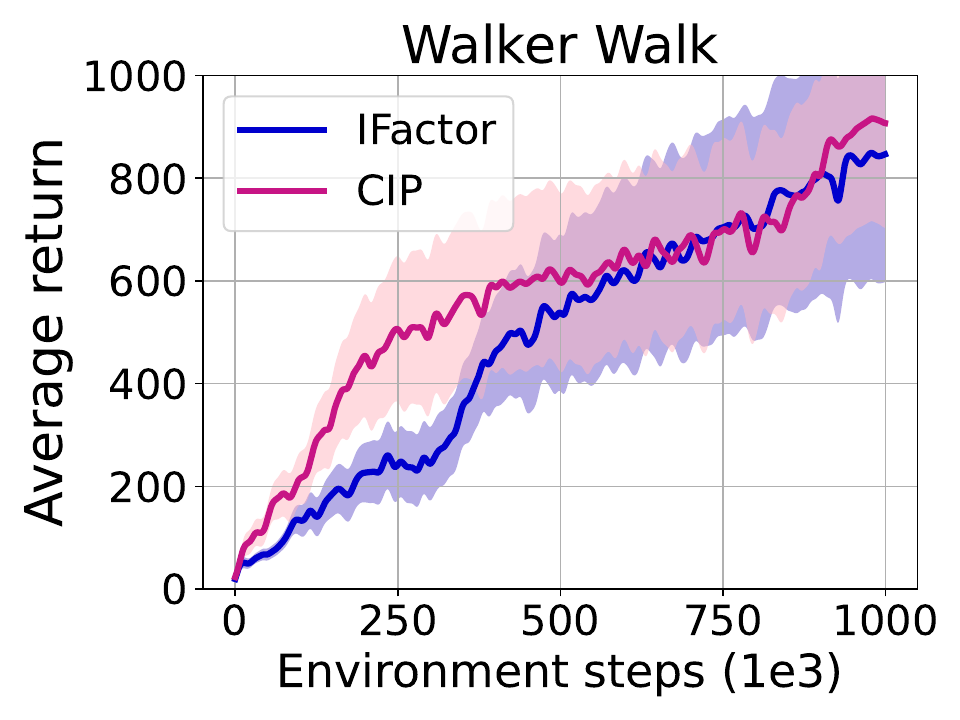}
    \includegraphics[width=0.24\textwidth]{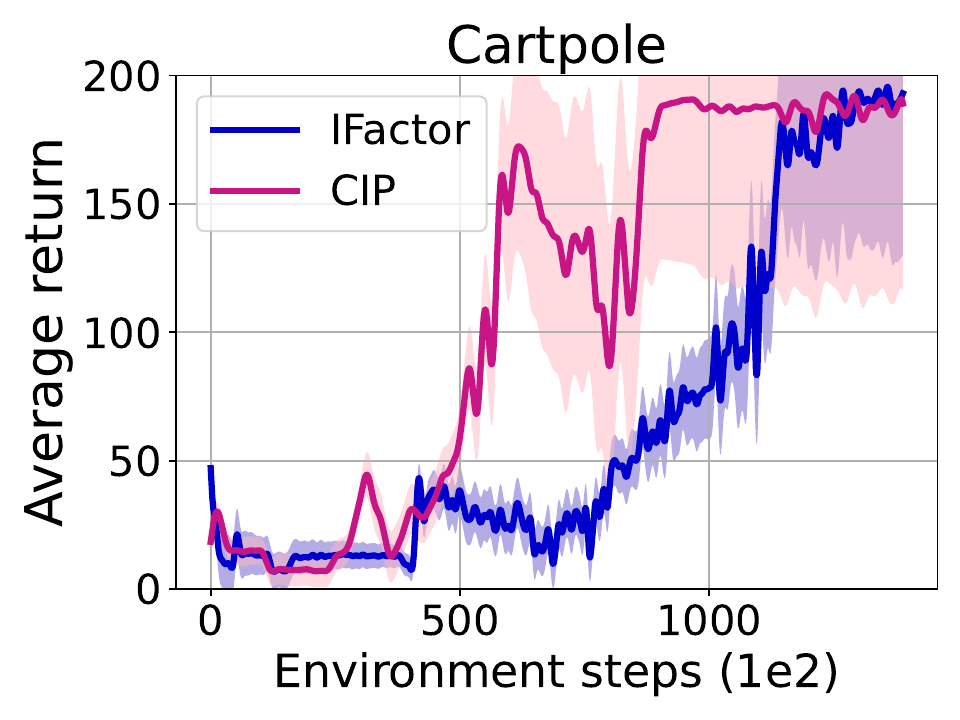}
    \caption{Experimental results of $4$ pixel-based tasks in DMControl and Cartpole environments.}
    \label{fig:appendix_pixel}
    \vspace{-3mm}
\end{figure}

\vspace{-3mm}
\subsection{Main Results}
\vspace{-3mm}
Figure~\ref{fig:main_results} presents the normalized scores of $\texttt{\textbf{CIP}}$ compare to other methods across $36$ tasks in $5$ environments. In $17$ Meta-World robot-arm tasks, $\texttt{\textbf{CIP}}$ achieves a near-perfect score of $100$, showcasing its exceptional performance in manipulation tasks. For locomotion tasks in DMControl and MuJoCo, $\texttt{\textbf{CIP}}$ consistently attains scores exceeding $80$, indicating robust performance across diverse locomotion challenges. Notably, $\texttt{\textbf{CIP}}$ exhibits significant performance improvements in challenging scenarios, such as adroit hand manipulation and $3$ tasks with the sparse reward setting. 
These results underscore the effectiveness in tackling complex, high-dimensional control problems. In next sections, we present a comprehensive analysis of $\texttt{\textbf{CIP}}$'s performance across diverse tasks. 

\textbf{Robot-arm manipulation.} 
Figure~\ref{fig:manipulation} presents the success rates across $7$ Meta-World robot-arm manipulation tasks including sparse reward settings. $\texttt{\textbf{CIP}}$ consistently outperforms all other methods across these tasks, demonstrating both faster policy learning and enhanced stability. 
In challenging tasks, such as disassemble, $\texttt{\textbf{CIP}}$ achieves an impressive 100\% success rate. 
The effectiveness of $\texttt{\textbf{CIP}}$ can be attributed to focus on causally relevant information within the state and action spaces. In sparse reward settings, the efficient extraction of causal state information and the prioritization of controllable actions enable effective task completion. 
By systematically eliminating noise from non-causal factors, $\texttt{\textbf{CIP}}$ allows the agent to construct a more controllable and efficient policy. 

\textbf{High-dimensional Adroit hand manipulation.}
To rigorously evaluate our method's efficacy in high-dimensional tasks, we conduct comparative experiments in the Adroit Hand environment of door open task. This challenging setup involves controlling a robotic hand with up to 28 actuated degrees of freedom ($\mathcal{A} \in \mathbb{R}^{28}$).
Figure~\ref{fig:manipulation} illustrates the success rates achieved across all methods. Notably, while the three other comparative methods fail to demonstrate significant progress on this challenging task, $\texttt{\textbf{CIP}}$ achieves a near $100\%$ success rate after $700k$ environment steps. 

\begin{figure}[h]
    \centering
     \includegraphics[width=1\linewidth]{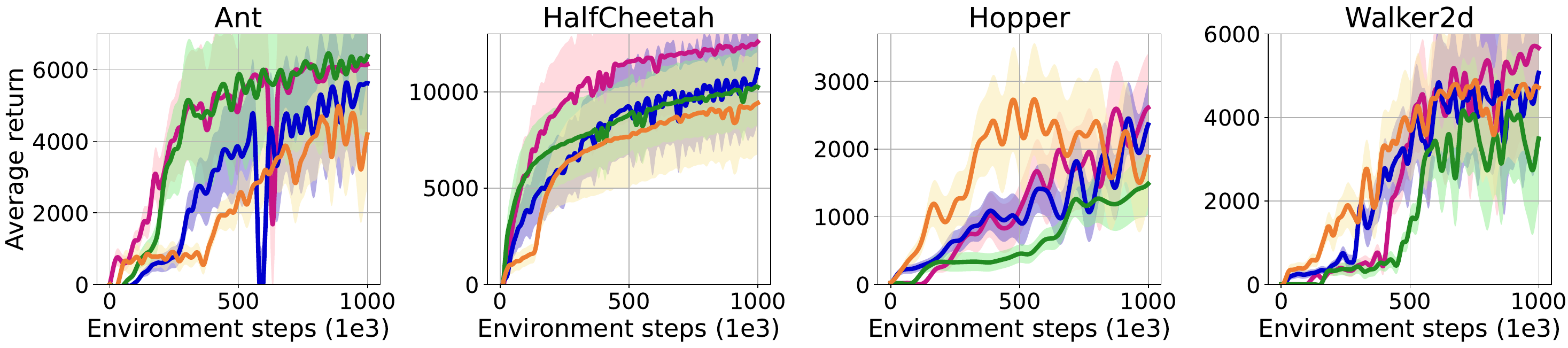}
    \includegraphics[width=1\linewidth]{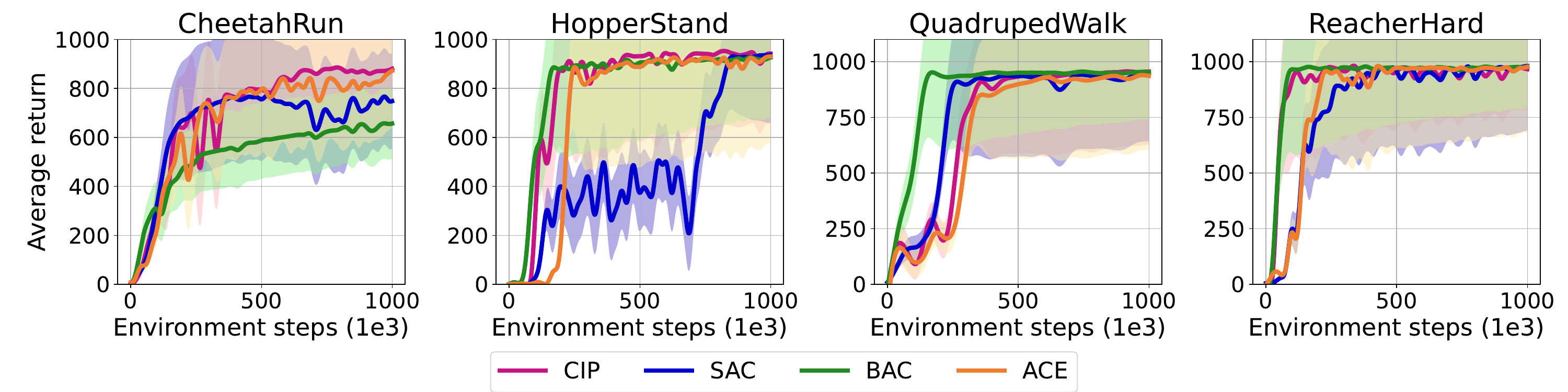}
    \caption{Experimental results with average return across $8$ tasks in locomotion tasks.}
    \label{fig:loco}
    \vspace{-3mm}
\end{figure}

\textbf{Locomotion.}
We further evaluate $\texttt{\textbf{CIP}}$ in another important category: locomotion. The part experimental results of average return in MuJoCo and DMControl environments are presented in Table~\ref{tab1:loco}. Learning curves are illustrated in Figure~\ref{fig:loco}. 
We observe that $\texttt{\textbf{CIP}}$ achieves the best performance in six tasks and sub-optimal in other tasks, and shows statistically significant improvements in 5 out of 8 tasks. Moreover, compared to the traditional method SAC, $\texttt{\textbf{CIP}}$ demonstrates significant performance improvements in more challenging tasks such as CheetahRun and Hopper. 
Compared to the causality-based method ACE, $\texttt{\textbf{CIP}}$ demonstrates improvements in all tasks. 
Overall, in locomotion tasks, $\texttt{\textbf{CIP}}$ achieves superior performance and attains high sample efficiency. \textcolor{black}{Detailed performance and statistical analyses are provided in Appendix~\ref{sec:full_results_appendix} and~\ref{sec:appendix_statis}. }

\begin{table}[t]
\footnotesize
\renewcommand{\arraystretch}{1.4}
\setlength{\tabcolsep}{2.1pt} 
\caption{The experimental results of average return in $8$ locomotion tasks. We bold the best scores, and underline second-best results, $\pm$ is the standard deviation, w/o represents without. \textcolor{black}{$\bullet$ indicates $\texttt{\textbf{CIP}}$ is statistically superior to compared method (pairwise \textit{t}-test at $95\%$ confidence interval).}} 
\label{tab1:loco}
\begin{tabular}{cccccccccc}
\toprule
\textbf{Method}        & \textbf{Ant}  & \textbf{HalfCheetah} & \textbf{Hopper} & \textbf{Walker2d} & \multicolumn{1}{c}{\textbf{\makecell{Cheetah \\ Run}}}
 & 
 \multicolumn{1}{c}{\textbf{\makecell{Hopper \\ Stand}}}
 & 
 \multicolumn{1}{c}{\textbf{\makecell{Quadruped \\ Walk}}}
 & 
 \multicolumn{1}{c}{\textbf{\makecell{Reacher \\ Hard}}}

  \\ 
\hline
\textbf{$\texttt{\textbf{CIP}}$}           & \underline{6418$\pm$81}    & \textbf{12594$\pm$210  }       & \textbf{2846$\pm$882}      & \textbf{5624$\pm$91}     & \textbf{893$\pm$12}        & \textbf{936$\pm$17}            & \underline{948$\pm$54}          & \textbf{991$\pm$11}              
\\
 \textbf{w/o Aug}
 & 6231$\pm$81    & \underline{12225$\pm$102}       & 2308$\pm$785      & 5294$\pm$41     & \underline{885$\pm$13}        & 931$\pm$22            & 945$\pm$35           & \underline{989$\pm$13}            
\\
\textbf{w/o Emp}
 & 6295$\pm$210          & 10986$\pm$572          & 2270$\pm$904           & \underline{5547$\pm$91}       & 876$\pm$21                & 785$\pm$114                  & 924$\pm$23                    & 971$\pm$13                     
\\ 
\hline
\textbf{SAC}           & 6062$\pm$105$\bullet$          & 10888$\pm$240$\bullet$                & 2266$\pm$981            & 5251$\pm$106              & 767$\pm$16$\bullet$                 & \textbf{936$\pm$8}         & 930$\pm$19 $\bullet$                   & 980$\pm$8$\bullet$         
\\
\textbf{BAC}           & \textbf{6511$\pm$30} & 10276$\pm$34$\bullet$                & 2263$\pm$1063$\bullet$   & 3316$\pm$702 $\bullet$             & 665$\pm$6$\bullet$                 & 932$\pm$4$\bullet$                  & \textbf{962$\pm$24}              & 974$\pm$16               
\\
\textbf{ACE}           & 5922$\pm$106 $\bullet$         & 9390$\pm$25$\bullet$                 & \underline{2312$\pm$673$\bullet$}            & 4922$\pm$96$\bullet$              & 863$\pm$23$\bullet$                 & 912$\pm$16                  & 933$\pm$57                    & 973$\pm$17               
\\ \bottomrule
\end{tabular}
\vspace{-2mm}
\end{table}

\textbf{Pixel-based task learning} 
To further validate the performance in pixel-based tasks, we use $3$ complex pixel-based DMControl tasks for evaluation, where video backgrounds serve as distractors. We apply the proposed counterfactual data augmentation and causal action empowerment to IFactor for comparison. 
As shown in Figure~\ref{fig:appendix_pixel}, \texttt{\textbf{CIP}} surpasses IFactor in terms of average return. These results underscore \texttt{\textbf{CIP}}'s efficacy in pixel-based tasks and its capacity to better overcome spurious correlations arising from video backgrounds, focusing on locomotion. \textcolor{black}{Moreover, the result of Cartpole task in Figure~\ref{fig:appendix_pixel} demonstrate the effectiveness in discrete action space environment.}
For visualization trajectories in pixel-based results, please refer to Appendix~\ref{sec:vis_pixel}. 

\vspace{-4mm}
\subsection{Analysis}
\vspace{-2mm}
\paragraph{Ablation study.}
We conduct ablation experiments involving $\texttt{\textbf{CIP}}$, $\texttt{\textbf{CIP}}$ without (w/o) counterfactual data augmentation (Aug), and $\texttt{\textbf{CIP}}$ w/o Empowerment (Emp). 
The results in $8$ locomotion tasks are shown in Table~\ref{tab1:loco}. And the learning curves of all tasks are depicted in Appendix~\ref{sec:pro_res_appendix}. 
The experimental results reveal that the variant without the empowerment learning objective performs poorly, underscoring the critical role of empowerment maximization in enhancing control capabilities. Additionally, $\texttt{\textbf{CIP}}$ without counterfactual data augmentation is less sample efficient than $\texttt{\textbf{CIP}}$, highlighting the importance of augmentation. 

\begin{figure}[t]
    \centering
     \includegraphics[width=0.245\linewidth]{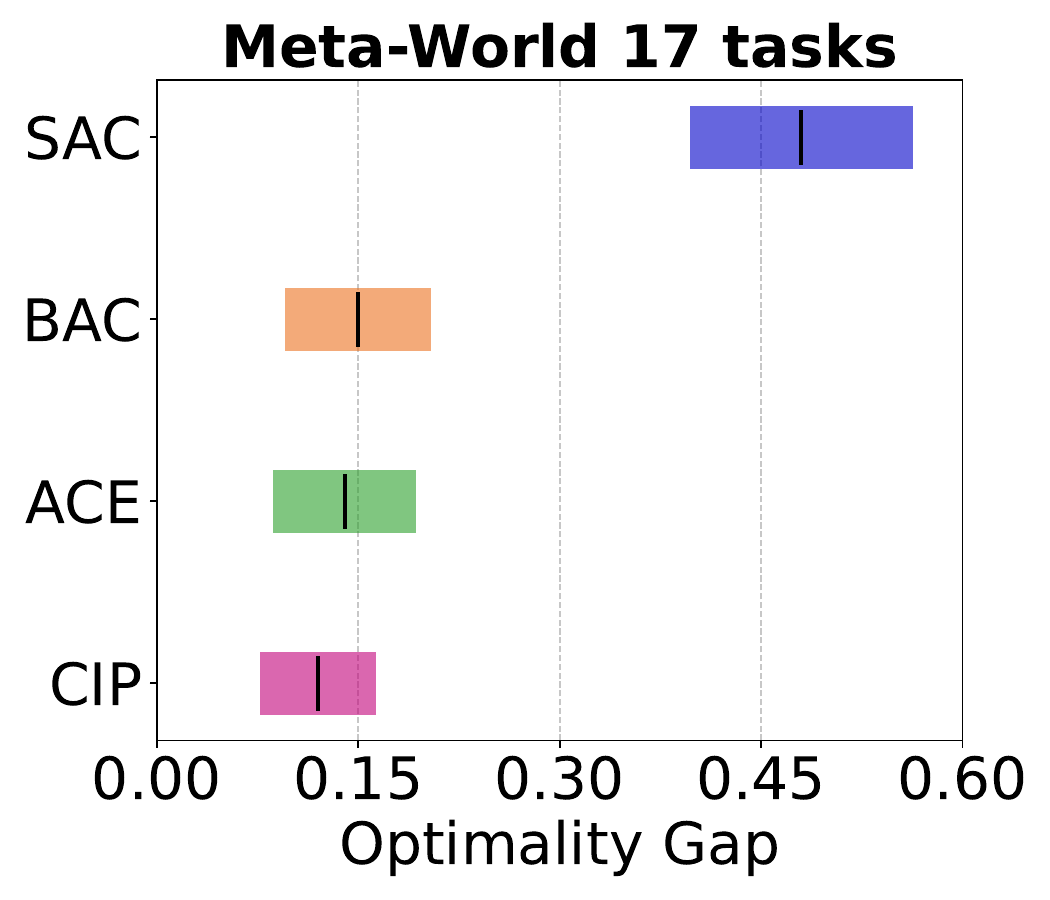}
    \includegraphics[width=0.245\linewidth]{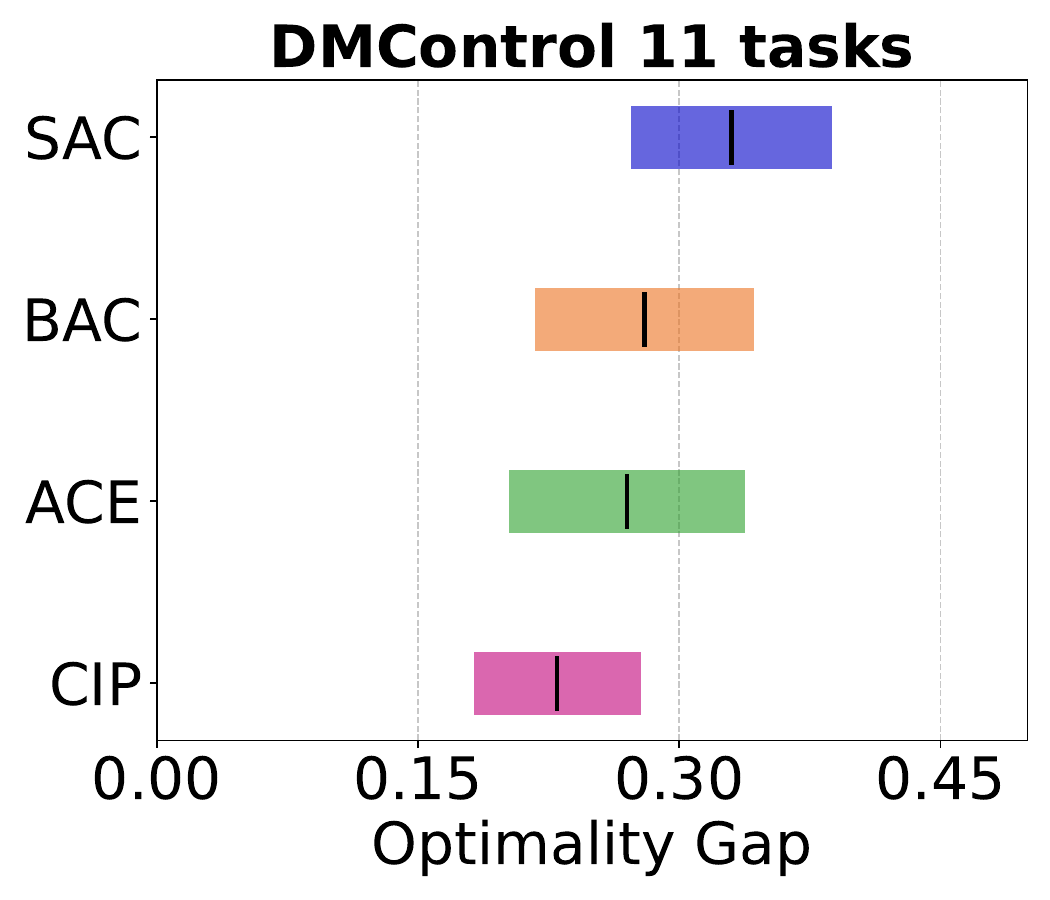}
    \includegraphics[width=0.245\linewidth]{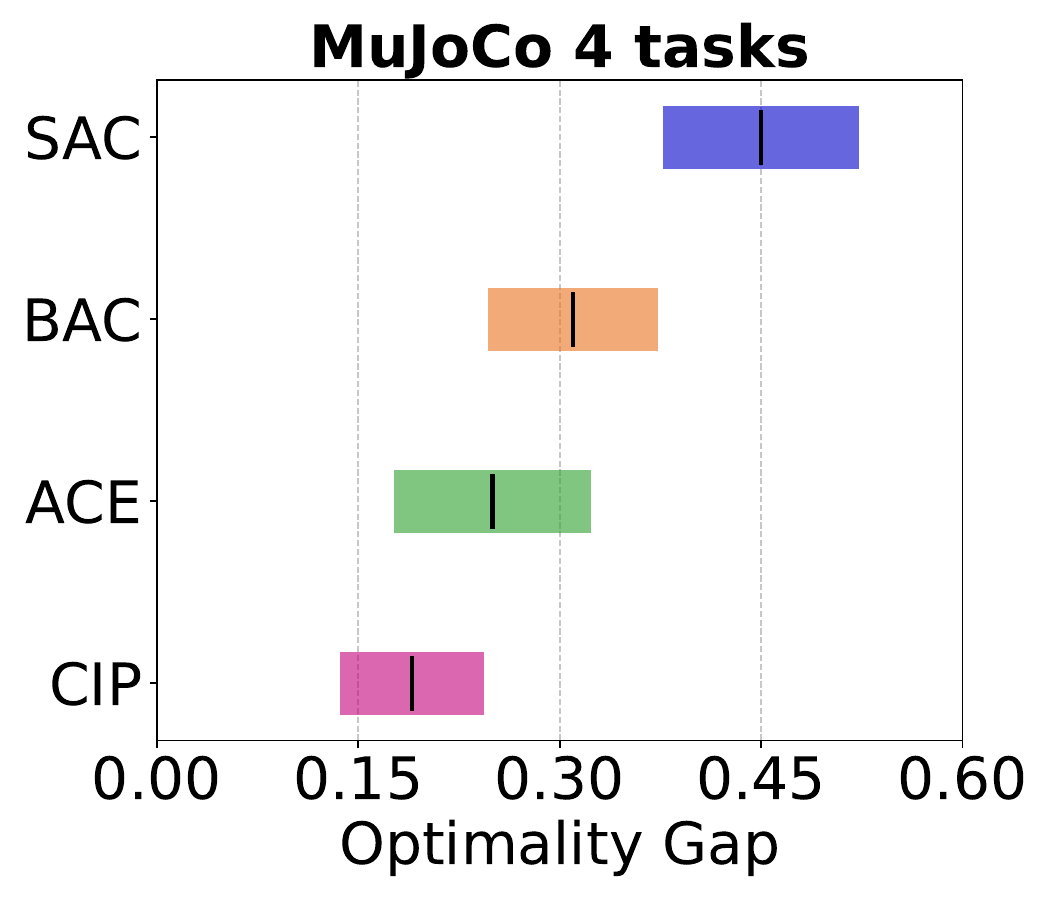}
     \includegraphics[width=0.245\linewidth]{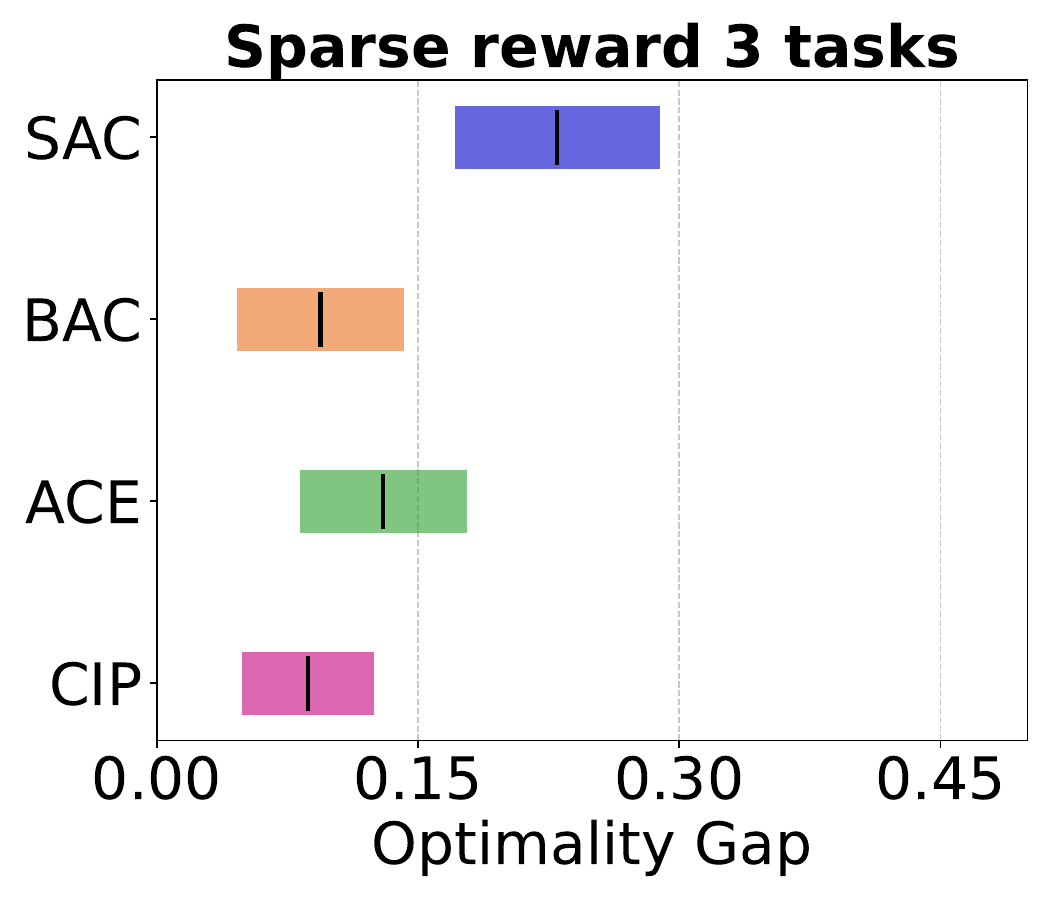}
    \caption{Experimental results of reliability evaluation by the metric Optimality Gap (lower values are better) on $4$ diverse environments across $35$ tasks. }
    \label{fig:rel}
    \vspace{-3mm}
\end{figure}

\vspace{-3mm}
\paragraph{Reliability evaluation.}
We evaluate $\texttt{\textbf{CIP}}$'s reliability across 35 tasks in 4 environments, excluding the Adroit Hand door task due to $\texttt{\textbf{CIP}}$'s exceptional performance there. Figure \ref{fig:rel} illustrates the experimental results using the Optimality Gap metric~\citep{agarwal2021deep}. $\texttt{\textbf{CIP}}$ consistently achieves the lowest values across all tasks in four environments, with lower values indicating superior performance. This consistent excellence across diverse scenarios underscores the robustness and reliability of our proposed method.

\vspace{-3mm}
\section{Conclusion}
\vspace{-3mm}
This study introduces an efficient RL framework, designed to enhance sample efficiency. This approach begins by counterfactual data augmentation using the causality between states and rewards, effectively mitigating interference from irrelevant states without additional environmental interactions. We then develop a reward-guided structural model that leverages causal awareness to prioritize causal actions through empowerment. 
We conduct extensive experiments across $39$ tasks spanning $5$ diverse continuous control environments which demonstrate the exceptional performance of our proposed method, showcasing its robustness and adaptability across challenging scenarios. 

\textbf{Limitation and Future Work}
\quad \textcolor{black}{The current limitations of our work are twofold. First, $\texttt{\textbf{CIP}}$ has not yet been extended to complex scenarios, such as real-world 3D robotics tasks. Potential approaches to address this limitation include leveraging object-centric models~\citep{wu2023slotdiffusion}, 3D perception models~\citep{wang2024rise}, and robotic foundation models~\citep{team2024octo, firoozi2023foundation} to construct essential variables for causal world modeling. Second, $\texttt{\textbf{CIP}}$ does not adequately consider non-stationarity and heterogeneity, which are critical challenges in causal discovery. Future work could integrate method designed to handle such complexities, such as CD-NOD~\citep{huang2020causal}.}

\vspace{-3mm}
\section*{Reproducibility Statement}
\vspace{-3mm}
We provide the core code of $\texttt{\textbf{CIP}}$ in the supplementary material. 
The implementation details are shown in Appendix~\ref{sec:experimental_setup_appendix}.

\section*{Acknowledgment}
This work was supported in part by the National Natural Science Foundation of China under Grant 62276128, Grant 62192783; in part by the Jiangsu Science and Technology Major Project BG2024031; in part by the Natural Science Foundation of Jiangsu Province under Grant BK20243051; the Fundamental Research Funds for the Central Universities(14380128); in part by the Collaborative Innovation Center of Novel Software Technology and Industrialization.

\bibliography{iclr2025_conference}

\begin{thebibliography}{88}
\providecommand{\natexlab}[1]{#1}
\providecommand{\url}[1]{\texttt{#1}}
\expandafter\ifx\csname urlstyle\endcsname\relax
  \providecommand{\doi}[1]{doi: #1}\else
  \providecommand{\doi}{doi: \begingroup \urlstyle{rm}\Url}\fi

\bibitem[Agarwal et~al.(2021)Agarwal, Schwarzer, Castro, Courville, and Bellemare]{agarwal2021deep}
Rishabh Agarwal, Max Schwarzer, Pablo~Samuel Castro, Aaron~C Courville, and Marc Bellemare.
\newblock Deep reinforcement learning at the edge of the statistical precipice.
\newblock \emph{Advances in neural information processing systems}, 34:\penalty0 29304--29320, 2021.

\bibitem[Bharadhwaj et~al.(2022)Bharadhwaj, Babaeizadeh, Erhan, and Levine]{bharadhwaj2022information}
Homanga Bharadhwaj, Mohammad Babaeizadeh, Dumitru Erhan, and Sergey Levine.
\newblock Information prioritization through empowerment in visual model-based rl.
\newblock In \emph{International Conference on Learning Representations}, 2022.

\bibitem[Brady et~al.(2023)Brady, Zimmermann, Sharma, Sch{\"o}lkopf, Von~K{\"u}gelgen, and Brendel]{brady2023provably}
Jack Brady, Roland~S Zimmermann, Yash Sharma, Bernhard Sch{\"o}lkopf, Julius Von~K{\"u}gelgen, and Wieland Brendel.
\newblock Provably learning object-centric representations.
\newblock In \emph{International Conference on Machine Learning}, pp.\  3038--3062. PMLR, 2023.

\bibitem[Cao et~al.(2023)Cao, Yang, Huo, Chen, and Gao]{cao2023enhancing}
Hongye Cao, Shangdong Yang, Jing Huo, Xingguo Chen, and Yang Gao.
\newblock Enhancing ood generalization in offline reinforcement learning with energy-based policy optimization.
\newblock In \emph{ECAI 2023}, pp.\  335--342. IOS Press, 2023.

\bibitem[Cao et~al.(2024)Cao, Feng, Fang, Dong, Huo, and Gao]{cao2024towards}
Hongye Cao, Fan Feng, Meng Fang, Shaokang Dong, Jing Huo, and Yang Gao.
\newblock Towards empowerment gain through causal structure learning in model-based rl.
\newblock In \emph{ICML 2024 Workshop: Foundations of Reinforcement Learning and Control--Connections and Perspectives}, 2024.

\bibitem[Chickering(2002)]{chickering2002optimal}
David~Maxwell Chickering.
\newblock Optimal structure identification with greedy search.
\newblock \emph{Journal of machine learning research}, 3\penalty0 (Nov):\penalty0 507--554, 2002.

\bibitem[Choi et~al.(2021)Choi, Sharma, Lee, Levine, and Gu]{choi2021variational}
Jongwook Choi, Archit Sharma, Honglak Lee, Sergey Levine, and Shixiang~Shane Gu.
\newblock Variational empowerment as representation learning for goal-based reinforcement learning.
\newblock \emph{arXiv preprint arXiv:2106.01404}, 2021.

\bibitem[Choi et~al.(2024)Choi, Lee, Wang, Sohn, and Lee]{choi2024unsupervised}
Jongwook Choi, Sungtae Lee, Xinyu Wang, Sungryull Sohn, and Honglak Lee.
\newblock Unsupervised object interaction learning with counterfactual dynamics models.
\newblock In \emph{Proceedings of the AAAI Conference on Artificial Intelligence}, volume~38, pp.\  11570--11578, 2024.

\bibitem[Deng et~al.(2023)Deng, Jiang, Long, and Zhang]{deng2023causal}
ZH~Deng, J~Jiang, G~Long, and C~Zhang.
\newblock Causal reinforcement learning: A survey.
\newblock \emph{Transactions on Machine Learning Research}, 2023.

\bibitem[Diuk et~al.(2008)Diuk, Cohen, and Littman]{diuk2008object}
Carlos Diuk, Andre Cohen, and Michael~L Littman.
\newblock An object-oriented representation for efficient reinforcement learning.
\newblock In \emph{Proceedings of the 25th international conference on Machine learning}, pp.\  240--247, 2008.

\bibitem[Eslami et~al.(2016)Eslami, Heess, Weber, Tassa, Szepesvari, Hinton, et~al.]{eslami2016attend}
SM~Eslami, Nicolas Heess, Theophane Weber, Yuval Tassa, David Szepesvari, Geoffrey~E Hinton, et~al.
\newblock Attend, infer, repeat: Fast scene understanding with generative models.
\newblock \emph{Advances in neural information processing systems}, 29, 2016.

\bibitem[Eysenbach et~al.(2018)Eysenbach, Gupta, Ibarz, and Levine]{eysenbach2018diversity}
Benjamin Eysenbach, Abhishek Gupta, Julian Ibarz, and Sergey Levine.
\newblock Diversity is all you need: Learning skills without a reward function.
\newblock In \emph{International Conference on Learning Representations}, 2018.

\bibitem[Feng \& Magliacane(2023)Feng and Magliacane]{feng2023learning}
Fan Feng and Sara Magliacane.
\newblock Learning dynamic attribute-factored world models for efficient multi-object reinforcement learning.
\newblock \emph{Advances in Neural Information Processing Systems}, 36, 2023.

\bibitem[Firoozi et~al.(2023)Firoozi, Tucker, Tian, Majumdar, Sun, Liu, Zhu, Song, Kapoor, Hausman, et~al.]{firoozi2023foundation}
Roya Firoozi, Johnathan Tucker, Stephen Tian, Anirudha Majumdar, Jiankai Sun, Weiyu Liu, Yuke Zhu, Shuran Song, Ashish Kapoor, Karol Hausman, et~al.
\newblock Foundation models in robotics: Applications, challenges, and the future.
\newblock \emph{The International Journal of Robotics Research}, pp.\  02783649241281508, 2023.

\bibitem[Guestrin et~al.(2001)Guestrin, Koller, and Parr]{guestrin2001multiagent}
Carlos Guestrin, Daphne Koller, and Ronald Parr.
\newblock Multiagent planning with factored mdps.
\newblock \emph{Advances in neural information processing systems}, 14, 2001.

\bibitem[Guestrin et~al.(2003)Guestrin, Koller, Parr, and Venkataraman]{guestrin2003efficient}
Carlos Guestrin, Daphne Koller, Ronald Parr, and Shobha Venkataraman.
\newblock Efficient solution algorithms for factored mdps.
\newblock \emph{Journal of Artificial Intelligence Research}, 19:\penalty0 399--468, 2003.

\bibitem[Haarnoja et~al.(2018)Haarnoja, Zhou, Abbeel, and Levine]{haarnoja2018soft}
Tuomas Haarnoja, Aurick Zhou, Pieter Abbeel, and Sergey Levine.
\newblock Soft actor-critic: Off-policy maximum entropy deep reinforcement learning with a stochastic actor.
\newblock In \emph{International conference on machine learning}, pp.\  1861--1870. PMLR, 2018.

\bibitem[Haramati et~al.(2023)Haramati, Daniel, and Tamar]{haramati2024entity}
Dan Haramati, Tal Daniel, and Aviv Tamar.
\newblock Entity-centric reinforcement learning for object manipulation from pixels.
\newblock In \emph{NeurIPS 2023 Workshop on Goal-Conditioned Reinforcement Learning}, 2023.

\bibitem[Huang et~al.(2020)Huang, Zhang, Zhang, Ramsey, Sanchez-Romero, Glymour, and Sch{\"o}lkopf]{huang2020causal}
Biwei Huang, Kun Zhang, Jiji Zhang, Joseph Ramsey, Ruben Sanchez-Romero, Clark Glymour, and Bernhard Sch{\"o}lkopf.
\newblock Causal discovery from heterogeneous/nonstationary data.
\newblock \emph{Journal of Machine Learning Research}, 21\penalty0 (89):\penalty0 1--53, 2020.

\bibitem[Huang et~al.(2022{\natexlab{a}})Huang, Feng, Lu, Magliacane, and Zhang]{huangadarl}
Biwei Huang, Fan Feng, Chaochao Lu, Sara Magliacane, and Kun Zhang.
\newblock Adarl: What, where, and how to adapt in transfer reinforcement learning.
\newblock In \emph{International Conference on Learning Representations}, 2022{\natexlab{a}}.

\bibitem[Huang et~al.(2022{\natexlab{b}})Huang, Lu, Leqi, Hern{\'a}ndez-Lobato, Glymour, Sch{\"o}lkopf, and Zhang]{huang2022action}
Biwei Huang, Chaochao Lu, Liu Leqi, Jos{\'e}~Miguel Hern{\'a}ndez-Lobato, Clark Glymour, Bernhard Sch{\"o}lkopf, and Kun Zhang.
\newblock Action-sufficient state representation learning for control with structural constraints.
\newblock In \emph{International Conference on Machine Learning}, pp.\  9260--9279. PMLR, 2022{\natexlab{b}}.

\bibitem[Ji et~al.(2024{\natexlab{a}})Ji, Liang, Zeng, Luo, Xu, Guo, Zheng, Huang, Sun, and Xu]{ji2024ace}
Tianying Ji, Yongyuan Liang, Yan Zeng, Yu~Luo, Guowei Xu, Jiawei Guo, Ruijie Zheng, Furong Huang, Fuchun Sun, and Huazhe Xu.
\newblock Ace: Off-policy actor-critic with causality-aware entropy regularization.
\newblock In \emph{Forty-first International Conference on Machine Learning}, 2024{\natexlab{a}}.

\bibitem[Ji et~al.(2024{\natexlab{b}})Ji, Luo, Sun, Zhan, Zhang, and Xu]{ji2023seizing}
Tianying Ji, Yu~Luo, Fuchun Sun, Xianyuan Zhan, Jianwei Zhang, and Huazhe Xu.
\newblock Seizing serendipity: Exploiting the value of past success in off-policy actor-critic.
\newblock In \emph{Forty-first International Conference on Machine Learning}, 2024{\natexlab{b}}.

\bibitem[Jiang et~al.(2019)Jiang, Janghorbani, De~Melo, and Ahn]{jiang2019scalor}
Jindong Jiang, Sepehr Janghorbani, Gerard De~Melo, and Sungjin Ahn.
\newblock Scalor: Generative world models with scalable object representations.
\newblock In \emph{International Conference on Learning Representations}, 2019.

\bibitem[Jiang et~al.(2023)Jiang, Deng, Singh, and Ahn]{jiang2023object}
Jindong Jiang, Fei Deng, Gautam Singh, and Sungjin Ahn.
\newblock Object-centric slot diffusion.
\newblock \emph{arXiv preprint arXiv:2303.10834}, 2023.

\bibitem[Jiang et~al.(2024)Jiang, Deng, Singh, Lee, and Ahn]{jiang2024slot}
Jindong Jiang, Fei Deng, Gautam Singh, Minseung Lee, and Sungjin Ahn.
\newblock Slot state space models.
\newblock \emph{arXiv preprint arXiv:2406.12272}, 2024.

\bibitem[Jung et~al.(2011)Jung, Polani, and Stone]{jung2011empowerment}
Tobias Jung, Daniel Polani, and Peter Stone.
\newblock Empowerment for continuous agent—environment systems.
\newblock \emph{Adaptive Behavior}, 19\penalty0 (1):\penalty0 16--39, 2011.

\bibitem[Kearns \& Koller(1999)Kearns and Koller]{kearns1999efficient}
Michael Kearns and Daphne Koller.
\newblock Efficient reinforcement learning in factored mdps.
\newblock In \emph{IJCAI}, volume~16, pp.\  740--747, 1999.

\bibitem[Klyubin et~al.(2005)Klyubin, Polani, and Nehaniv]{klyubin2005empowerment}
Alexander~S Klyubin, Daniel Polani, and Chrystopher~L Nehaniv.
\newblock Empowerment: A universal agent-centric measure of control.
\newblock In \emph{2005 ieee congress on evolutionary computation}, volume~1, pp.\  128--135. IEEE, 2005.

\bibitem[Kori et~al.(2024)Kori, Locatello, Santhirasekaram, Toni, Glocker, and Ribeiro]{kori2024identifiable}
Avinash Kori, Francesco Locatello, Ainkaran Santhirasekaram, Francesca Toni, Ben Glocker, and Fabio De~Sousa Ribeiro.
\newblock Identifiable object-centric representation learning via probabilistic slot attention.
\newblock \emph{arXiv preprint arXiv:2406.07141}, 2024.

\bibitem[Kosiorek et~al.(2018)Kosiorek, Kim, Teh, and Posner]{kosiorek2018sequential}
Adam Kosiorek, Hyunjik Kim, Yee~Whye Teh, and Ingmar Posner.
\newblock Sequential attend, infer, repeat: Generative modelling of moving objects.
\newblock \emph{Advances in Neural Information Processing Systems}, 31, 2018.

\bibitem[Kossen et~al.(2019)Kossen, Stelzner, Hussing, Voelcker, and Kersting]{kossen2019structured}
Jannik Kossen, Karl Stelzner, Marcel Hussing, Claas Voelcker, and Kristian Kersting.
\newblock Structured object-aware physics prediction for video modeling and planning.
\newblock In \emph{International Conference on Learning Representations}, 2019.

\bibitem[Kroemer et~al.(2021)Kroemer, Niekum, and Konidaris]{kroemer2021review}
Oliver Kroemer, Scott Niekum, and George Konidaris.
\newblock A review of robot learning for manipulation: Challenges, representations, and algorithms.
\newblock \emph{Journal of machine learning research}, 22\penalty0 (30):\penalty0 1--82, 2021.

\bibitem[Lachapelle et~al.(2024)Lachapelle, Mahajan, Mitliagkas, and Lacoste-Julien]{lachapelle2024additive}
S{\'e}bastien Lachapelle, Divyat Mahajan, Ioannis Mitliagkas, and Simon Lacoste-Julien.
\newblock Additive decoders for latent variables identification and cartesian-product extrapolation.
\newblock \emph{Advances in Neural Information Processing Systems}, 36, 2024.

\bibitem[Leibfried et~al.(2019)Leibfried, Pascual-Diaz, and Grau-Moya]{leibfried2019unified}
Felix Leibfried, Sergio Pascual-Diaz, and Jordi Grau-Moya.
\newblock A unified bellman optimality principle combining reward maximization and empowerment.
\newblock \emph{Advances in Neural Information Processing Systems}, 32, 2019.

\bibitem[Li et~al.(2020)Li, Jabri, Darrell, and Agrawal]{li2020towards}
Richard Li, Allan Jabri, Trevor Darrell, and Pulkit Agrawal.
\newblock Towards practical multi-object manipulation using relational reinforcement learning.
\newblock In \emph{2020 ieee international conference on robotics and automation (icra)}, pp.\  4051--4058. IEEE, 2020.

\bibitem[Li et~al.(2024)Li, Zhang, Geng, Geng, Long, Shen, Zhang, Liu, and Dong]{li2024manipllm}
Xiaoqi Li, Mingxu Zhang, Yiran Geng, Haoran Geng, Yuxing Long, Yan Shen, Renrui Zhang, Jiaming Liu, and Hao Dong.
\newblock Manipllm: Embodied multimodal large language model for object-centric robotic manipulation.
\newblock In \emph{Proceedings of the IEEE/CVF Conference on Computer Vision and Pattern Recognition}, pp.\  18061--18070, 2024.

\bibitem[Li \& Pathak(2024)Li and Pathak]{li2024objectaware}
Yulong Li and Deepak Pathak.
\newblock Object-aware gaussian splatting for robotic manipulation.
\newblock In \emph{ICRA 2024 Workshop on 3D Visual Representations for Robot Manipulation}, 2024.
\newblock URL \url{https://openreview.net/forum?id=gdRI43hDgo}.

\bibitem[Lin et~al.(2020)Lin, Wu, Peri, Fu, Jiang, and Ahn]{lin2020improving}
Zhixuan Lin, Yi-Fu Wu, Skand Peri, Bofeng Fu, Jindong Jiang, and Sungjin Ahn.
\newblock Improving generative imagination in object-centric world models.
\newblock In \emph{International conference on machine learning}, pp.\  6140--6149. PMLR, 2020.

\bibitem[Liu et~al.(2024)Liu, Huang, Zhu, Tian, Gong, Yu, and Zhang]{liu2024learning}
Yuren Liu, Biwei Huang, Zhengmao Zhu, Honglong Tian, Mingming Gong, Yang Yu, and Kun Zhang.
\newblock Learning world models with identifiable factorization.
\newblock \emph{Advances in Neural Information Processing Systems}, 36, 2024.

\bibitem[Locatello et~al.(2020)Locatello, Weissenborn, Unterthiner, Mahendran, Heigold, Uszkoreit, Dosovitskiy, and Kipf]{locatello2020object}
Francesco Locatello, Dirk Weissenborn, Thomas Unterthiner, Aravindh Mahendran, Georg Heigold, Jakob Uszkoreit, Alexey Dosovitskiy, and Thomas Kipf.
\newblock Object-centric learning with slot attention.
\newblock \emph{Advances in neural information processing systems}, 33:\penalty0 11525--11538, 2020.

\bibitem[Mambelli et~al.(2022)Mambelli, Tr{\"a}uble, Bauer, Sch{\"o}lkopf, and Locatello]{mambelli2022compositional}
Davide Mambelli, Frederik Tr{\"a}uble, Stefan Bauer, Bernhard Sch{\"o}lkopf, and Francesco Locatello.
\newblock Compositional multi-object reinforcement learning with linear relation networks.
\newblock In \emph{ICLR2022 Workshop on the Elements of Reasoning: Objects, Structure and Causality}, 2022.

\bibitem[Mitash et~al.(2024)Mitash, Hussein, Vanbaar, Terhuja, and Katyal]{mitash2024scaling}
Chaitanya Mitash, Mostafa Hussein, Jeroen Vanbaar, Vikedo Terhuja, and Kapil Katyal.
\newblock Scaling object-centric robotic manipulation with multimodal object identification.
\newblock 2024.

\bibitem[Mohamed \& Jimenez~Rezende(2015)Mohamed and Jimenez~Rezende]{mohamed2015variational}
Shakir Mohamed and Danilo Jimenez~Rezende.
\newblock Variational information maximisation for intrinsically motivated reinforcement learning.
\newblock \emph{Advances in neural information processing systems}, 28, 2015.

\bibitem[Mutti et~al.(2023)Mutti, De~Santi, Rossi, Calderon, Bronstein, and Restelli]{mutti2023provably}
Mirco Mutti, Riccardo De~Santi, Emanuele Rossi, Juan~Felipe Calderon, Michael Bronstein, and Marcello Restelli.
\newblock Provably efficient causal model-based reinforcement learning for systematic generalization.
\newblock In \emph{Proceedings of the AAAI Conference on Artificial Intelligence}, volume~37, pp.\  9251--9259, 2023.

\bibitem[Park et~al.(2021)Park, Seo, Liu, Zhao, Qin, Shin, and Liu]{park2021object}
Jongjin Park, Younggyo Seo, Chang Liu, Li~Zhao, Tao Qin, Jinwoo Shin, and Tie-Yan Liu.
\newblock Object-aware regularization for addressing causal confusion in imitation learning.
\newblock \emph{Advances in Neural Information Processing Systems}, 34:\penalty0 3029--3042, 2021.

\bibitem[Pearl(2009)]{pearl2009causality}
Judea Pearl.
\newblock \emph{Causality}.
\newblock Cambridge university press, 2009.

\bibitem[Pitis et~al.(2020)Pitis, Creager, and Garg]{pitis2020counterfactual}
Silviu Pitis, Elliot Creager, and Animesh Garg.
\newblock Counterfactual data augmentation using locally factored dynamics.
\newblock \emph{Advances in Neural Information Processing Systems}, 33:\penalty0 3976--3990, 2020.

\bibitem[Pitis et~al.(2022)Pitis, Creager, Mandlekar, and Garg]{pitis2022mocoda}
Silviu Pitis, Elliot Creager, Ajay Mandlekar, and Animesh Garg.
\newblock Mocoda: Model-based counterfactual data augmentation.
\newblock \emph{Advances in Neural Information Processing Systems}, 35:\penalty0 18143--18156, 2022.

\bibitem[Rajeswaran et~al.(2018)Rajeswaran, Kumar, Gupta, Vezzani, Schulman, Todorov, and Levine]{rajeswaran2017learning}
Aravind Rajeswaran, Vikash Kumar, Abhishek Gupta, Giulia Vezzani, John Schulman, Emanuel Todorov, and Sergey Levine.
\newblock Learning complex dexterous manipulation with deep reinforcement learning and demonstrations.
\newblock \emph{Robotics: Science and Systems XIV}, 2018.

\bibitem[Rezaei-Shoshtari et~al.(2022)Rezaei-Shoshtari, Zhao, Panangaden, Meger, and Precup]{rezaei2022continuous}
Sahand Rezaei-Shoshtari, Rosie Zhao, Prakash Panangaden, David Meger, and Doina Precup.
\newblock Continuous mdp homomorphisms and homomorphic policy gradient.
\newblock \emph{Advances in Neural Information Processing Systems}, 35:\penalty0 20189--20204, 2022.

\bibitem[Richens \& Everitt(2024)Richens and Everitt]{richens2024robust}
Jonathan Richens and Tom Everitt.
\newblock Robust agents learn causal world models.
\newblock \emph{arXiv preprint arXiv:2402.10877}, 2024.

\bibitem[Salge et~al.(2014)Salge, Glackin, and Polani]{salge2014empowerment}
Christoph Salge, Cornelius Glackin, and Daniel Polani.
\newblock Empowerment--an introduction.
\newblock \emph{Guided Self-Organization: Inception}, pp.\  67--114, 2014.

\bibitem[Savva et~al.(2019)Savva, Kadian, Maksymets, Zhao, Wijmans, Jain, Straub, Liu, Koltun, Malik, et~al.]{savva2019habitat}
Manolis Savva, Abhishek Kadian, Oleksandr Maksymets, Yili Zhao, Erik Wijmans, Bhavana Jain, Julian Straub, Jia Liu, Vladlen Koltun, Jitendra Malik, et~al.
\newblock Habitat: A platform for embodied ai research.
\newblock In \emph{Proceedings of the IEEE/CVF international conference on computer vision}, pp.\  9339--9347, 2019.

\bibitem[Scholz et~al.(2014)Scholz, Levihn, Isbell, and Wingate]{scholz2014physics}
Jonathan Scholz, Martin Levihn, Charles Isbell, and David Wingate.
\newblock A physics-based model prior for object-oriented mdps.
\newblock In \emph{International Conference on Machine Learning}, pp.\  1089--1097. PMLR, 2014.

\bibitem[Seitzer et~al.(2021)Seitzer, Sch{\"o}lkopf, and Martius]{seitzer2021causal}
Maximilian Seitzer, Bernhard Sch{\"o}lkopf, and Georg Martius.
\newblock Causal influence detection for improving efficiency in reinforcement learning.
\newblock \emph{Advances in Neural Information Processing Systems}, 34:\penalty0 22905--22918, 2021.

\bibitem[Shimizu et~al.(2011)Shimizu, Inazumi, Sogawa, Hyvarinen, Kawahara, Washio, Hoyer, Bollen, and Hoyer]{shimizu2011directlingam}
Shohei Shimizu, Takanori Inazumi, Yasuhiro Sogawa, Aapo Hyvarinen, Yoshinobu Kawahara, Takashi Washio, Patrik~O Hoyer, Kenneth Bollen, and Patrik Hoyer.
\newblock Directlingam: A direct method for learning a linear non-gaussian structural equation model.
\newblock \emph{Journal of Machine Learning Research-JMLR}, 12\penalty0 (Apr):\penalty0 1225--1248, 2011.

\bibitem[Silver et~al.(2017)Silver, Schrittwieser, Simonyan, Antonoglou, Huang, Guez, Hubert, Baker, Lai, Bolton, et~al.]{silver2017mastering}
David Silver, Julian Schrittwieser, Karen Simonyan, Ioannis Antonoglou, Aja Huang, Arthur Guez, Thomas Hubert, Lucas Baker, Matthew Lai, Adrian Bolton, et~al.
\newblock Mastering the game of go without human knowledge.
\newblock \emph{nature}, 550\penalty0 (7676):\penalty0 354--359, 2017.

\bibitem[Spirtes et~al.(2001)Spirtes, Glymour, and Scheines]{spirtes2001causation}
Peter Spirtes, Clark Glymour, and Richard Scheines.
\newblock \emph{Causation, prediction, and search}.
\newblock MIT press, 2001.

\bibitem[Sun \& Wang()Sun and Wang]{sun2022toward}
Hao Sun and Taiyi Wang.
\newblock Toward causal-aware rl: State-wise action-refined temporal difference.
\newblock In \emph{Deep Reinforcement Learning Workshop NeurIPS 2022}.

\bibitem[Sun et~al.(2024)Sun, Wang, Huang, Lu, Feng, Sun, and Zhang]{sun2024acamda}
Yuewen Sun, Erli Wang, Biwei Huang, Chaochao Lu, Lu~Feng, Changyin Sun, and Kun Zhang.
\newblock Acamda: Improving data efficiency in reinforcement learning through guided counterfactual data augmentation.
\newblock In \emph{Proceedings of the AAAI Conference on Artificial Intelligence}, volume~38, pp.\  15193--15201, 2024.

\bibitem[Sutton(2018)]{sutton2018reinforcement}
Richard~S Sutton.
\newblock Reinforcement learning: An introduction.
\newblock \emph{A Bradford Book}, 2018.

\bibitem[Tassa et~al.(2018)Tassa, Doron, Muldal, Erez, Li, Casas, Budden, Abdolmaleki, Merel, Lefrancq, et~al.]{tassa2018deepmind}
Yuval Tassa, Yotam Doron, Alistair Muldal, Tom Erez, Yazhe Li, Diego de~Las Casas, David Budden, Abbas Abdolmaleki, Josh Merel, Andrew Lefrancq, et~al.
\newblock Deepmind control suite.
\newblock \emph{arXiv preprint arXiv:1801.00690}, 2018.

\bibitem[Team et~al.(2024)Team, Ghosh, Walke, Pertsch, Black, Mees, Dasari, Hejna, Kreiman, Xu, et~al.]{team2024octo}
Octo~Model Team, Dibya Ghosh, Homer Walke, Karl Pertsch, Kevin Black, Oier Mees, Sudeep Dasari, Joey Hejna, Tobias Kreiman, Charles Xu, et~al.
\newblock Octo: An open-source generalist robot policy.
\newblock \emph{arXiv preprint arXiv:2405.12213}, 2024.

\bibitem[Todorov et~al.(2012)Todorov, Erez, and Tassa]{todorov2012mujoco}
Emanuel Todorov, Tom Erez, and Yuval Tassa.
\newblock Mujoco: A physics engine for model-based control.
\newblock In \emph{2012 IEEE/RSJ international conference on intelligent robots and systems}, pp.\  5026--5033. IEEE, 2012.

\bibitem[Urp{\'\i} et~al.(2024)Urp{\'\i}, Bagatella, Vlastelica, and Martius]{urpicausal}
N{\'u}ria~Armengol Urp{\'\i}, Marco Bagatella, Marin Vlastelica, and Georg Martius.
\newblock Causal action influence aware counterfactual data augmentation.
\newblock In \emph{Forty-first International Conference on Machine Learning}, 2024.

\bibitem[Van~der Pol et~al.(2020)Van~der Pol, Kipf, Oliehoek, and Welling]{van2020plannable}
Elise Van~der Pol, Thomas Kipf, Frans~A Oliehoek, and Max Welling.
\newblock Plannable approximations to mdp homomorphisms: Equivariance under actions.
\newblock \emph{arXiv preprint arXiv:2002.11963}, 2020.

\bibitem[Wandzel et~al.(2019)Wandzel, Oh, Fishman, Kumar, Wong, and Tellex]{wandzel2019multi}
Arthur Wandzel, Yoonseon Oh, Michael Fishman, Nishanth Kumar, Lawson~LS Wong, and Stefanie Tellex.
\newblock Multi-object search using object-oriented pomdps.
\newblock In \emph{2019 International Conference on Robotics and Automation (ICRA)}, pp.\  7194--7200. IEEE, 2019.

\bibitem[Wang et~al.(2024{\natexlab{a}})Wang, Fang, Fang, and Lu]{wang2024rise}
Chenxi Wang, Hongjie Fang, Hao-Shu Fang, and Cewu Lu.
\newblock Rise: 3d perception makes real-world robot imitation simple and effective.
\newblock \emph{arXiv preprint arXiv:2404.12281}, 2024{\natexlab{a}}.

\bibitem[Wang et~al.(2021)Wang, Xiao, Zhu, and Stone]{wang2021task}
Zizhao Wang, Xuesu Xiao, Yuke Zhu, and Peter Stone.
\newblock Task-independent causal state abstraction.
\newblock In \emph{Proceedings of the 35th International Conference on Neural Information Processing Systems, Robot Learning workshop}, 2021.

\bibitem[Wang et~al.(2022)Wang, Xiao, Xu, Zhu, and Stone]{wang2022causal}
Zizhao Wang, Xuesu Xiao, Zifan Xu, Yuke Zhu, and Peter Stone.
\newblock Causal dynamics learning for task-independent state abstraction.
\newblock In \emph{International Conference on Machine Learning}, pp.\  23151--23180. PMLR, 2022.

\bibitem[Wang et~al.(2024{\natexlab{b}})Wang, Hu, Chuck, Chen, Mart{\'\i}n-Mart{\'\i}n, Zhang, Niekum, and Stone]{wang2024skild}
Zizhao Wang, Jiaheng Hu, Caleb Chuck, Stephen Chen, Roberto Mart{\'\i}n-Mart{\'\i}n, Amy Zhang, Scott Niekum, and Peter Stone.
\newblock Skild: Unsupervised skill discovery guided by factor interactions.
\newblock In \emph{The Thirty-eighth Annual Conference on Neural Information Processing Systems}, 2024{\natexlab{b}}.

\bibitem[Wang et~al.(2024{\natexlab{c}})Wang, Hu, Stone, and Mart{\'\i}n-Mart{\'\i}n]{hu2023elden}
Zizhao Wang, Jiaheng Hu, Peter Stone, and Roberto Mart{\'\i}n-Mart{\'\i}n.
\newblock Elden: exploration via local dependencies.
\newblock \emph{Advances in Neural Information Processing Systems}, 36, 2024{\natexlab{c}}.

\bibitem[Wang et~al.(2024{\natexlab{d}})Wang, Wang, Xiao, Zhu, and Stone]{wang2024building}
Zizhao Wang, Caroline Wang, Xuesu Xiao, Yuke Zhu, and Peter Stone.
\newblock Building minimal and reusable causal state abstractions for reinforcement learning.
\newblock \emph{arXiv preprint arXiv:2401.12497}, 2024{\natexlab{d}}.

\bibitem[Watters et~al.(2019)Watters, Matthey, Bosnjak, Burgess, and Lerchner]{watters2019cobra}
Nicholas Watters, Loic Matthey, Matko Bosnjak, Christopher~P Burgess, and Alexander Lerchner.
\newblock Cobra: Data-efficient model-based rl through unsupervised object discovery and curiosity-driven exploration.
\newblock \emph{arXiv preprint arXiv:1905.09275}, 2019.

\bibitem[Wolfe \& Barto(2006)Wolfe and Barto]{wolfe2006defining}
Alicia~Peregrin Wolfe and Andrew~G Barto.
\newblock Defining object types and options using mdp homomorphisms.
\newblock In \emph{Proceedings of the ICML-06 Workshop on Structural Knowledge Transfer for Machine Learning}, 2006.

\bibitem[Wu et~al.(2022)Wu, Dvornik, Greff, Kipf, and Garg]{wu2022slotformer}
Ziyi Wu, Nikita Dvornik, Klaus Greff, Thomas Kipf, and Animesh Garg.
\newblock Slotformer: Unsupervised visual dynamics simulation with object-centric models.
\newblock In \emph{The Eleventh International Conference on Learning Representations}, 2022.

\bibitem[Wu et~al.(2023)Wu, Hu, Lu, Gilitschenski, and Garg]{wu2023slotdiffusion}
Ziyi Wu, Jingyu Hu, Wuyue Lu, Igor Gilitschenski, and Animesh Garg.
\newblock Slotdiffusion: Object-centric generative modeling with diffusion models.
\newblock \emph{Advances in Neural Information Processing Systems}, 36:\penalty0 50932--50958, 2023.

\bibitem[Wu et~al.(2024)Wu, Rubanova, Kabra, Hudson, Gilitschenski, Aytar, van Steenkiste, Allen, and Kipf]{wu2024neural}
Ziyi Wu, Yulia Rubanova, Rishabh Kabra, Drew~A Hudson, Igor Gilitschenski, Yusuf Aytar, Sjoerd van Steenkiste, Kelsey~R Allen, and Thomas Kipf.
\newblock Neural assets: 3d-aware multi-object scene synthesis with image diffusion models.
\newblock \emph{arXiv preprint arXiv:2406.09292}, 2024.

\bibitem[Yu et~al.(2020)Yu, Quillen, He, Julian, Hausman, Finn, and Levine]{yu2020meta}
Tianhe Yu, Deirdre Quillen, Zhanpeng He, Ryan Julian, Karol Hausman, Chelsea Finn, and Sergey Levine.
\newblock Meta-world: A benchmark and evaluation for multi-task and meta reinforcement learning.
\newblock In \emph{Conference on robot learning}, pp.\  1094--1100. PMLR, 2020.

\bibitem[Yuan et~al.(2022)Yuan, Paxton, Desingh, and Fox]{yuan2022sornet}
Wentao Yuan, Chris Paxton, Karthik Desingh, and Dieter Fox.
\newblock Sornet: Spatial object-centric representations for sequential manipulation.
\newblock In \emph{Conference on Robot Learning}, pp.\  148--157. PMLR, 2022.

\bibitem[Zadaianchuk et~al.(2021)Zadaianchuk, Seitzer, and Martius]{zadaianchuk2020self}
Andrii Zadaianchuk, Maximilian Seitzer, and Georg Martius.
\newblock Self-supervised visual reinforcement learning with object-centric representations.
\newblock In \emph{International Conference on Learning Representations}, 2021.

\bibitem[Zadaianchuk et~al.(2022)Zadaianchuk, Martius, and Yang]{zadaianchuk2022self}
Andrii Zadaianchuk, Georg Martius, and Fanny Yang.
\newblock Self-supervised reinforcement learning with independently controllable subgoals.
\newblock In \emph{Conference on Robot Learning}, pp.\  384--394. PMLR, 2022.

\bibitem[Zadaianchuk et~al.(2023)Zadaianchuk, Seitzer, and Martius]{zadaianchuk2023objectcentric}
Andrii Zadaianchuk, Maximilian Seitzer, and Georg Martius.
\newblock Object-centric learning for real-world videos by predicting temporal feature similarities.
\newblock In \emph{Thirty-seventh Conference on Neural Information Processing Systems (NeurIPS 2023)}, 2023.

\bibitem[Ze et~al.(2024)Ze, Liu, Shi, Qin, Yuan, Wang, and Xu]{ze2024h}
Yanjie Ze, Yuyao Liu, Ruizhe Shi, Jiaxin Qin, Zhecheng Yuan, Jiashun Wang, and Huazhe Xu.
\newblock H-index: Visual reinforcement learning with hand-informed representations for dexterous manipulation.
\newblock \emph{Advances in Neural Information Processing Systems}, 36, 2024.

\bibitem[Zeng et~al.(2023)Zeng, Cai, Sun, Huang, and Hao]{zeng2023survey}
Yan Zeng, Ruichu Cai, Fuchun Sun, Libo Huang, and Zhifeng Hao.
\newblock A survey on causal reinforcement learning.
\newblock \emph{arXiv preprint arXiv:2302.05209}, 2023.

\bibitem[Zhang et~al.(2024)Zhang, Du, Huang, Wang, Wang, Fang, and Pechenizkiy]{zhang2024interpretable}
Yudi Zhang, Yali Du, Biwei Huang, Ziyan Wang, Jun Wang, Meng Fang, and Mykola Pechenizkiy.
\newblock Interpretable reward redistribution in reinforcement learning: a causal approach.
\newblock \emph{Advances in Neural Information Processing Systems}, 36, 2024.

\bibitem[Zhao et~al.(2022)Zhao, Kong, Walters, and Wong]{zhao2022toward}
Linfeng Zhao, Lingzhi Kong, Robin Walters, and Lawson~LS Wong.
\newblock Toward compositional generalization in object-oriented world modeling.
\newblock In \emph{International Conference on Machine Learning}, pp.\  26841--26864. PMLR, 2022.

\end{thebibliography}
\bibliographystyle{iclr2025_conference}

\maketitle

\clearpage
\appendix
\tableofcontents
\clearpage

\section{Broader Impact}
\label{Broader Impact}
To avoid blind exploration and improve sample efficiency, we propose $\texttt{\textbf{CIP}}$ for efficient reinforcement learning. $\texttt{\textbf{CIP}}$ leverages the causal relationships among states, actions, and rewards to prioritize causal information for efficient policy learning. 
$\texttt{\textbf{CIP}}$ first learns a causal matrix between states and rewards to execute counterfactual data augmentation, prioritizing important state features without additional environmental interactions. Subsequently, it learns a causal reweight matrix between actions and rewards to prioritize causally-informed behaviors. We then introduce a causal action empowerment term into the learning objective to enhance the controllability. By prioritizing the causal information, $\texttt{\textbf{CIP}}$ enables agents to focus on behaviors that have causally significant effects on their tasks. 
$\texttt{\textbf{CIP}}$ offers substantial broader impact by prioritizing causal information through individual assessment of how different factors contribute to rewards. Our novel empowerment learning objective achieves efficient policy optimization by leveraging entropy via the policy and learned inverse dynamics model. This approach shows promise for extension into research frameworks centered on maximum entropy algorithms. 

Despite its strengths, $\texttt{\textbf{CIP}}$ has limitations beyond its reliance on the method DirectLiNGAM. There's potential to explore alternative causal discovery techniques for more robust relationship mapping. Moreover, analyzing inter-entity causal connections could lead to better disentanglement of diverse behaviors. 
Our future work will investigate a range of causal discovery methods to refine our approach. We aim to extend $\texttt{\textbf{CIP}}$ to model-based RL frameworks, focusing on building causal world models to enhance generalization.

\section{Assumptions and Propositions}
\label{sec:app_ass}
\begin{assumption}
(d-separation~\citep{pearl2009causality}) d-separation is a graphical criterion used to determine, from a given causal graph, if a set of variables X is conditionally independent of another set Y, given a third set of variables Z. 
In a directed acyclic graph (DAG) $\mathcal{G}$, a path between nodes $n_1$ and $n_m$ is said to be blocked by a set $S$ if there exists a node $n_k$, for $k = 2, \cdots, m-1$, that satisfies one of the following two conditions:

(i) $n_k \in S$, and the path between $n_{k-1}$ and $n_{k+1}$ forms ($n_{k-1} \rightarrow n_k \rightarrow n_{k+1}$), ($n_{k-1} \leftarrow n_k \leftarrow n_{k+1}$), or ($n_{k-1} \leftarrow n_k \rightarrow n_{k+1}$). 

(ii) Neither $n_k$ nor any of its descendants is in $S$, and the path between $n_{k-1}$ and $n_{k+1}$ forms ($n_{k-1} \rightarrow n_k \leftarrow n_{k+1}$). 

In a DAG, we say that two nodes $n_a$ and $n_b$ are d-separated by a third node $n_c$ if every path between nodes $n_a$ and $n_b$ is blocked by $n_c$, denoted as $n_a  \! \perp \!\!\! \perp n_b|n_c$. 
\end{assumption}

\begin{assumption}
    (Global Markov Condition~\citep{spirtes2001causation, pearl2009causality}) The state is fully observable and the dynamics is Markovian. The distribution $p$ over a set of variables $\mathcal{V}=(s^1_{t},\cdots,s^d_{t},a^1_{t},\cdots,a^d_{t},r_t)^T$ satisfies the global Markov condition on the graph if for any partition $(\mathcal{S, A, R})$ in $\mathcal{V}$ such that if $\mathcal{A}$ d-separates $\mathcal{S}$ from $\mathcal{R}$, then $p(\mathcal{S},\mathcal{R}|\mathcal{A}) = p(\mathcal{S}|\mathcal{A})\cdot p(\mathcal{R}|\mathcal{A})$
\end{assumption}

\begin{assumption}
    (Faithfulness Assumption~\citep{spirtes2001causation, pearl2009causality}) 
For a set of variables $\mathcal{V}=(s^1_{t},\cdots,s^d_{t},a^1_{t},\cdots,a^d_{t},r_t)^T$, there are no independencies between variables that are not implied by the Markovian Condition.
\end{assumption}

\begin{assumption}
Under the assumptions that the causal graph is Markov and faithful to the observations, the edge $s^i_t \to  s^i_{t+1}$ exists for all state variables $s^i$.
\end{assumption}

\begin{assumption}
No simultaneous or backward edges in time.
\end{assumption}

\paragraph{Proposition 1}
\textit{Under the assumptions that the causal graph is Markov and faithful to the observations, there exists an edge from $a^i_t \to r_t$ if and only if $a^i_t \not \! \perp \!\!\! \perp r_t | { a_t \setminus a^i_t, s_t }$.}

\textit{Proof:} We proceed by proving both directions of the if and only if statement.

($\Rightarrow$) Suppose there exists an edge from $a^i_t$ to $r_t$. We prove that $a^i_t \not \! \perp \!\!\! \perp r_t | { a_t \setminus a^i_t, s_t }$ by contradiction. Assume $a^i_t \! \perp \!\!\! \perp r_t | { a_t \setminus a^i_t, s_t }$. By the faithfulness assumption, this independence must be reflected in the graph structure. However, this implies the absence of a directed path from $a^i_t$ to $r_t$, contradicting the existence of the edge. Thus, $a^i_t \not \! \perp \!\!\! \perp r_t | { a_t \setminus a^i_t, s_t }$.

($\Leftarrow$) Now, suppose $a^i_t \not \! \perp \!\!\! \perp r_t | { a_t \setminus a^i_t, s_t }$. We prove the existence of an edge from $a^i_t$ to $r_t$ by contradiction. Assume no such edge exists. By the Markov assumption, the absence of this edge implies $a^i_t  \! \perp \!\!\! \perp r_t | { a_t \setminus a^i_t, s_t }$, contradicting our initial supposition. Therefore, an edge from $a^i_t$ to $r_t$ must exist.
Thus, we have shown that an edge from $a^i_t$ to $r_t$ exists if and only if $a^i_t \not \! \perp \!\!\! \perp r_t | { a_t \setminus a^i_t, s_t }$, completing the proof.
\paragraph{Proposition 2} 
\textit{Under the assumptions that the causal graph is Markov and faithful to the observations, there exists an edge from $s^i_t \to r_t$ if and only if $s^i_t \not \! \perp \!\!\! \perp r_t | \{a_t, s_t  \setminus r_t\}$.}

The proof of Proposition 2 follows a similar line of reasoning as that of Proposition 1.

\textbf{Theorem 1} \textit{Based on above $5$ assumptions and $2$ propositions, suppose $s_t, a_t, s_t$ follow the factored MDP reward function Eq.~\ref{eq:gen}, the causal matrices $M^{s \to r}$ and $M^{a \to r}$ are identifiable}.

\section{Extended Related Work}
\label{sec:appendix_related_work}
We categorize existing causal RL approaches based on problem domains and task types, providing a systematic analysis of how different methods explore causal relationships between states, actions, and rewards, as illustrated in Table~\ref{tab:causal_rl}.

In the single-task learning domain, methods such as ACE~\citep{ji2024ace} and IFactor~\citep{liu2024learning} have shown success in learning policies for manipulation and locomotion tasks. However, both approaches are limited by focusing on a single reward-guided causal relationship.
Regarding generalization, AdaRL~\citep{huangadarl} effectively leverages both state-reward and action-reward causal relationships. 
However, AdaRL focuses primarily on applying causal inference to address generalization challenges in locomotion tasks. 
Its application is limited to locomotion tasks, leaving more complex manipulation tasks unaddressed. Since our work focuses on the single-task problem domain, we do not provide a direct comparison with AdaRL. 
Conversely, CBM~\citep{wang2024building} considers the causal relationship between states and rewards but overlooks the causal link between actions and rewards.
In the problem domain of counterfactual data augmentation, current causal RL methods~\citep{urpicausal,pitis2020counterfactual,pitis2022mocoda} have not yet explored the inference and utilization of both causal relationships.

In summary, current research on reward-guided causal discovery remains incomplete and lacks validation across a broader spectrum of tasks. This gap underscores the need for more comprehensive investigation and application in the field of causal reinforcement learning. 

\textcolor{black}{
\subsection{Extended Discussion on object-centric RL and 3D world models}
\label{ocrl}The main similarity lies between  our framework and object-centric RL is both are learning and using factored MDPs~\citep{kearns1999efficient}, but they differ in granularity: our framework operates at the component level (e.g., raw state variables), whereas object-centric RL factors states based on objects.}

\textcolor{black}{Although our work is orthogonal to object-centric RL, we believe certain elements of object-centric RL could complement our framework in specific applications, particularly in real-world robotic manipulation tasks. Potential future work include:
\begin{itemize}
    \item \textbf{Using object-centric representation as input}: Object-centric models can help identify object-factored variables, such as object attributes, geometry, and physical states, which are useful for planning~\citep{jiang2019scalor,lin2020improving,kossen2019structured, mambelli2022compositional,feng2023learning, choi2024unsupervised, zadaianchuk2022self, park2021object, zadaianchuk2020self, yuan2022sornet, li2020towards, mitash2024scaling, haramati2024entity, li2024manipllm}. In this case, states are factored as objects, and we can learn causal graphs over these variables. This is useful in robotic environments involving numerous objects. We will leave this as a future work for adapting our current framework to the applications of the object-centric robotic task. 
    \item \textbf{Learning more compact factored object representations with our framework}: Our structure learning approach could benefit object-centric RL by disentangling the internal representations of individual objects to the reward-relevant and reward-irrelevant groups by learning the causal structures. This can enhance the compactness and interpretability of object-centric representations.
    \item \textbf{Using object-aware 3D world models for applications}: In 3D environments, object-aware 3D world models~\citep{li2024objectaware} can provide essential representations of objects. Our framework could then build causal structures on top of these factored 3D-object representations.
\end{itemize}
}

\textcolor{black}{While these directions are promising and could advance the applicability of our framework in certain domains, they are outside the primary focus of this work. We plan to explore these ideas as part of future work.}

\begin{table}[t]
\centering
\renewcommand{\arraystretch}{1.1}
\setlength{\tabcolsep}{1pt}
\caption{Categorization of different causal RL methods with two different causal relationship of state-to-reward (state-reward) and action-to-reward (action-reward).}
\label{tab:causal_rl}
\begin{tabular}{ccccc}
\toprule
\multirow{2}{*}{\textbf{Problem domain}} & \multirow{2}{*}{\textbf{Task type}} & \multirow{2}{*}{\textbf{Method}} & \multicolumn{2}{c}{\textbf{Causal relationship}} \\ \cline{4-5} 
                         &                            &                         & \textbf{state-reward}       & \textbf{action-reward}      \\ \toprule
\textbf{Single-task}              & manipulation; locomotion    & ACE~\citep{ji2024ace}                     & \xmark                  & \cmark                  \\
                         & manipulation; locomotion    & IFactor~\citep{liu2024learning}                 & \cmark                 & \xmark                   \\ 
                          & manipulation    & CAI~\citep{seitzer2021causal}                 & \xmark                 & \xmark                   \\\hline
\textbf{Generalization}           & manipulation               & CDL~\citep{wang2022causal}                     & \xmark                   & \xmark                   \\
                         & locomotion                 & AdaRL~\citep{huangadarl}                   & \cmark                  & \cmark                  \\ 
                          & manipulation; locomotion                 & CBM~\citep{wang2024building}                   & \cmark                  & \xmark                  \\ \hline
\textbf{Augmentation}             & manipulation               & CAIAC~\citep{urpicausal}                   & \xmark                   & \xmark                   \\
             & manipulation               & CoDA~\citep{pitis2020counterfactual}                   & \xmark                   & \xmark                   \\  & manipulation               & MoCoDA~\citep{pitis2022mocoda}                   & \xmark                   & \xmark                   \\ \bottomrule
\end{tabular}
\end{table}

\section{Details on Experimental Design and Results}
\label{Details on Experimental Design and Results}

\subsection{Experimental setup}
\label{sec:experimental_setup_appendix}
We present the detailed hyperparameter settings of the proposed method $\texttt{\textbf{CIP}}$ across all $5$ environments in Table~\ref{tab:hps}. Additionally, the Q-value and V-value networks are used MLP with 512 hidden size. And the policy network is the Gaussian MLP with 512 hidden size. Moreover, we set the target update interval of $2$. For fair comparison, the hyperparameters of the baseline methods (SAC~\citep{haarnoja2018soft}, BAC~\citep{ji2023seizing}, ACE~\citep{ji2024ace}) follow the same settings in the experiments.

For pixel-based DMControl environments, we employ IFactor~\citep{liu2024learning} to encode latent states and integrate the \texttt{\textbf{CIP}} framework for policy learning. We utilize the $s_{t}^{\bar{r}}$ state features in IFactor as uncontrollable states unrelated to rewards to execute counterfactual data augmentation. Furthermore, for simplicity, we maximize the mutual information between future states and actions to facilitate empowerment. All parameter settings in these three tasks adhere to those specified in IFactor. Additionally, We use the same background video for the comparison. 

\begin{table}[h]
 \centering
\renewcommand{\arraystretch}{1.2}
\setlength{\tabcolsep}{4pt}
\caption{Hyperparameter settings of $\texttt{\textbf{CIP}}$ in $5$ environments}
\label{tab:hps}
\begin{tabular}{cccccc}
\hline
\multirow{2}{*}{\textbf{Hyperparameter}} & \multicolumn{5}{c}{\textbf{Environment}}                        \\ \cline{2-6} 
                                & \textbf{Meta-World} & \textbf{Sparse} & \textbf{MuJoCo} & \textbf{DMControl} & \textbf{Adroit Hand} \\ \hline
\textbf{batch size }                     & 512        & 512    & 256    & 512       & 256         \\
\textbf{hidden size  }                   & 1024       & 1024   & 256    & 1024      & 256         \\
\textbf{Q-value network hidden size}                       & \multicolumn{5}{c}{512}                            \\
\textbf{V-value network hidden size}                       & \multicolumn{5}{c}{512}                            \\
\textbf{policy network hidden size}                       & \multicolumn{5}{c}{512}                            \\
\textbf{learning step}                       & \multicolumn{5}{c}{1000000}                            \\
\textbf{replay size}                     & \multicolumn{5}{c}{1000000}                            \\
\textbf{causal sample size}              & \multicolumn{5}{c}{10000}                              \\
\textbf{gamma}                           & \multicolumn{5}{c}{0.99}                               \\
\textbf{learning rate}                   & \multicolumn{5}{c}{0.0003}                             \\ 
\textbf{update interval}                   & \multicolumn{5}{c}{2}                             \\
\hline
\end{tabular}
\end{table}

\begin{figure}[t]
    \centering
    \includegraphics[width=1\textwidth]{figs/meta/meta_1_success.pdf}
    \includegraphics[width=1\textwidth]{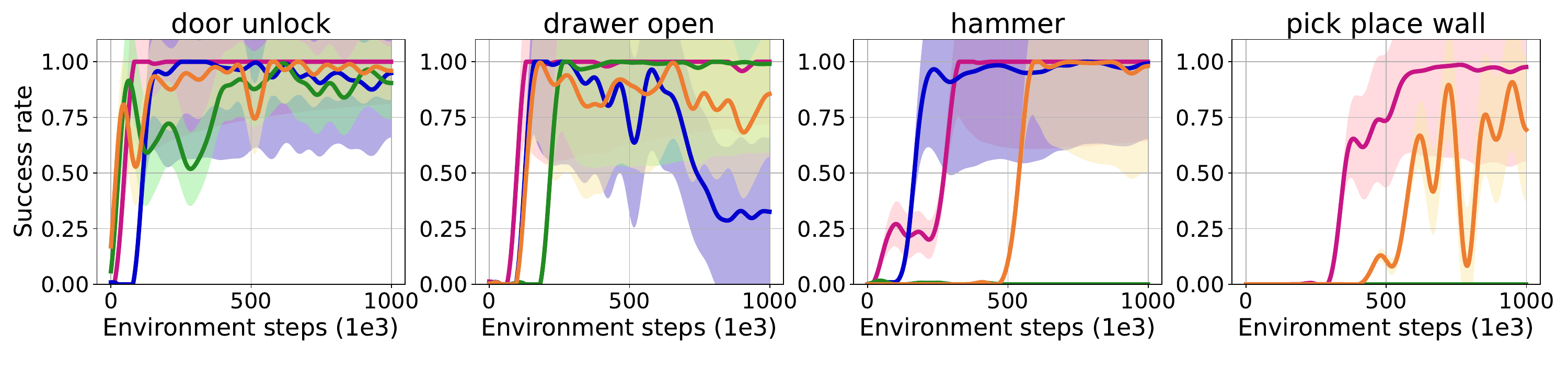}
    \includegraphics[width=1\textwidth]{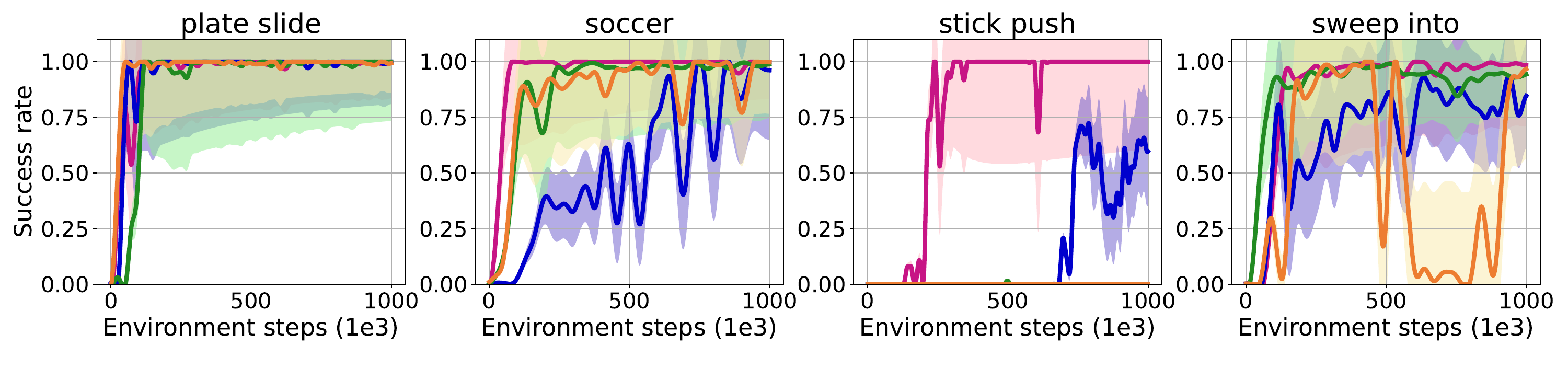}
    \includegraphics[width=1\textwidth]{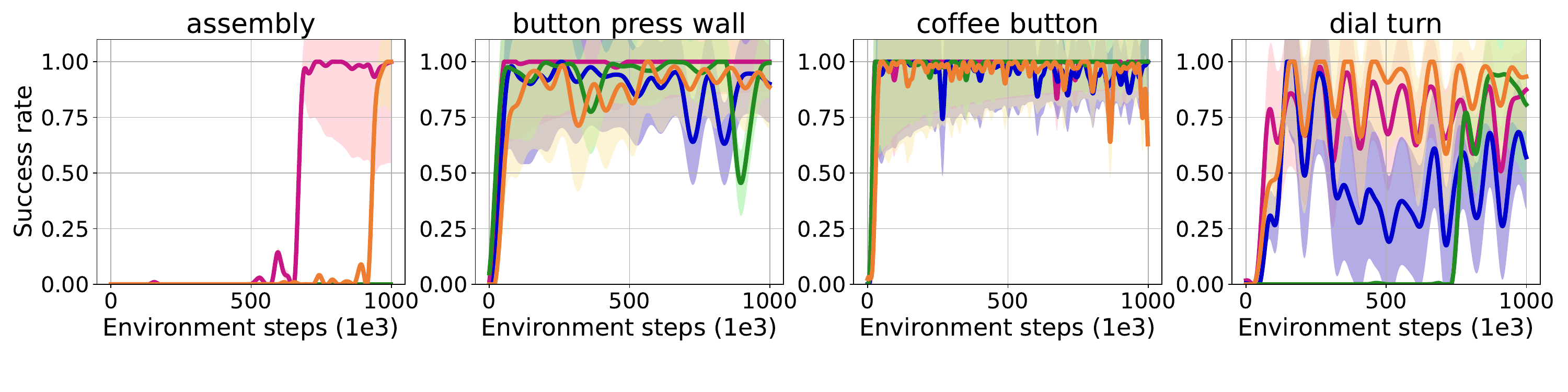}
    \includegraphics[width=0.3\textwidth]{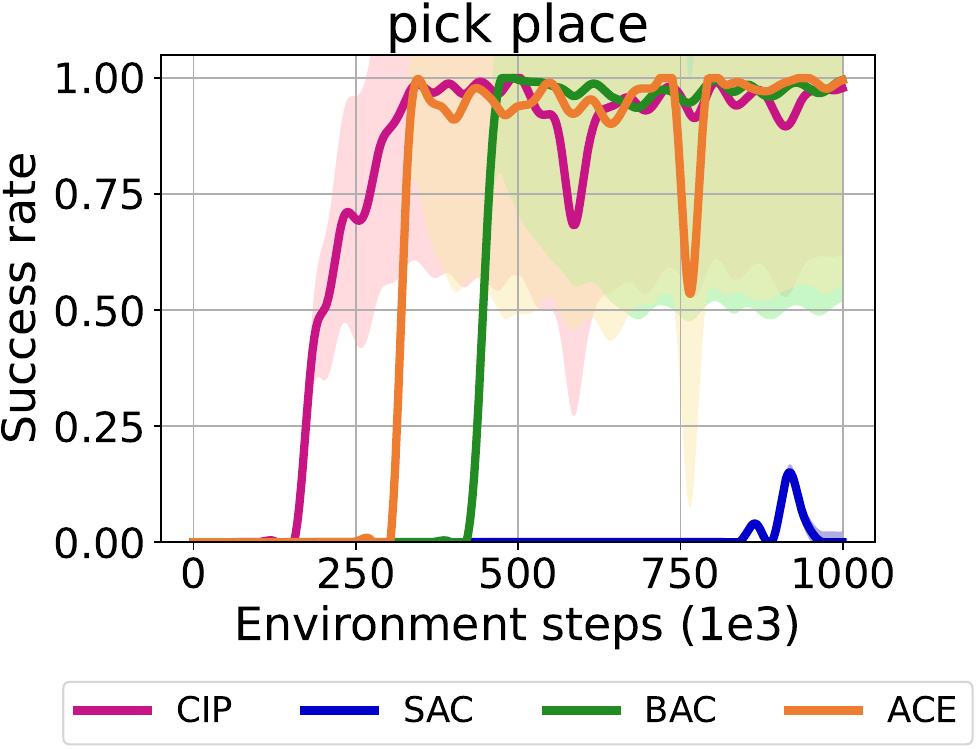} 
    \caption{Experimental results across $17$ manipulation skill learning tasks in Meta-World.}
    \label{fig:appendix_manipulation}
\end{figure}

\begin{figure}[h]
    \centering
    \includegraphics[width=1\linewidth]{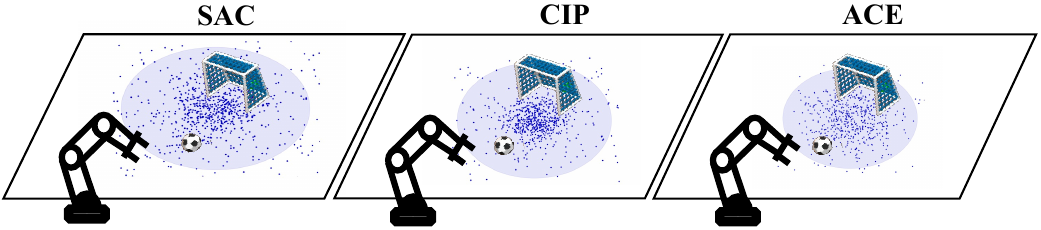}
    \caption{Visualization of the trajectories in soccer task.}
    \label{fig:aug}
    \vspace{-3mm}
\end{figure}

\subsection{Full Results}
\label{sec:full_results_appendix}

\subsubsection{Effectiveness in robot arm manipulation}
\begin{figure}[t]
    \centering
    \includegraphics[width=1\textwidth]{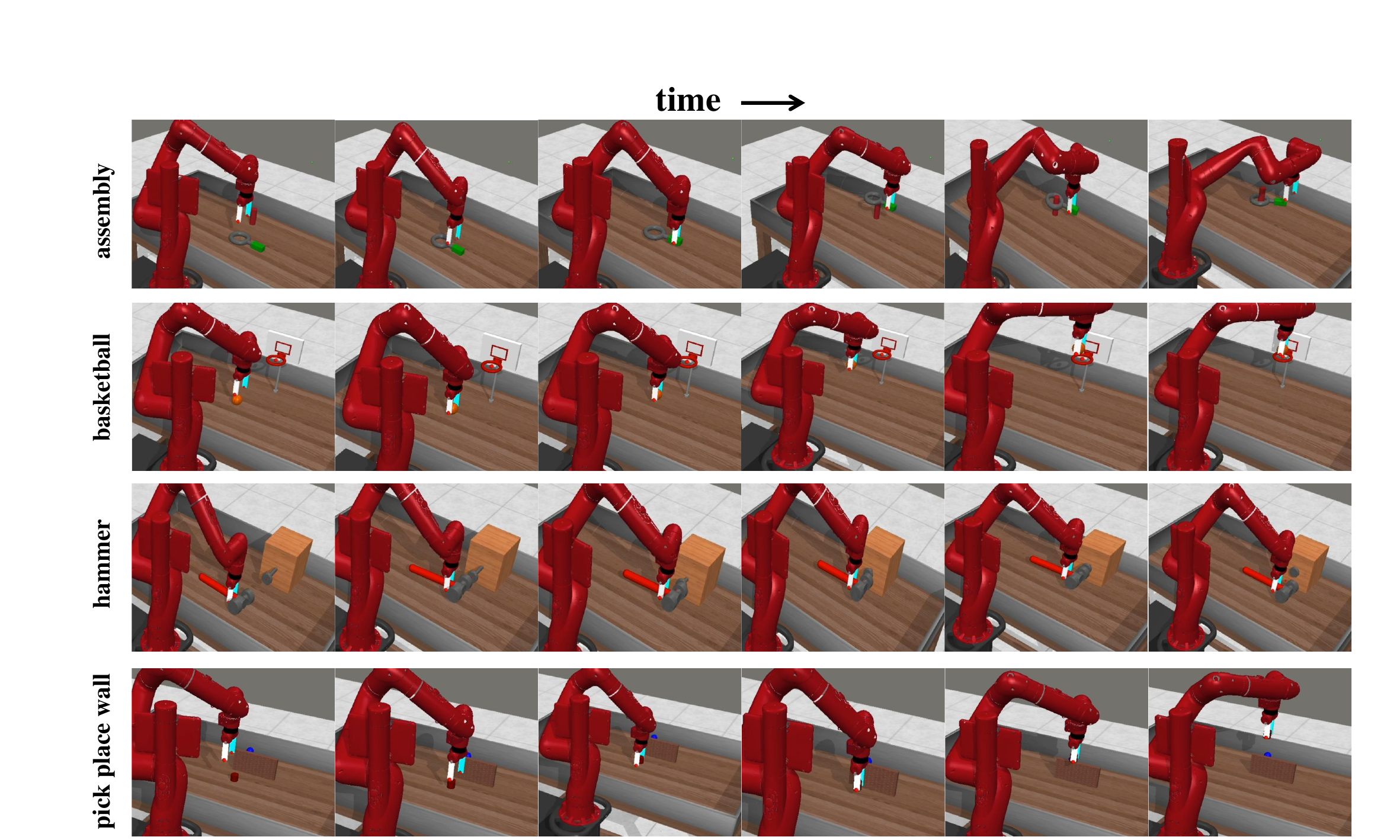}
    \caption{Visualization trajectories of $4$ manipulation skill learning tasks in Meta-World environment.}
    \label{fig:appendix_meta_vis}
\end{figure}

\begin{figure}[t]
    \centering
    \includegraphics[width=1\textwidth]{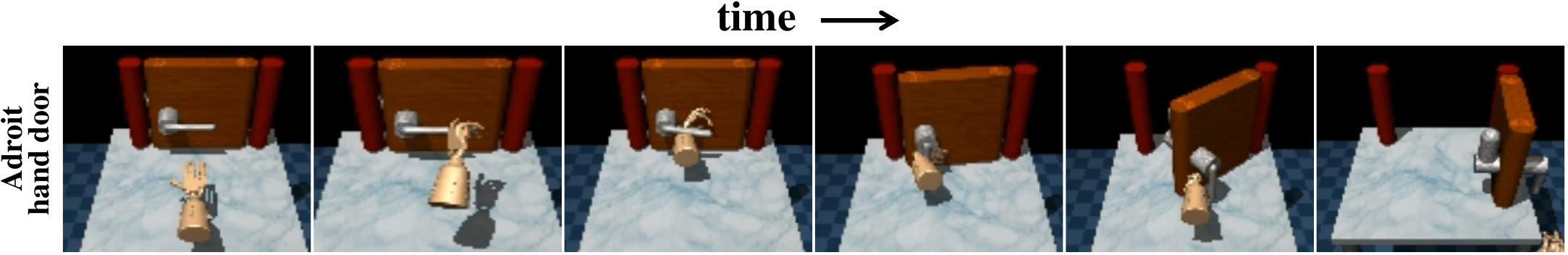}
    \caption{Visualization trajectory of Adroit Hand door open task.}
    \label{fig:appendix_adroit_vis}
\end{figure}

Figure~\ref{fig:appendix_manipulation} presents the learning curves for all 17 manipulation skill tasks within the Meta-World environment. The \(\texttt{\textbf{CIP}}\) framework demonstrates superior learning outcomes and efficiency compared to the three baseline methods, despite exhibiting minor instabilities in the basketball and dial-turn tasks. Notably, \(\texttt{\textbf{CIP}}\) achieves a 100\% success rate in more complex tasks, such as pick-place-wall and assembly. 
The visualization results presented in Figures~\ref{fig:appendix_meta_vis} and \ref{fig:appendix_adroit_vis} further demonstrate \(\texttt{\textbf{CIP}}\)'s ability to effectively and efficiently complete tasks, even in high-dimensional action spaces such as the Adroit Hand environment. 

In the hammer task, \(\texttt{\textbf{CIP}}\) allows the robot arm to execute reach and pick actions with precision, enabling it to accurately identify the nail's position and successfully perform the hammering action.
In the Adroit Hand door task, \(\texttt{\textbf{CIP}}\) effectively controls the complex joints to grasp the doorknob and applies the appropriate force to twist it, thereby opening the door. 

These findings affirm the effectiveness of \(\texttt{\textbf{CIP}}\) in robot arm manipulation skill learning, highlighting its capacity to enhance sample efficiency while mitigating the risks associated with blind exploration.

\vspace{-3mm}
\paragraph{Visualization.} 
We employ trajectory visualization to comparatively validate the efficacy of our method. As depicted in Figure~\ref{fig:aug}, the light-shaded regions delineate the policy exploration space, while the point clustering area indicates the area of frequent interaction. Our analysis reveals that $\texttt{\textbf{CIP}}$, leveraging counterfactual data augmentation, achieves substantially broader exploration compared to ACE and SAC. Concurrently, the causal information prioritization framework facilitates more focused execution in critical state regions. These visual findings provide robust empirical support for the effectiveness of our proposed augmentation framework. 

\subsubsection{Effectiveness in spare reward settings}

\begin{figure}[t]
    \centering
    \includegraphics[width=0.9\textwidth]{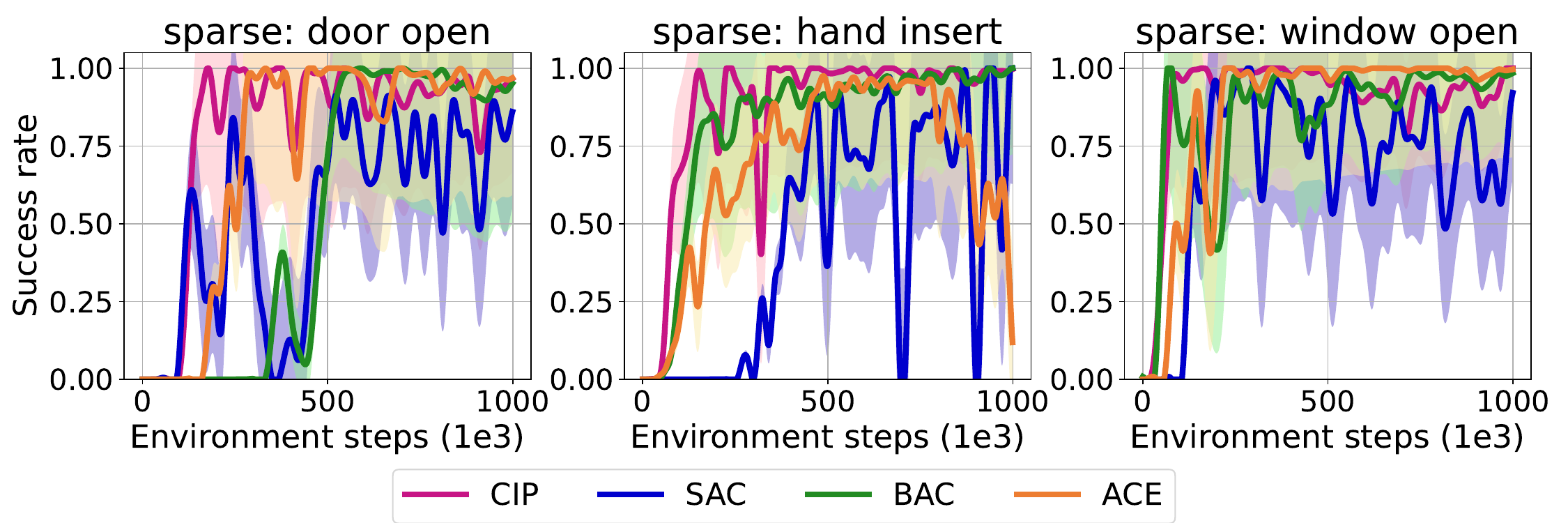}
    \caption{Experimental results across $3$ manipulation skill learning tasks in sparse reward settings of Meta-World environment.}
    \label{fig:appendix_meta_sparse}
\end{figure}

\begin{figure}[t]
    \centering
    \includegraphics[width=1\textwidth]{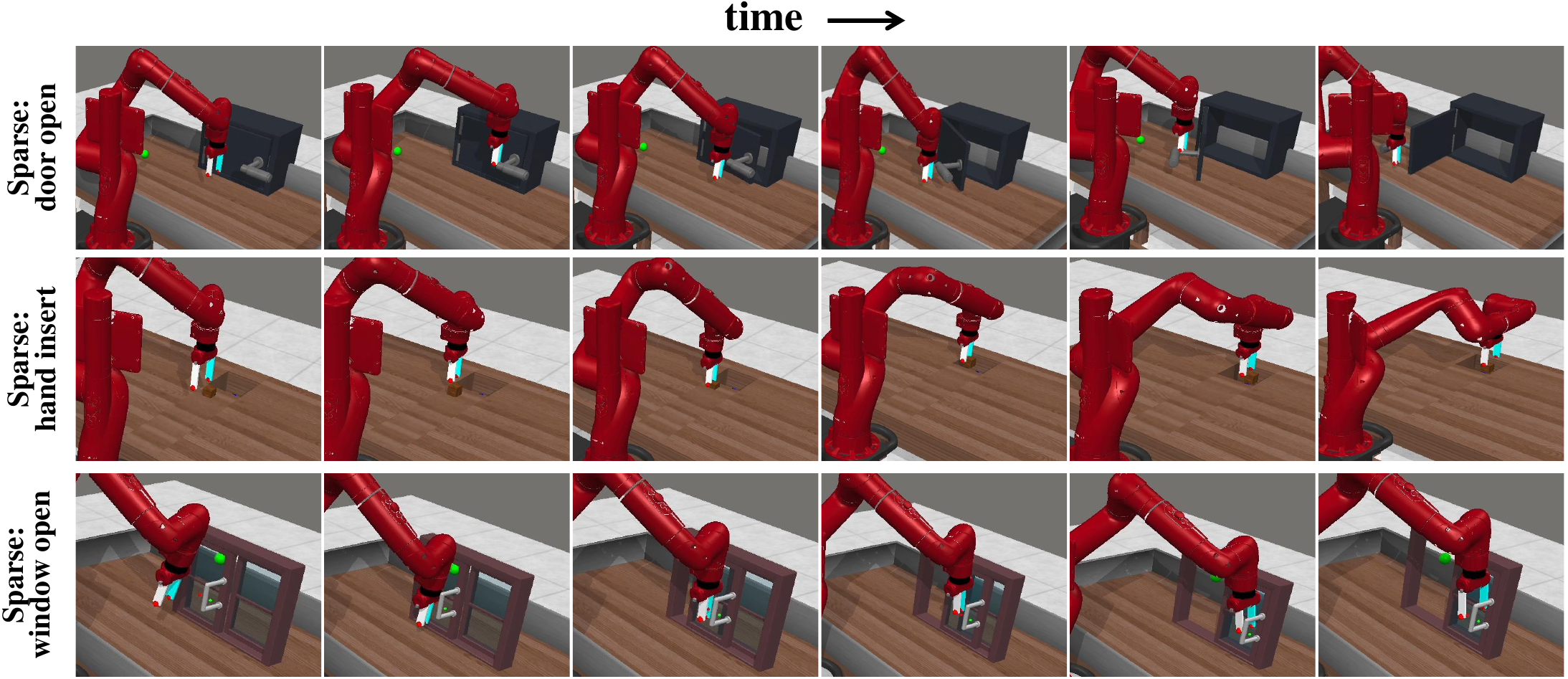}
    \caption{Visualization trajectories of $3$ manipulation skill learning tasks in sparse reward settings of Meta-World environment.}
    \label{fig:appendix_meta_sparse_vis}
\end{figure}

Figure~\ref{fig:appendix_meta_sparse} presents the learning curves for all three sparse reward setting tasks within the Meta-World environment, while Figure~\ref{fig:appendix_meta_sparse_vis} showcases their corresponding visualization trajectories. These findings reveal that \(\texttt{\textbf{CIP}}\) not only achieves superior learning efficiency but also adeptly executes critical actions necessary for task completion, such as opening the door and window and maneuvering the node to the target place.

These results substantiate the effectiveness of \(\texttt{\textbf{CIP}}\) in sparse reward scenarios. The counterfactual data augmentation process prioritizes salient state information, effectively filtering out irrelevant factors that could hinder learning. Meanwhile, causal action empowerment enhances policy controllability by focusing on actions that are causally linked to desired outcomes. This dual approach not only accelerates the learning process but also fosters a more robust policy capable of navigating the complexities inherent in sparse reward settings. Overall, these findings underscore \(\texttt{\textbf{CIP}}\)'s potential to significantly improve performance in challenging environments characterized by limited feedback.

\subsubsection{Effectiveness in locomotion}
\begin{figure}[t]
    \centering
    \includegraphics[width=1\textwidth]{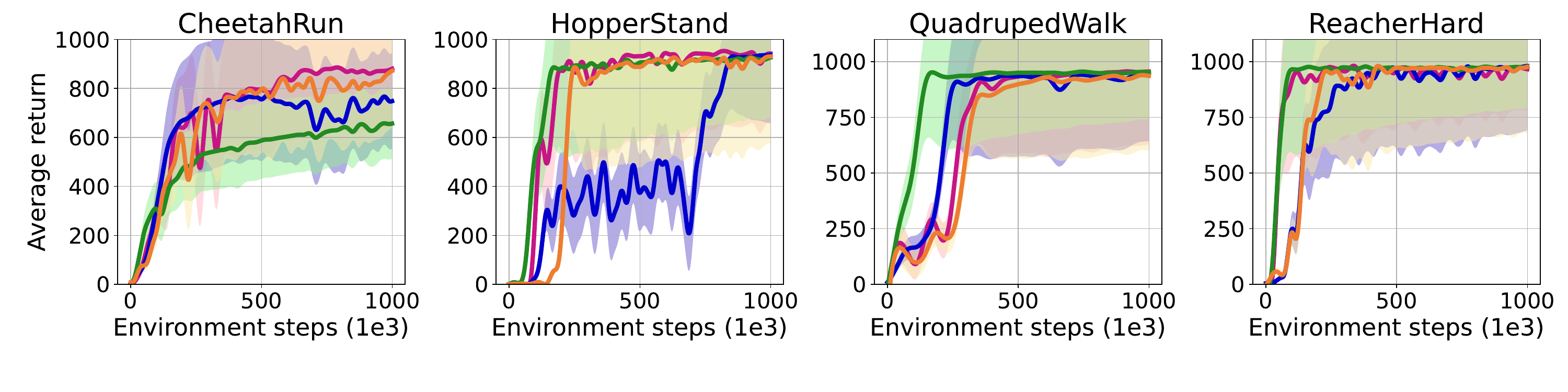}
    \includegraphics[width=1\textwidth]{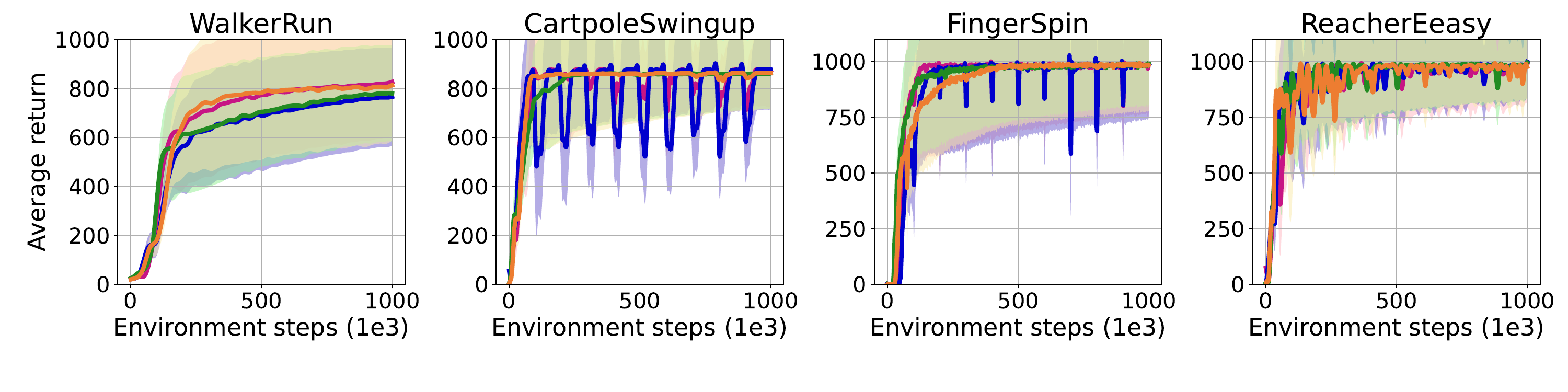}
    \includegraphics[width=0.75\textwidth]{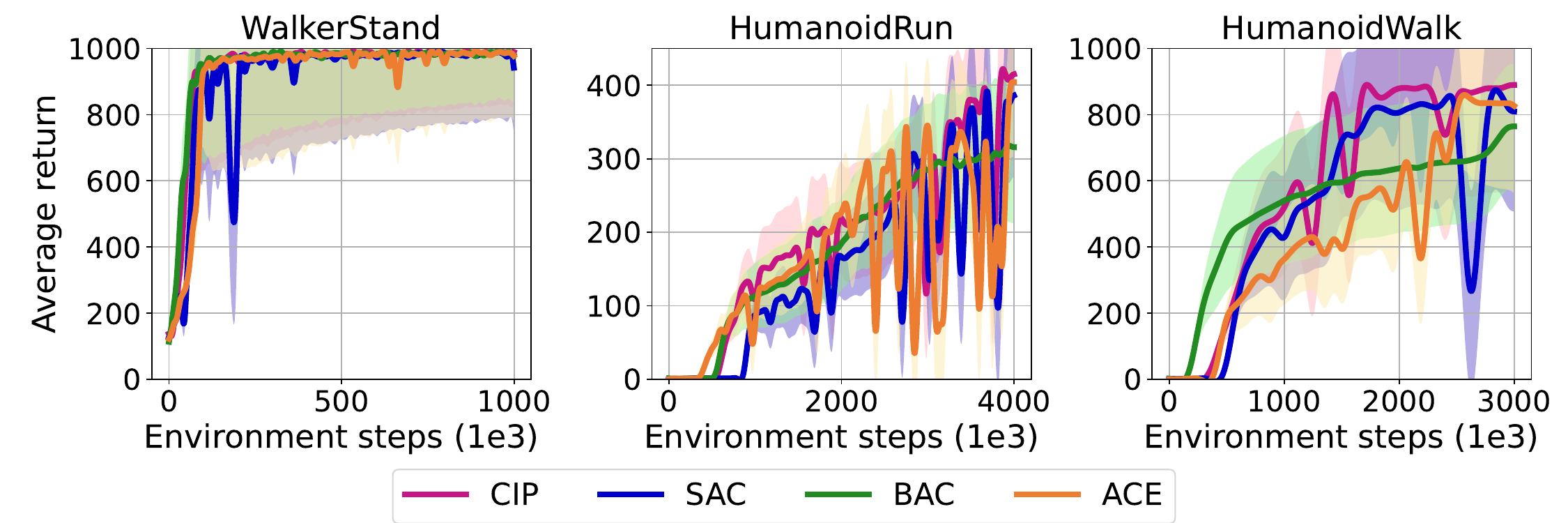}
    \caption{Experimental results across $11$ locomotion tasks in DMControl environment.}
    \label{fig:appendix_locomotion}
\end{figure}

\begin{figure}[t]
    \centering
    \includegraphics[width=1\textwidth]{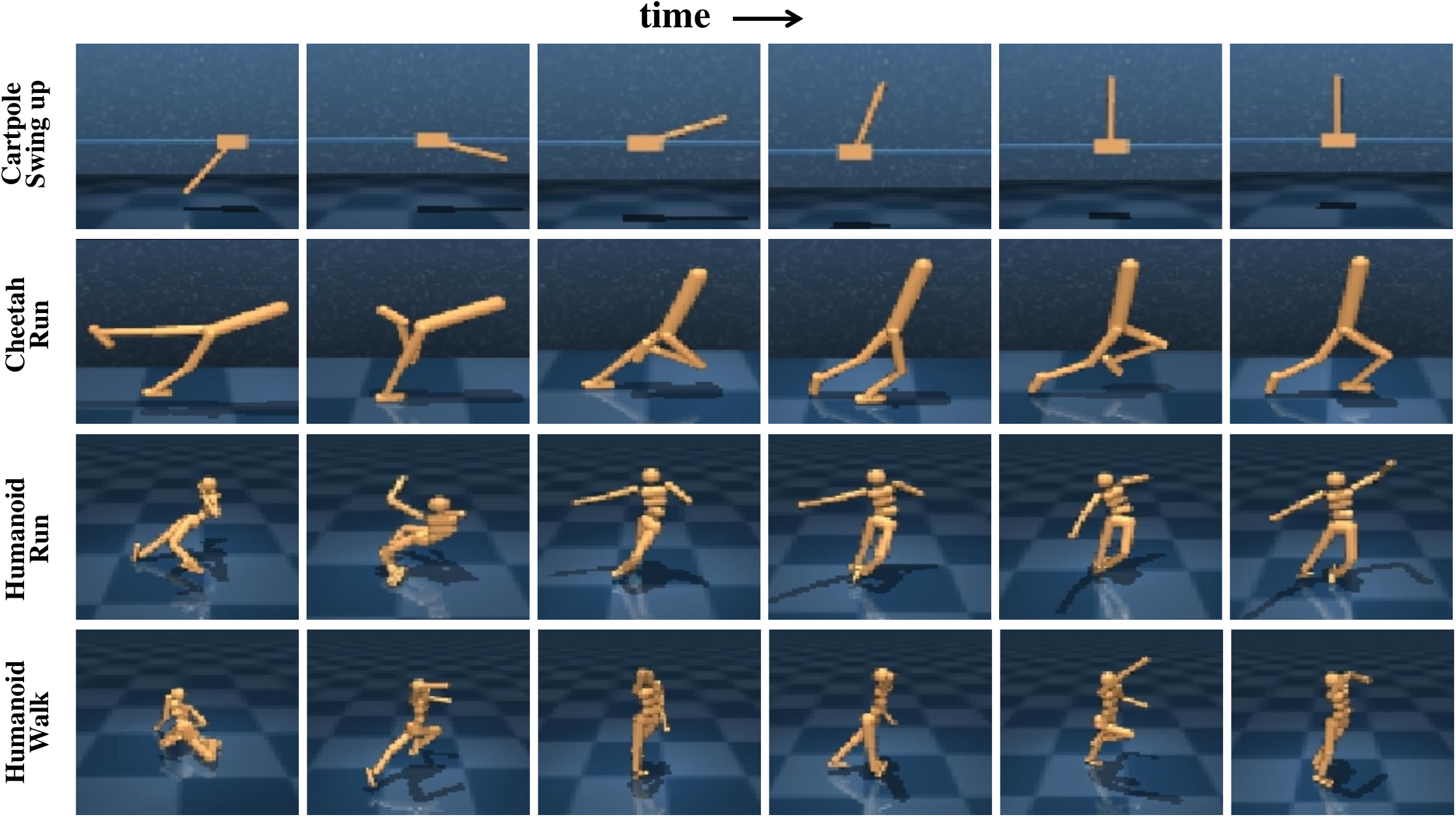}
    \caption{Visualization trajectories of $4$ locomotion tasks in DMControl environment.}
    \label{fig:appendix_loco_vis}
\end{figure}

We further evaluate \(\texttt{\textbf{CIP}}\) in $15$ locomotion tasks in DMControl and MuJoCo environments. Figure~\ref{fig:appendix_locomotion} presents the learning curves, while Figure~\ref{fig:appendix_loco_vis} showcases the corresponding visualization trajectories in $4$ specific tasks. 
A comprehensive analysis indicates that \(\texttt{\textbf{CIP}}\) achieves faster learning efficiency and greater stability compared to ACE and SAC, while demonstrating comparable policy learning performance to BAC, which is known for its proficiency in control tasks. The visualization results reveal that \(\texttt{\textbf{CIP}}\) effectively executes running and walking actions in complex humanoid scenarios.

These findings collectively underscore the efficacy of \(\texttt{\textbf{CIP}}\) in locomotion tasks, highlighting its potential to advance the state-of-the-art in reinforcement learning for intricate motor control problems. The method's success across varied environments suggests a robust framework that could generalize effectively to other challenging domains within robotics and control systems, paving the way for future research and applications in these areas.

\clearpage

\subsubsection{Effectiveness in pixel-based tasks}
\label{sec:vis_pixel}
To further validate the effectiveness of our proposed framework in pixel-based environments, we evaluated $\texttt{\textbf{CIP}}$ on three DMControl pixel-based tasks. We leverage IFactor for latent state processing and differentiation of uncontrollable state features to execute counterfactual data augmentation, alongside maximizing the mutual information between future states and actions for empowerment. 

\begin{figure}[t]
    \centering
    \includegraphics[width=1\textwidth]{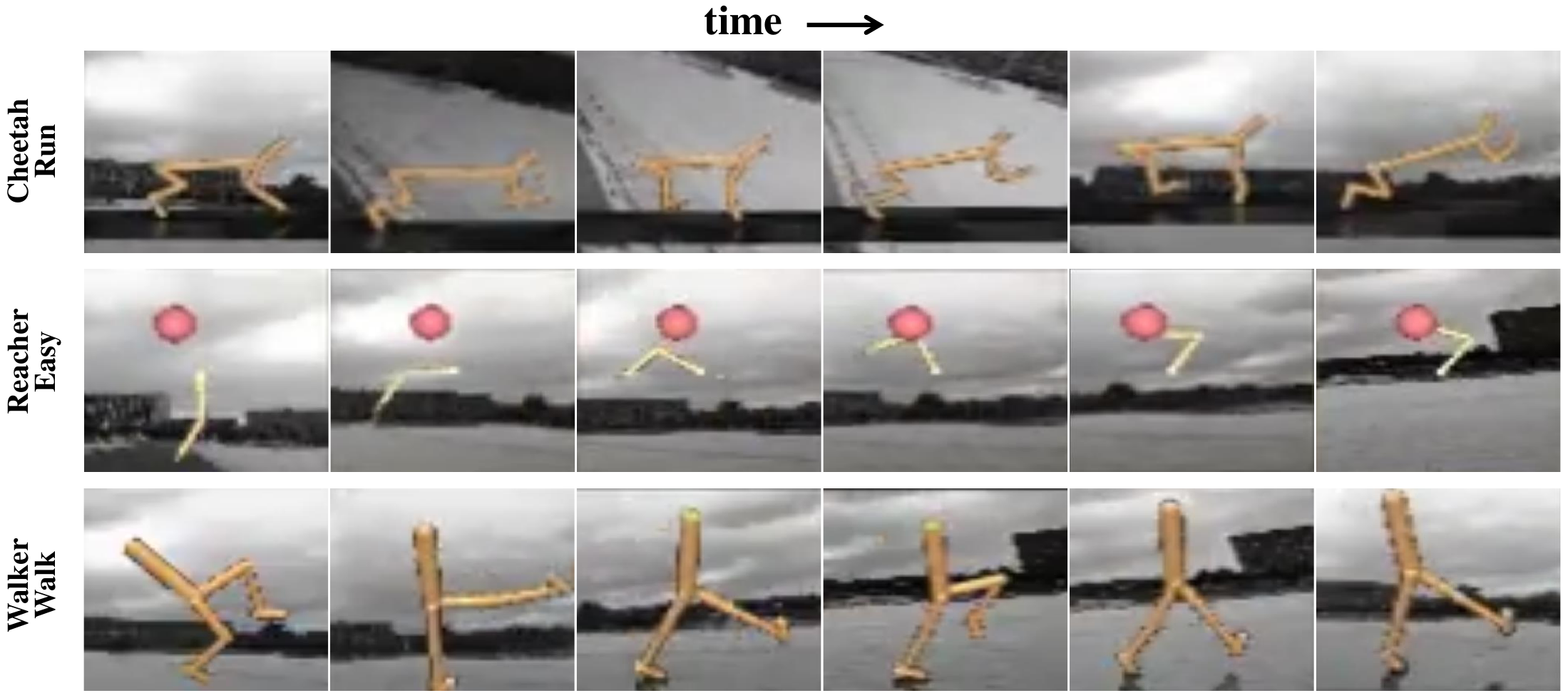}
    \caption{Visualization trajectories in $3$ pixel-based locomotion tasks of DMControl environment with video backgrounds as distractors.}
    \label{fig:appendix_pixel_vis}
\end{figure}

Figure~\ref{fig:appendix_pixel} presents the learning curves, while Figure~\ref{fig:appendix_pixel_vis} shows the visualization trajectories. The proposed framework exhibits enhanced policy learning performance and effectively mitigates interference from background video, facilitating efficient locomotion. These findings reinforce the effectiveness and extensibility of our causal information prioritization framework, highlighting its potential to improve learning in complex, pixel-based environments. 

\begin{figure}[t]
    \centering
    \includegraphics[width=0.24\textwidth]{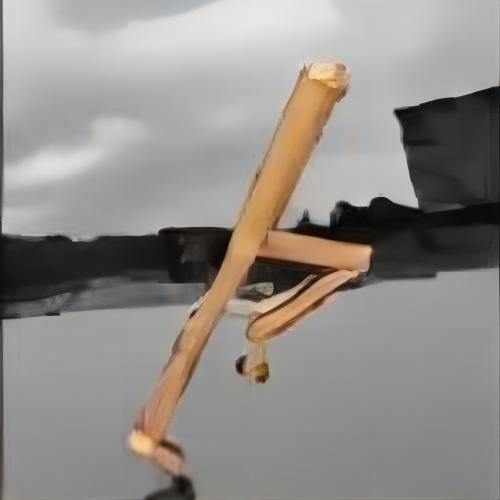}
    \includegraphics[width=0.24\textwidth]{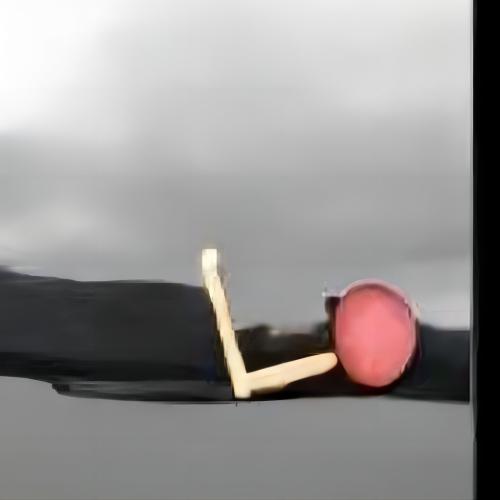}
    \includegraphics[width=0.24\textwidth]{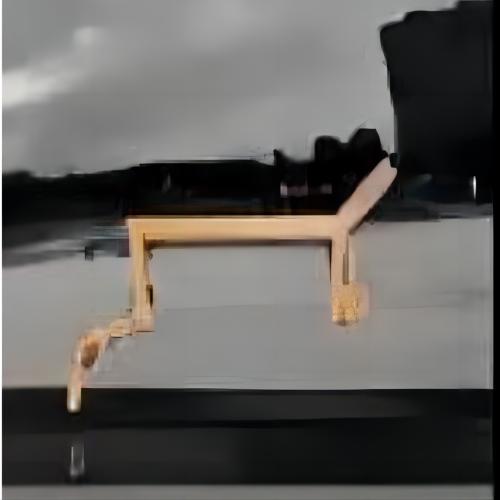}
    \includegraphics[width=0.24\textwidth]{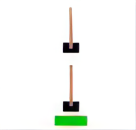}
    \caption{The DMControl environment of $3$ pxiel-based tasks (Walker Walk, Cheetah Run, Reacher Easy) and $1$ task in Cartpole environment~\citep{liu2024learning}.}
    \label{fig:appendix_pixel_env}
\end{figure}

\subsection{Property Analysis}
\label{sec:pro_res_appendix}

\subsubsection{Analysis for replacing counterfactual data augmentation}
\label{sec:appendix_cda_replace}

In $\texttt{\textbf{CIP}}$, we exploit the causal relationship between states and rewards to perform counterfactual data augmentation on irrelevant state features, thus prioritizing critical state information. We compare this approach with an alternative method: masking irrelevant state features to achieve state abstraction for subsequent causal action empowerment and policy learning. To evaluate the efficacy of both approaches, we conduct experiments with $\texttt{\textbf{CIP}}$ with counterfactual data augmentation ($\texttt{\textbf{CIP}}$ w/i Cda) and $\texttt{\textbf{CIP}}$ with causally-informed states ($\texttt{\textbf{CIP}}$ w/i Cs) across three distinct environments.

\begin{figure}[t]
    \centering
    \includegraphics[width=1\textwidth]{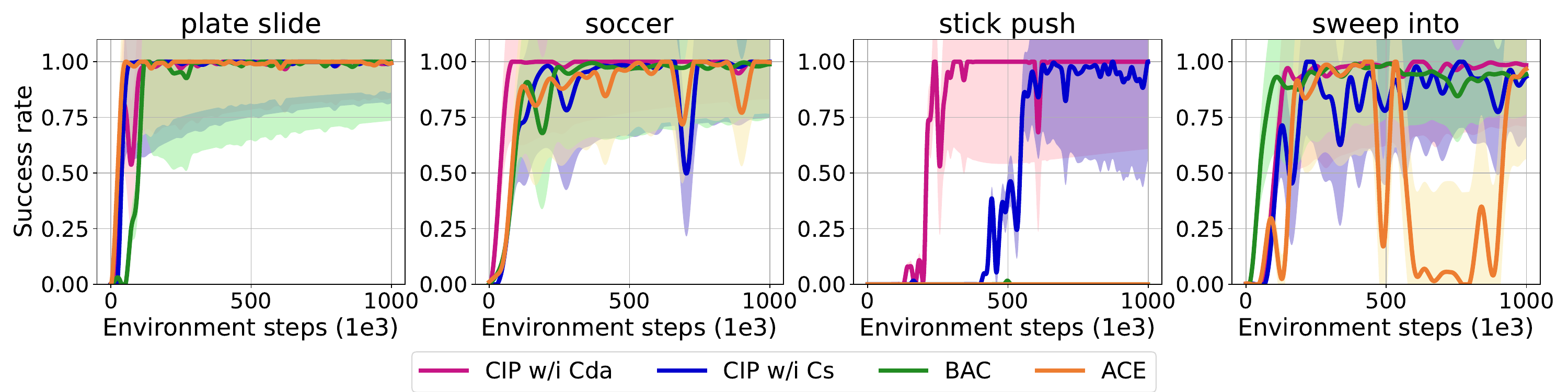}
    \caption{Experimental results in $4$ manipulation skill learning tasks of Meta-World environment. w/i stands for with.}
    \label{fig:cda_meta}
\end{figure}

\begin{figure}[t]
    \centering
    \includegraphics[width=0.9\textwidth]{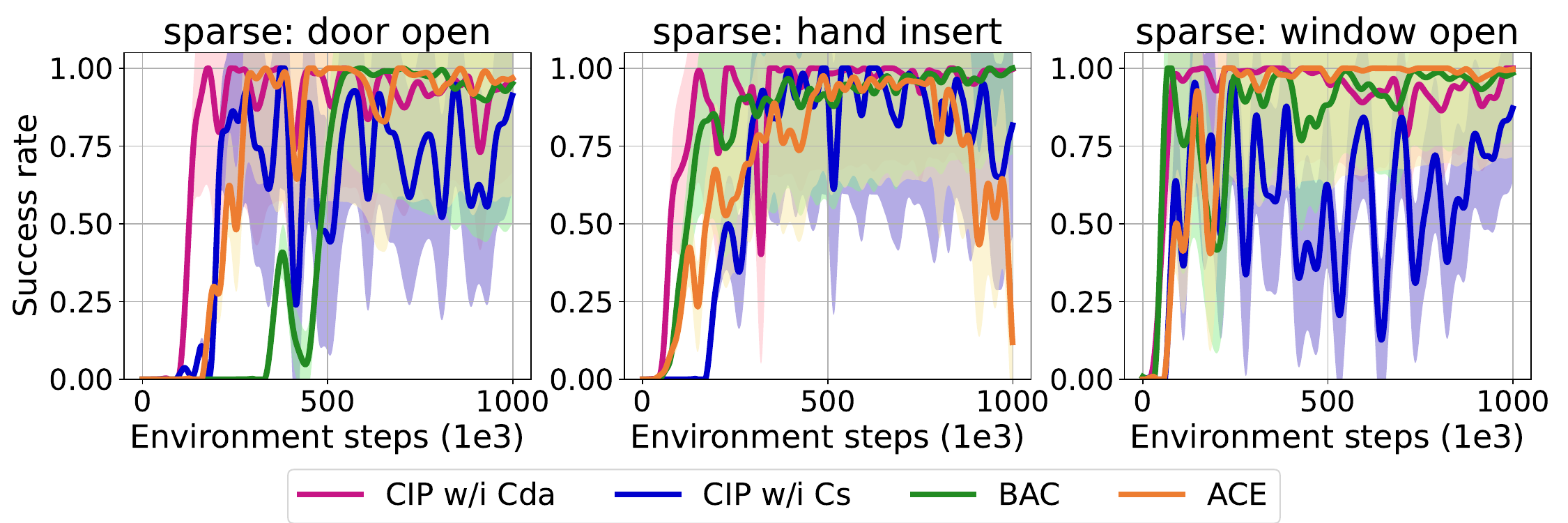}
    \caption{Experimental results in $3$ manipulation skill learning tasks of Meta-World environment with sparse reward settings.}
    \label{fig:cda_sparse}
\end{figure}

Figure \ref{fig:cda_meta} illustrates comparative results for four manipulation skill learning tasks in the Meta-World environment. Both $\texttt{\textbf{CIP}}$ variants achieve 100\% task success rates with high sample efficiency, validating their effectiveness. Notably, $\texttt{\textbf{CIP}}$ w/i Cda exhibits superior learning efficiency compared to $\texttt{\textbf{CIP}}$ w/i Cs, underscoring the value of our counterfactual data augmentation approach in enhancing training data without additional environmental interactions. 
In three sparse reward setting tasks (Figure \ref{fig:cda_sparse}), $\texttt{\textbf{CIP}}$ w/i Cda demonstrates superior policy performance. Further experiments across four locomotion environmetal tasks corroborate these findings, consistently favoring the counterfactual data augmentation approach.
These comprehensive experimental results strongly support the effectiveness and significance of incorporating counterfactual data augmentation in $\texttt{\textbf{CIP}}$, highlighting its potential to enhance reinforcement learning across diverse task domains.

\begin{figure}[t]
    \centering
    \includegraphics[width=1\textwidth]{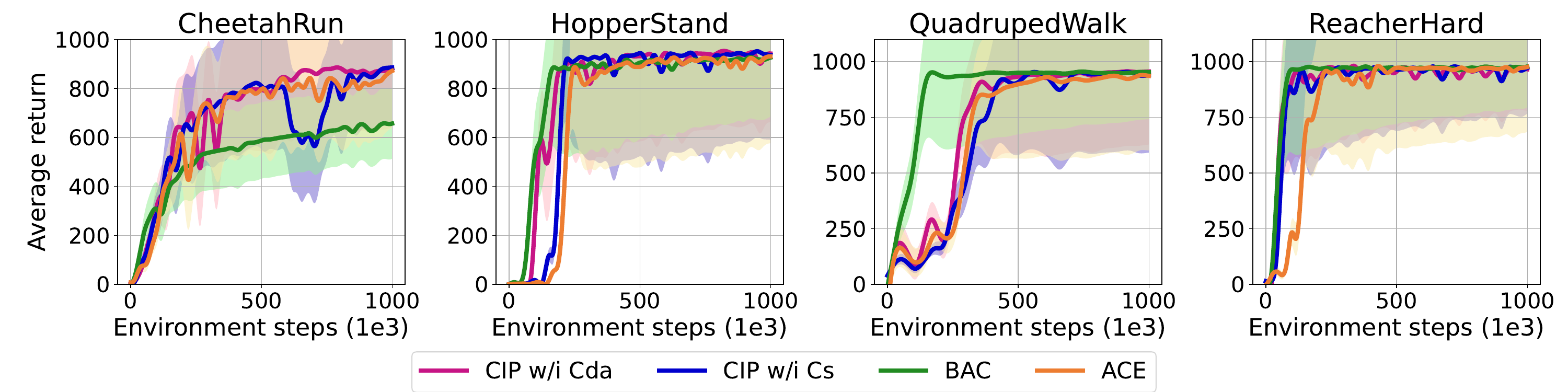}
    \caption{Experimental results in $4$ locomotion tasks of DMControl environment.}
    \label{fig:cda_dmc}
\end{figure}

\subsubsection{Extensive ablation study}
\paragraph{Robot arm manipulation}

\begin{figure}[t]
    \centering
    \includegraphics[width=1\textwidth]{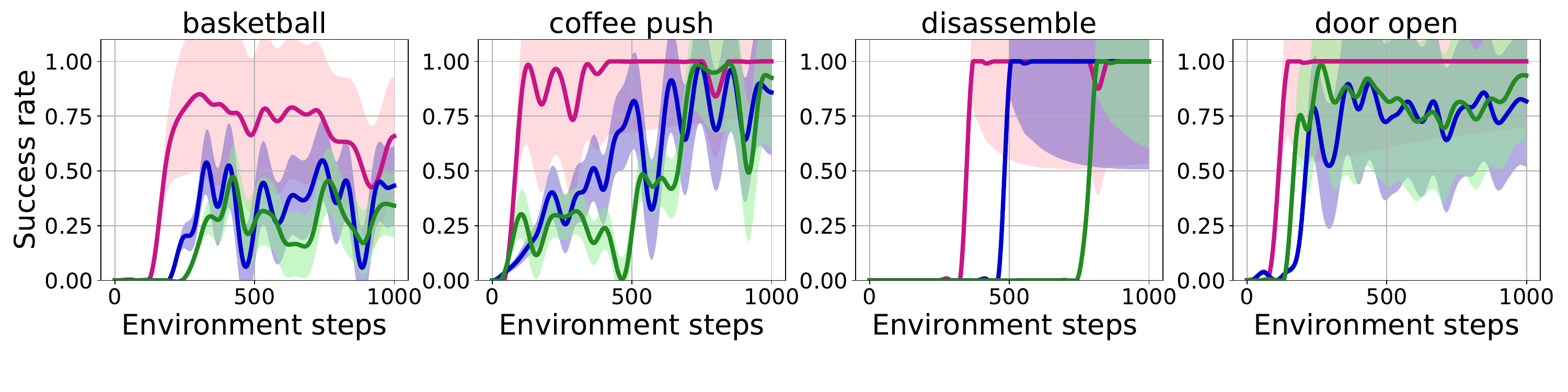}
    \includegraphics[width=1\textwidth]{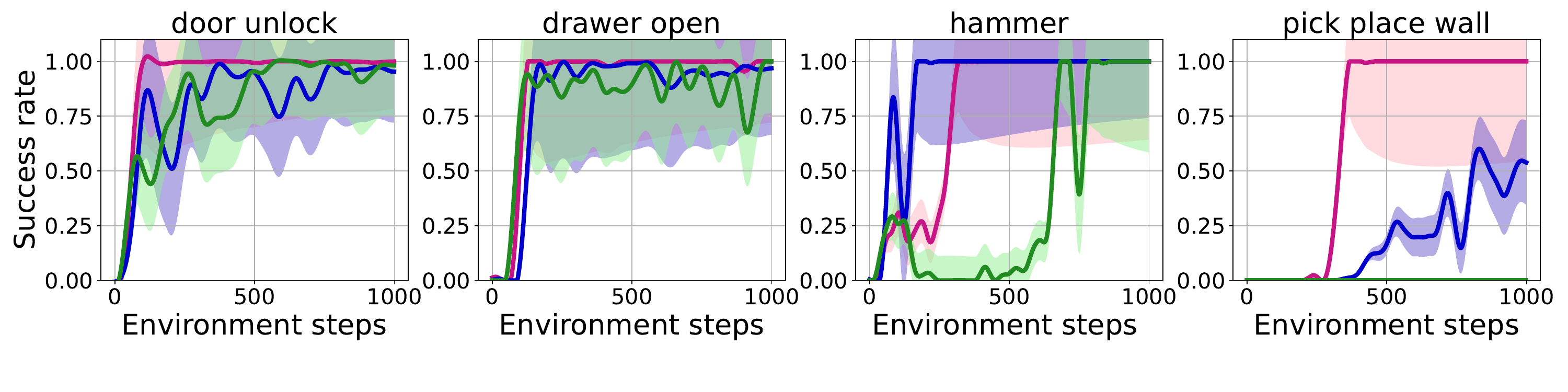}
    \includegraphics[width=1\textwidth]{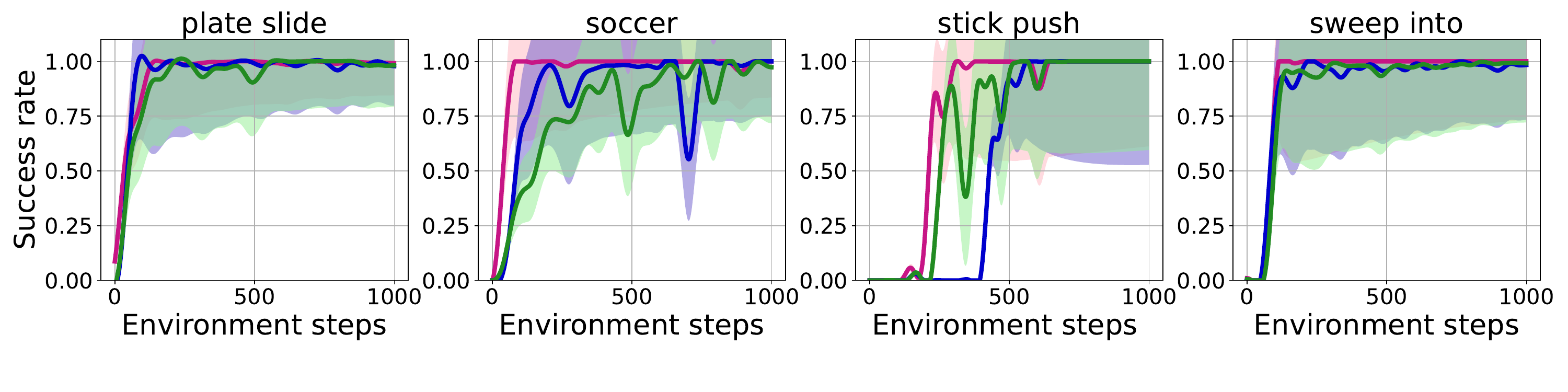}
    \includegraphics[width=1\textwidth]{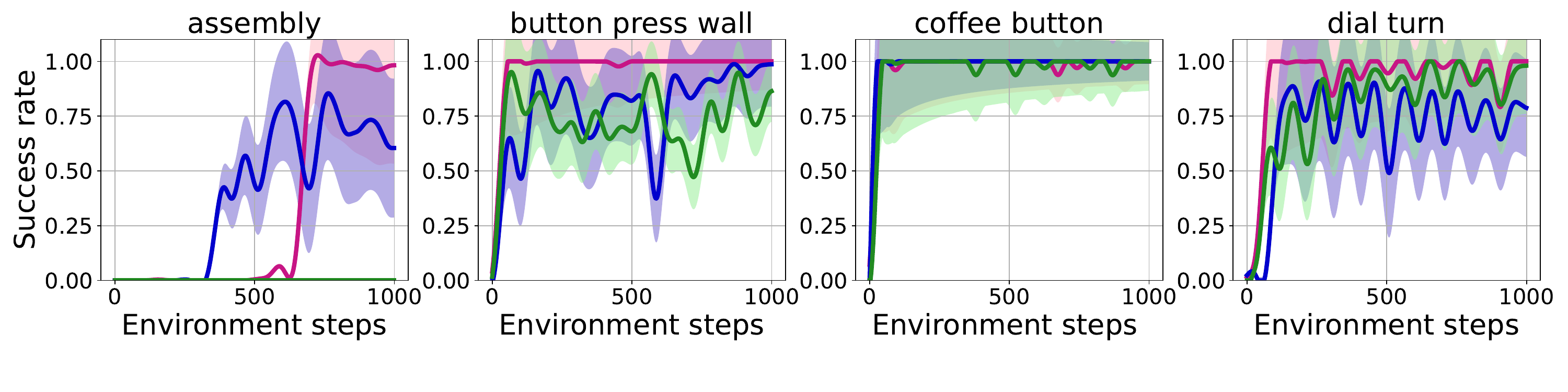}
    \includegraphics[width=1\textwidth]{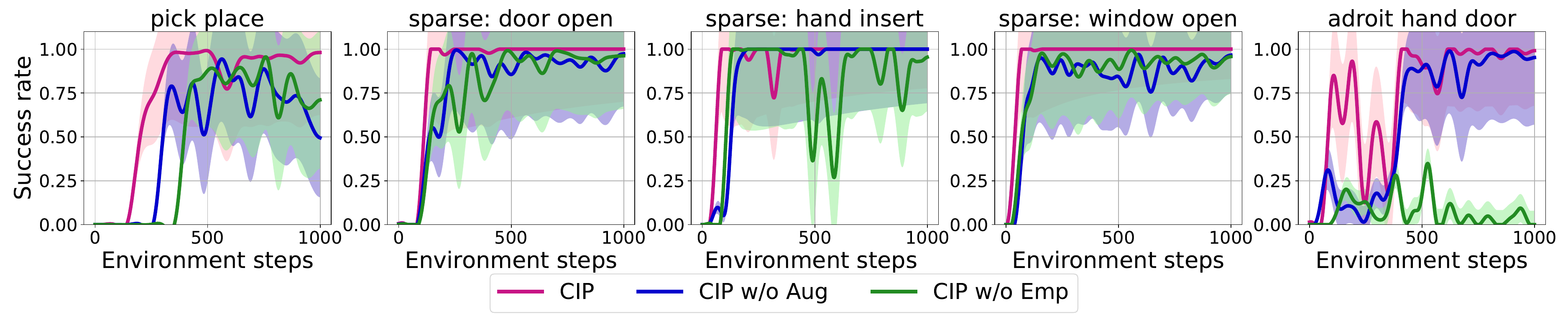} 
    \caption{Ablation results across $21$ manipulation skill learning tasks in Meta-World including sparse reward settings and adroit hand.}
    \label{fig:appendix_abl_manipulation}
\end{figure}

The ablation study results in the Meta-World and Adroit Hand environments are presented in Figure~\ref{fig:appendix_abl_manipulation}. The findings indicate that $\texttt{\textbf{CIP}}$ without counterfactual data augmentation exhibits reduced learning efficiency and is unable to successfully complete tasks such as pick-and-place. This underscores the importance of incorporating counterfactual data augmentation, which prioritizes causal state information, to enhance learning efficiency by mitigating the influence of irrelevant state information and preventing policy divergence.

Furthermore, $\texttt{\textbf{CIP}}$ without causal action empowerment demonstrates a significant decline in policy performance across robot arm manipulation tasks. In complex scenarios, such as Adroit Hand door opening and assembly, it fails to learn effective strategies for task completion. This outcome further corroborates the efficacy of the proposed causal action empowerment mechanism, as prioritizing causally informed actions facilitates more efficient exploration of the environment, ultimately enabling successful policy learning.

\paragraph{Sparse reward settings}
Figure~\ref{fig:appendix_abl_manipulation} presents the results of the ablation study conducted across three sparse reward setting tasks. These findings underscore the substantial influence of causal action empowerment on the efficacy of policy learning, demonstrating its critical role in enhancing performance in challenging environments. Additionally, the incorporation of counterfactual data augmentation proves effective in mitigating the need for additional environmental interactions, thereby significantly improving sample efficiency. This approach not only facilitates more rapid learning but also ensures that the agent can effectively navigate sparse reward scenarios by focusing on the most relevant causal information. 

\begin{figure}[t]
    \centering
    \includegraphics[width=1\textwidth]{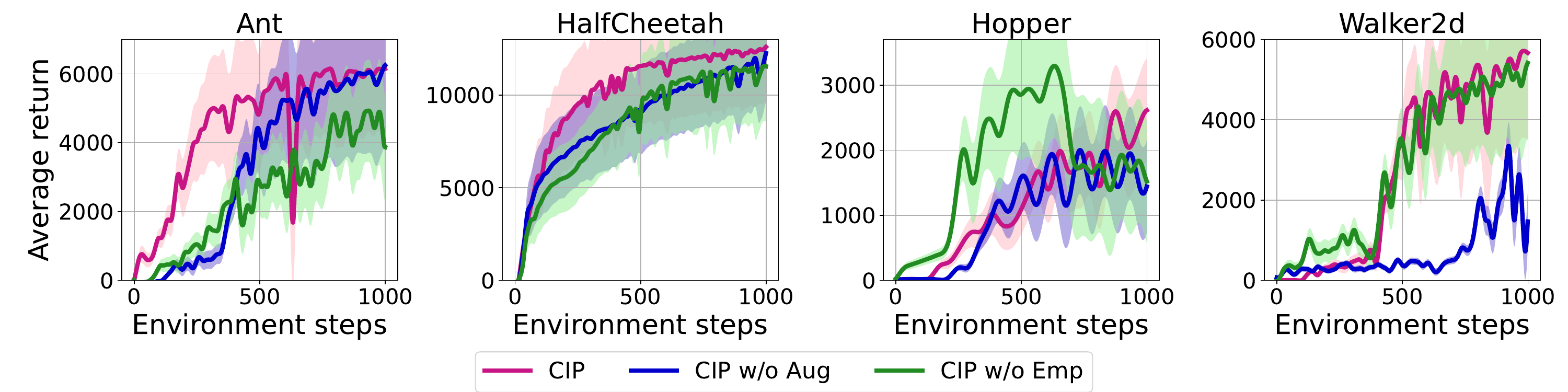}
    \caption{Ablation results across $4$ locomotion tasks in MuJoCo environment.}
    \label{fig:appendix_abl_mujoco}
\end{figure}

\paragraph{Locomotion}
We further conducted ablation experiments on locomotion tasks. The experimental results in the MuJoCo environment are shown in Figure~\ref{fig:appendix_abl_mujoco}, where it is evident that the performance of $\texttt{\textbf{CIP}}$ without causal action empowerment declines significantly. Similarly, $\texttt{\textbf{CIP}}$ without counterfactual data augmentation also exhibits reduced learning efficiency. Notably, in the 11 DMControl tasks, the decline in performance for $\texttt{\textbf{CIP}}$ without causal action empowerment is particularly pronounced. 

These experimental results further validate the effectiveness of our proposed method, which systematically analyzes the causal relationships between states, actions, and rewards. This analysis enables the execution of counterfactual data augmentation to avoid interference from irrelevant factors while prioritizing important state information. Subsequently, by leveraging the causal relationships between actions and rewards, we reweight actions to prioritize causally informed actions, thereby enhancing the agent’s controllability and overall learning efficacy.

\begin{figure}[t]
    \centering
    \includegraphics[width=1\textwidth]{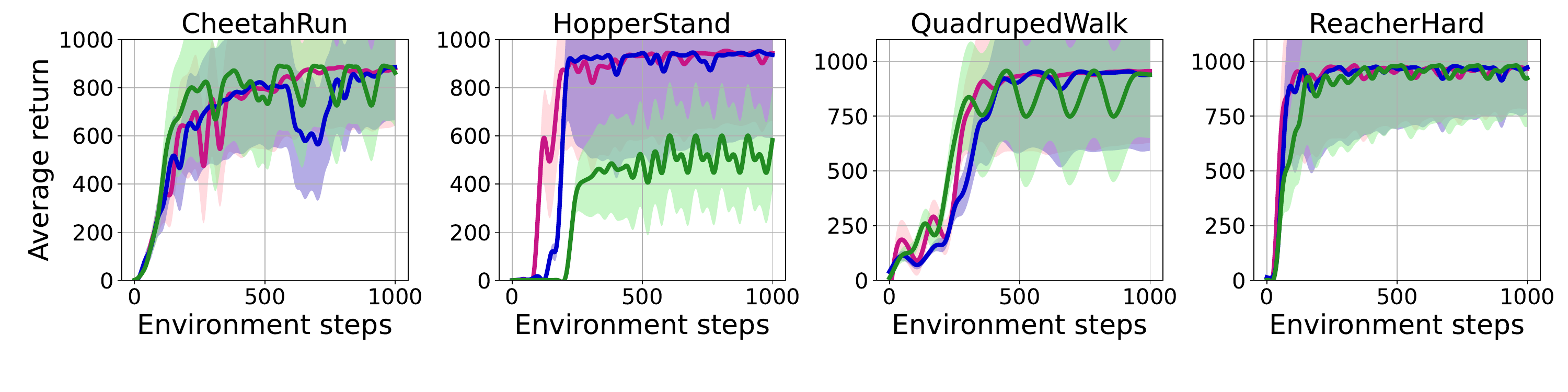}
    \includegraphics[width=1\textwidth]{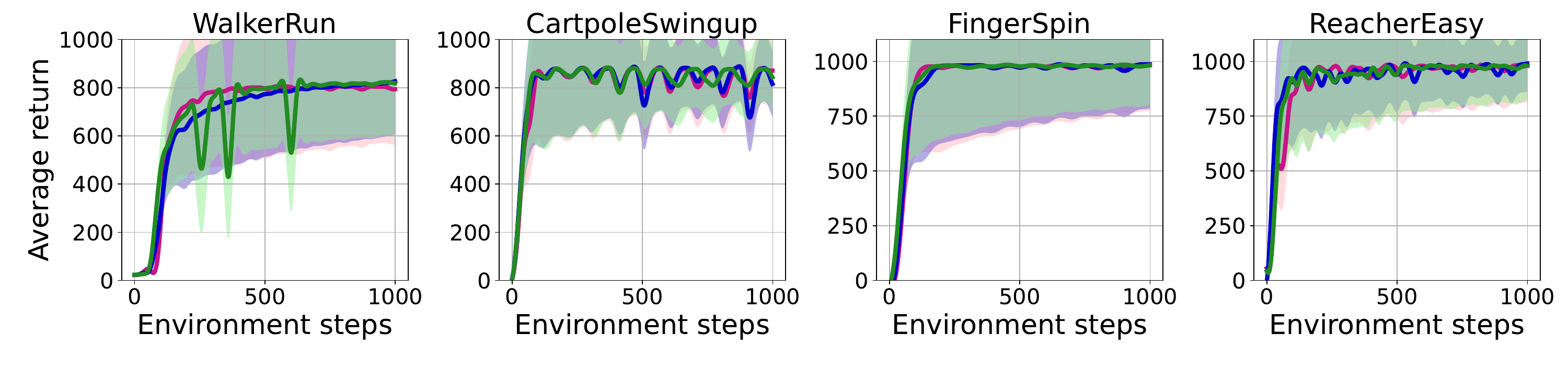}
    \includegraphics[width=0.75\textwidth]{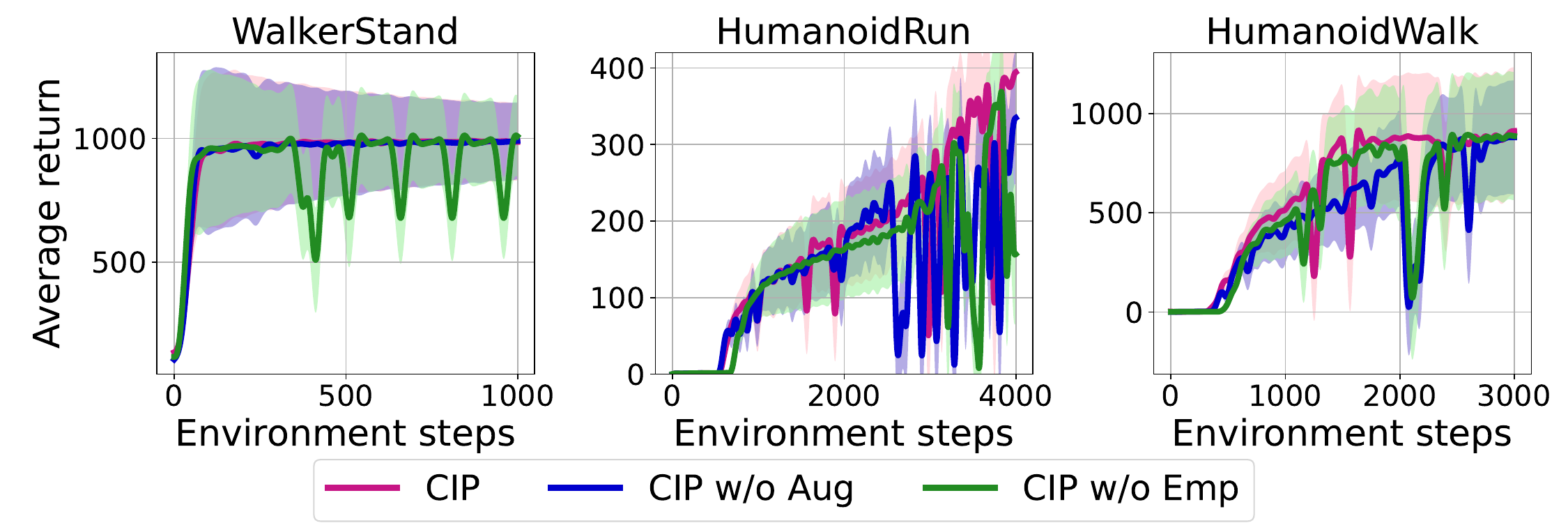}
    \caption{Ablation results across $11$ locomotion tasks in DMControl environment.}
    \label{fig:appendix_abl_dmc}
\end{figure}

\subsubsection{Hyperparameter analysis}
\begin{figure}[h]
    \centering
    \includegraphics[width=1\linewidth]{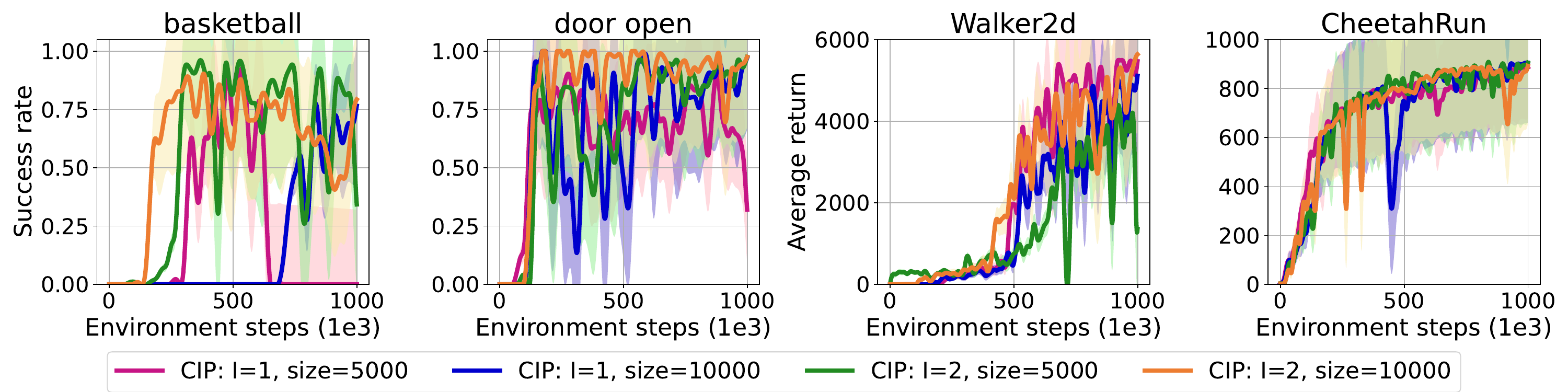}
    \caption{Hyperparameter study. Learning curves of $\texttt{\textbf{CIP}}$ with different hyperparameter settings. The shaded regions are the standard deviation of each policy.}
    \label{fig:hps}
\end{figure}

We conduct a detailed analysis of the hyperparameters associated with the causal update interval (\(I\)) and sample size within the $\texttt{\textbf{CIP}}$ framework. The experimental results for four distinct tasks are illustrated in Figure~\ref{fig:hps}. Across all tasks,  $\texttt{\textbf{CIP}}$ demonstrates optimal performance with a causal update interval of \(I = 2\) and a sample size of 10,000. 

Our findings suggest that while a reduction in the causal update interval can lead to improved performance, it may also result in heightened computational costs. Additionally, we observe that higher update frequencies and increased sample sizes introduce greater instability, which significantly raises computational demands. This analysis underscores the importance of carefully balancing hyperparameter settings to optimize both performance and efficiency within the $\texttt{\textbf{CIP}}$.

\begin{figure}[t]
    \centering
    \includegraphics[width=1\textwidth]{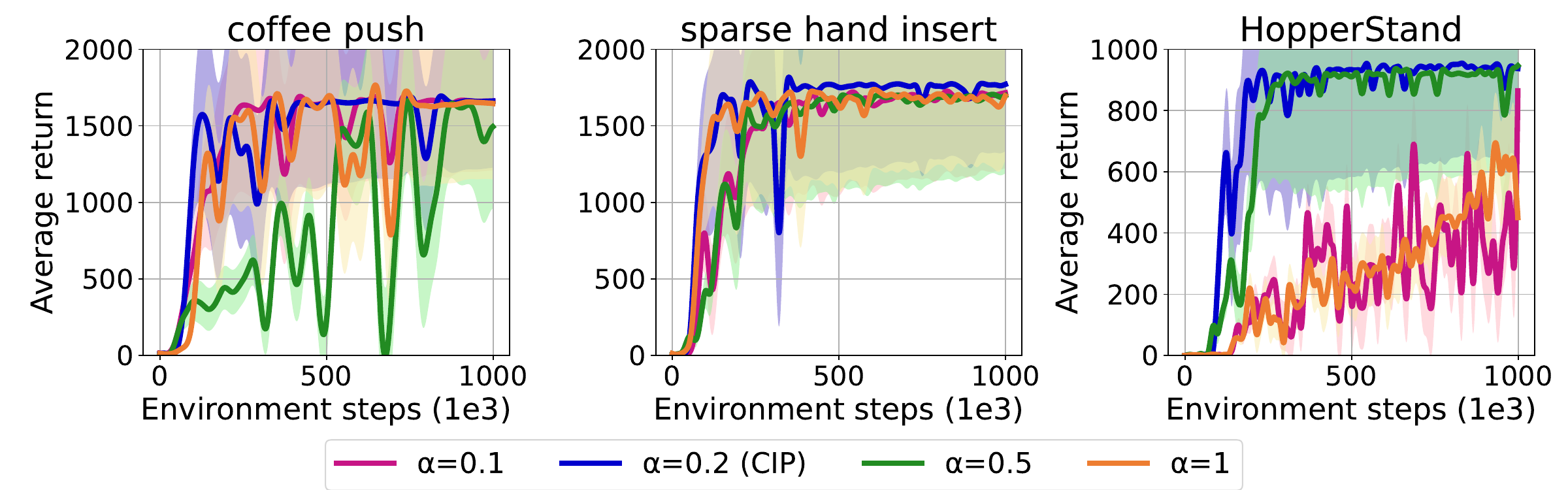}
    \caption{Hyperparameter analysis of temperature factor $\alpha$ across $3$ task.}
    \label{fig:appendix_hps_a}
\end{figure}

\textcolor{black}{Furthermore, we analyze the performance under different settings of the temperature factor $\alpha$ proposed in Eq.~\ref{eq:alpha}. The results across $3$ tasks are shown in Figure~\ref{fig:appendix_hps_a}. Our analysis reveals that $\texttt{\textbf{CIP}}$ demonstrates robust performance across different values of $\alpha$ in manipulation tasks, while showing some instability in locomotion tasks when $\alpha$ is either too small or too large. Moreover, we observe that setting $\alpha$ to 0.2 yields optimal performance across all tasks, which motivated our choice of $\alpha = 0.2$ for all experiments.}

\begin{figure}[t]
    \centering
    \includegraphics[width=1\textwidth]{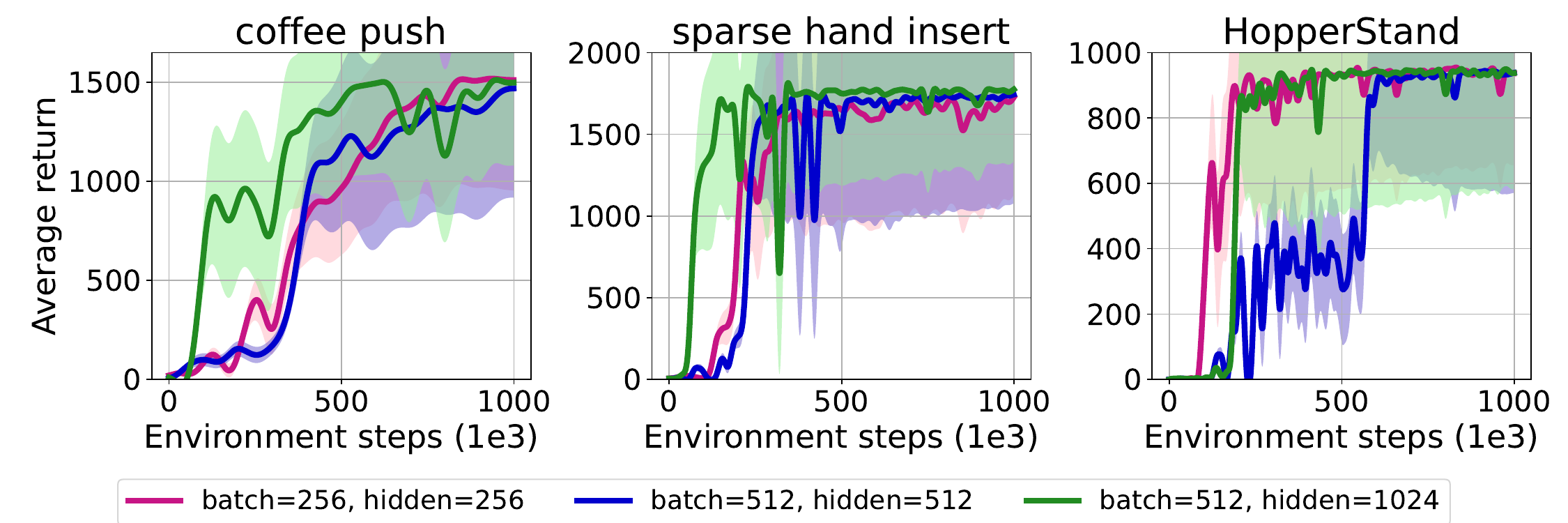}
    \caption{Hyperparameter analysis of batch size and hidden size across $3$ task.}
    \label{fig:appendix_hps_size}
\end{figure}

\textcolor{black}{Finally, we analyze the performance under different settings of the batch size and hidden size. The results across $3$ tasks are shown in Figure~\ref{fig:appendix_hps_size}. Our experimental results demonstrate that $\texttt{\textbf{CIP}}$ exhibits robust performance across various parameter settings in coffee push and sparse hand insert tasks, while maintaining strong performance in hopper stand task. 
Based on these experimental results, we configure the hyperparameters as follows: for manipulation tasks, we set the batch size to 512 and hidden size to 1024, while for locomotion tasks, we use a batch size of 256 and hidden size of 256. All other hyperparameters remain constant across all tasks, as detailed in Table~\ref{tab:hps}.}

\begin{figure}[t]
    \centering
    \includegraphics[width=1\textwidth]{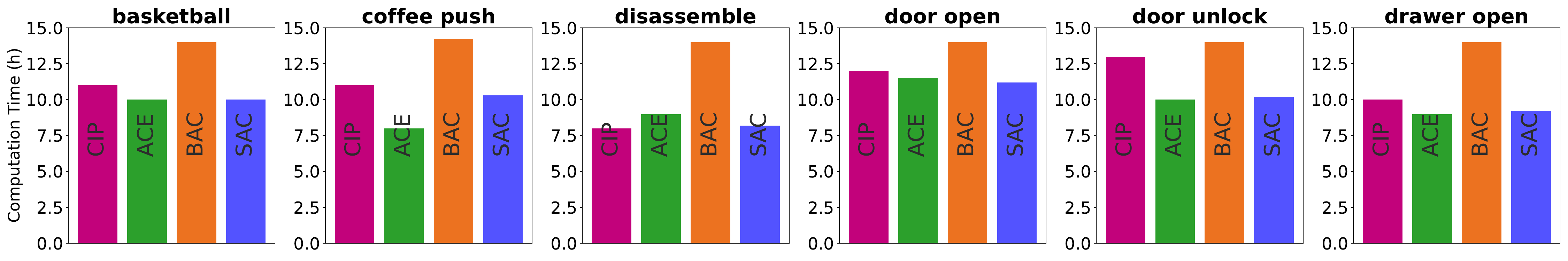}
    \includegraphics[width=1\textwidth]{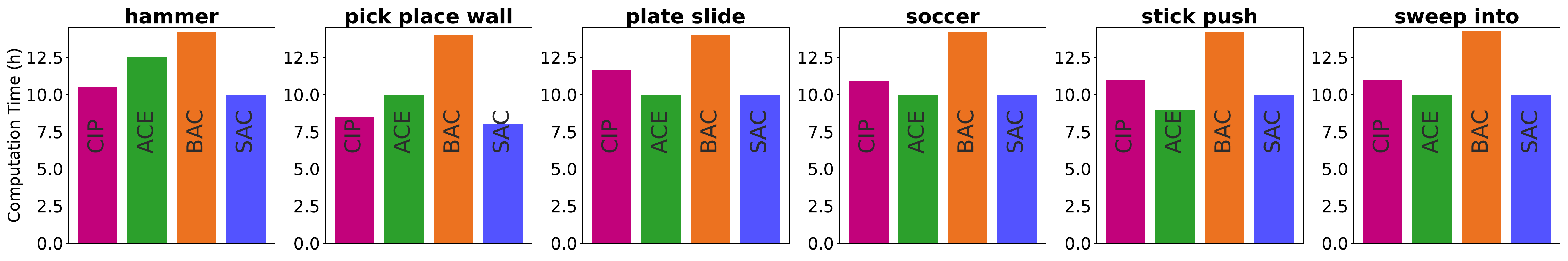}
    \includegraphics[width=1\textwidth]{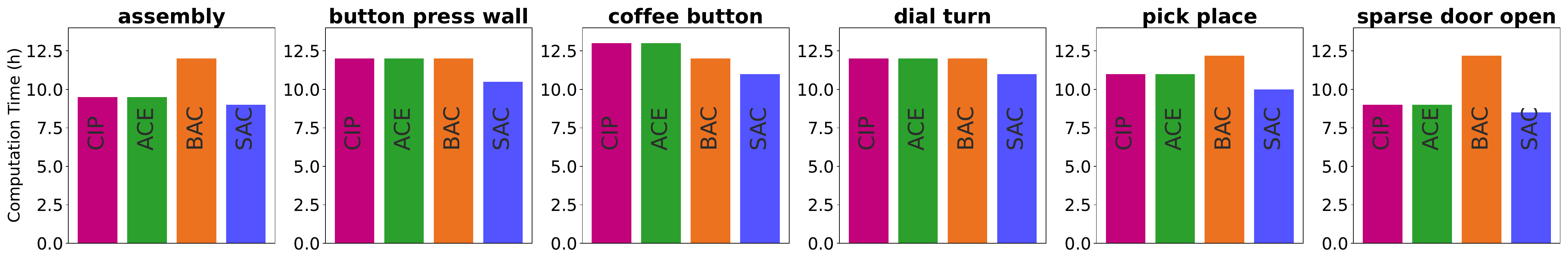}
    \includegraphics[width=1\textwidth]{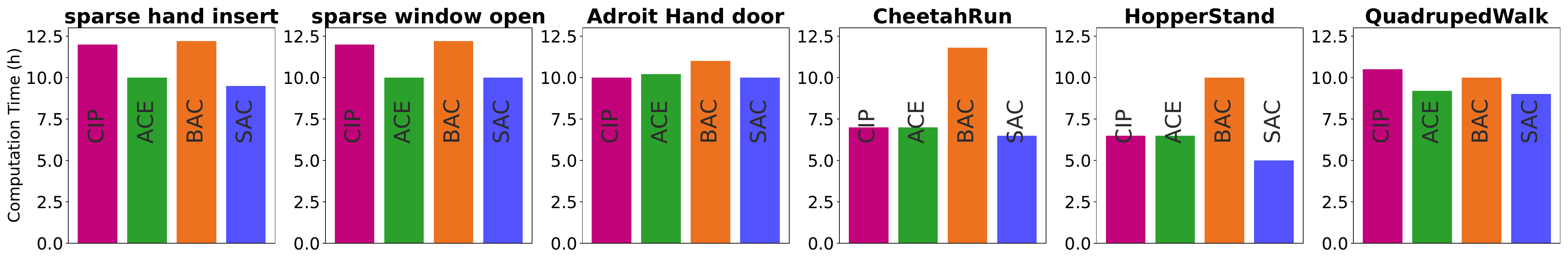}
    \includegraphics[width=1\textwidth]{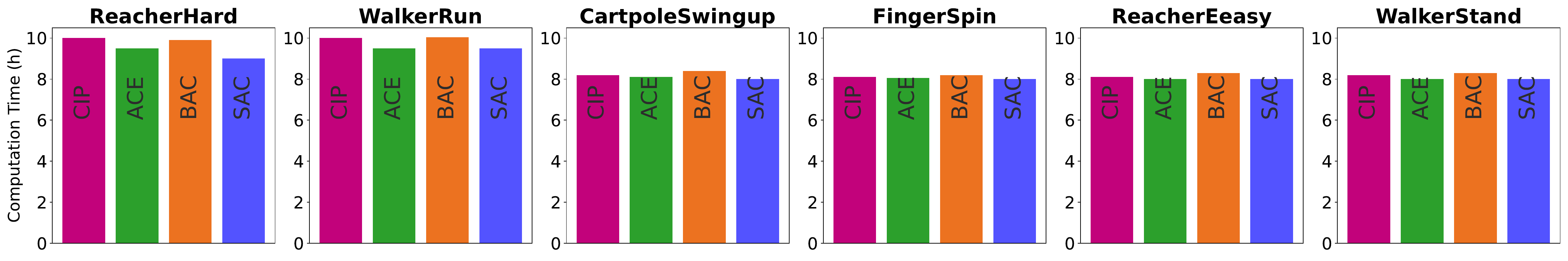} 
    \includegraphics[width=1\textwidth]{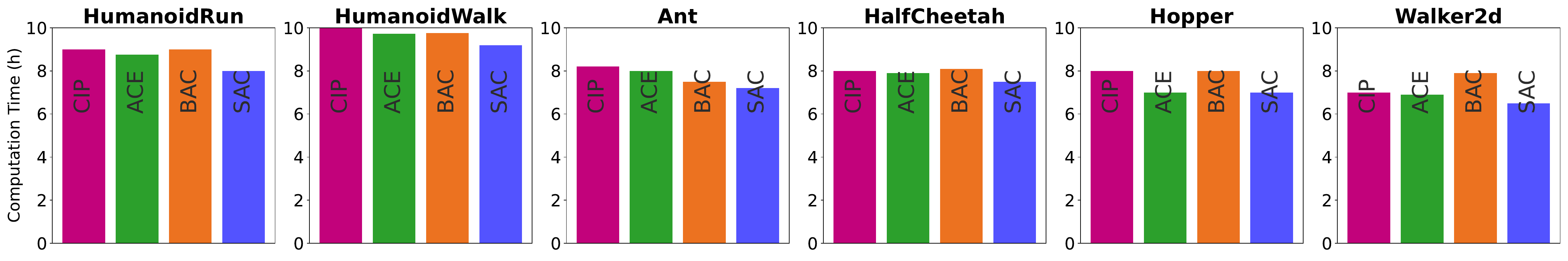} 
    \caption{Computation time in $36$ tasks.}
    \label{fig:appendix_time}
\end{figure}

\textcolor{black}{
\subsubsection{Computation cost analysis}
We analyze the computational cost of the proposed framework. The computation time for all methods across 36 tasks is shown in Figure~\ref{fig:appendix_time}. Our experimental results demonstrate that CIP achieves its performance improvements with minimal additional computational burden - specifically less than $10\%$ increase compared to SAC, less than $5\%$ increase compared to ACE, and actually requiring less computation time than BAC. All experiments were conducted on the same computing platform with the same computational resources detailed in Appendix~\ref{exp}. 
These experimental results verify that our proposed method achieves performance improvements without incurring significant additional computational costs.}

\begin{figure}[t]
    \centering
    \includegraphics[width=0.32\textwidth]{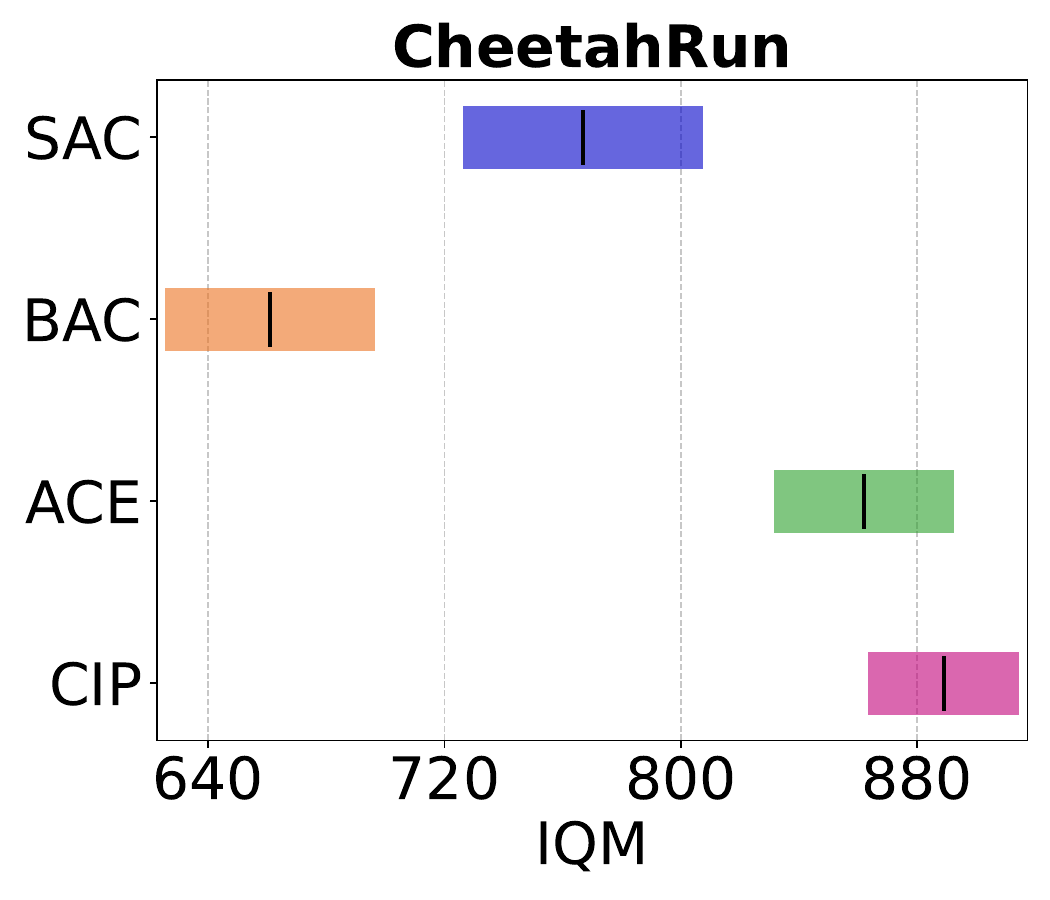}
    \includegraphics[width=0.32\textwidth]{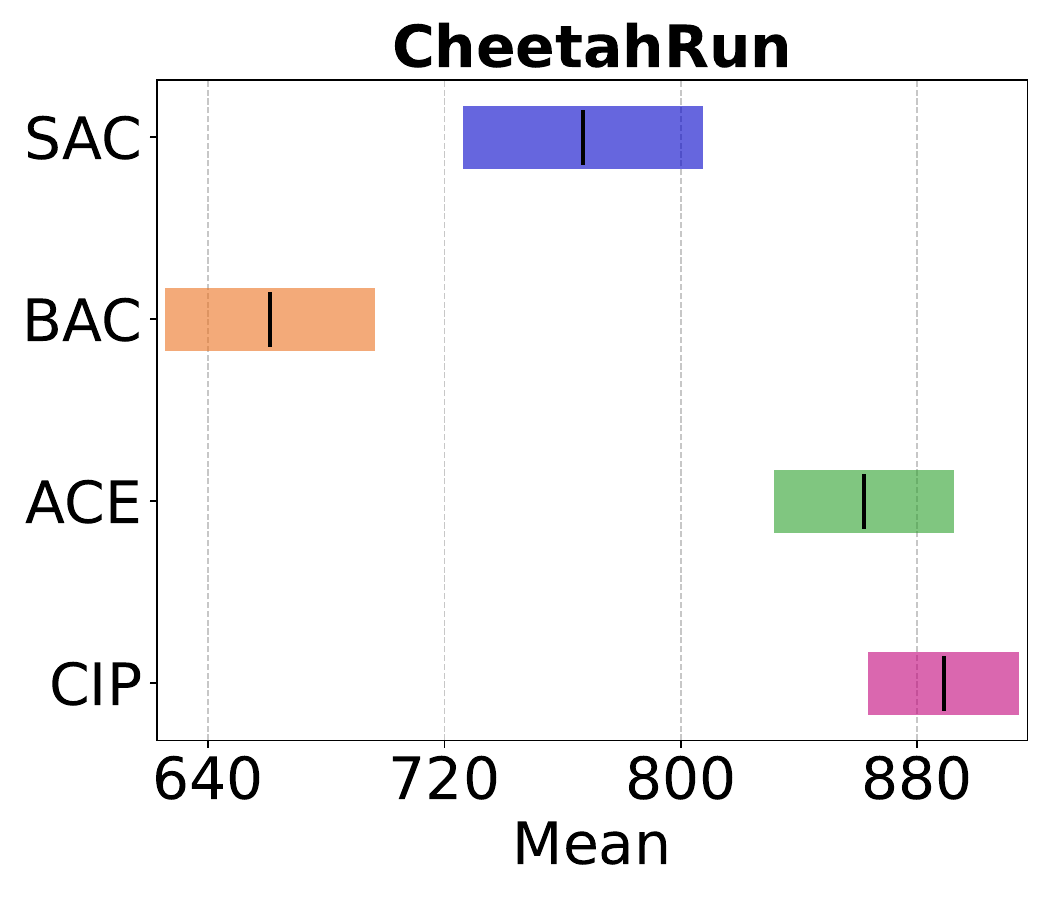}
    \includegraphics[width=0.32\textwidth]{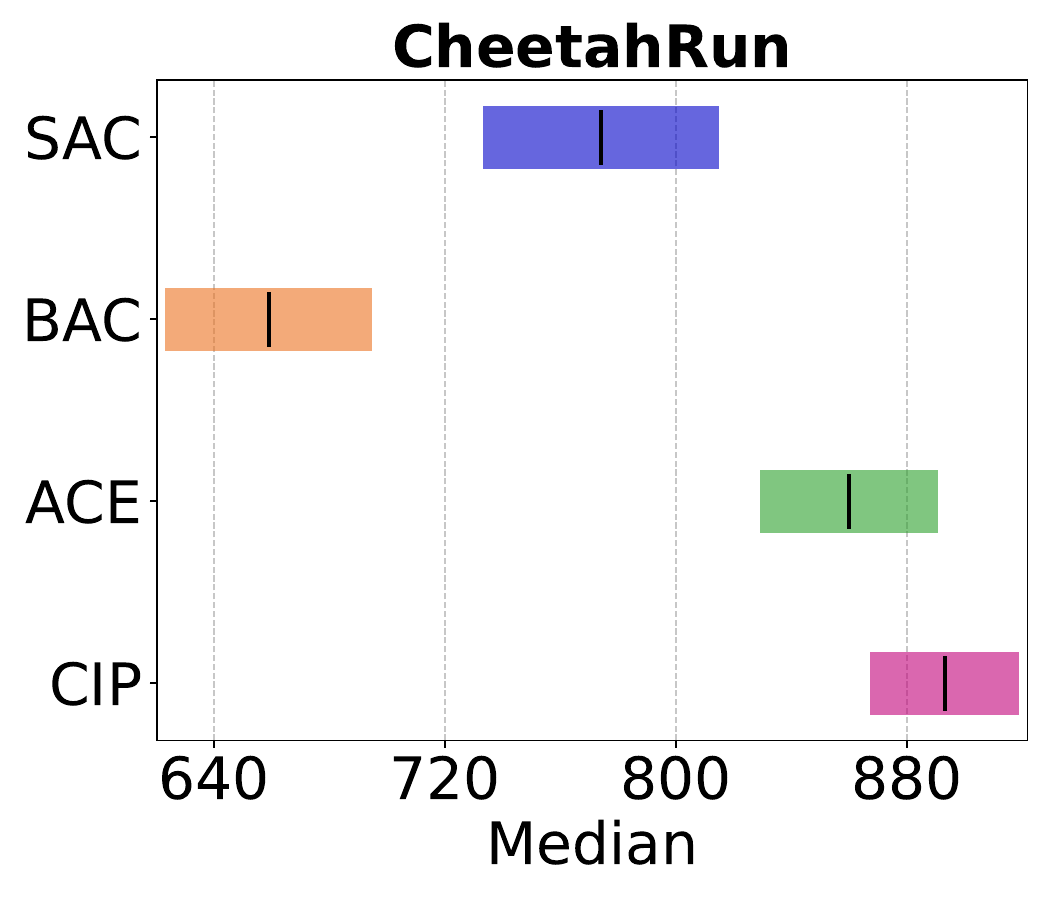}
    \includegraphics[width=0.32\textwidth]{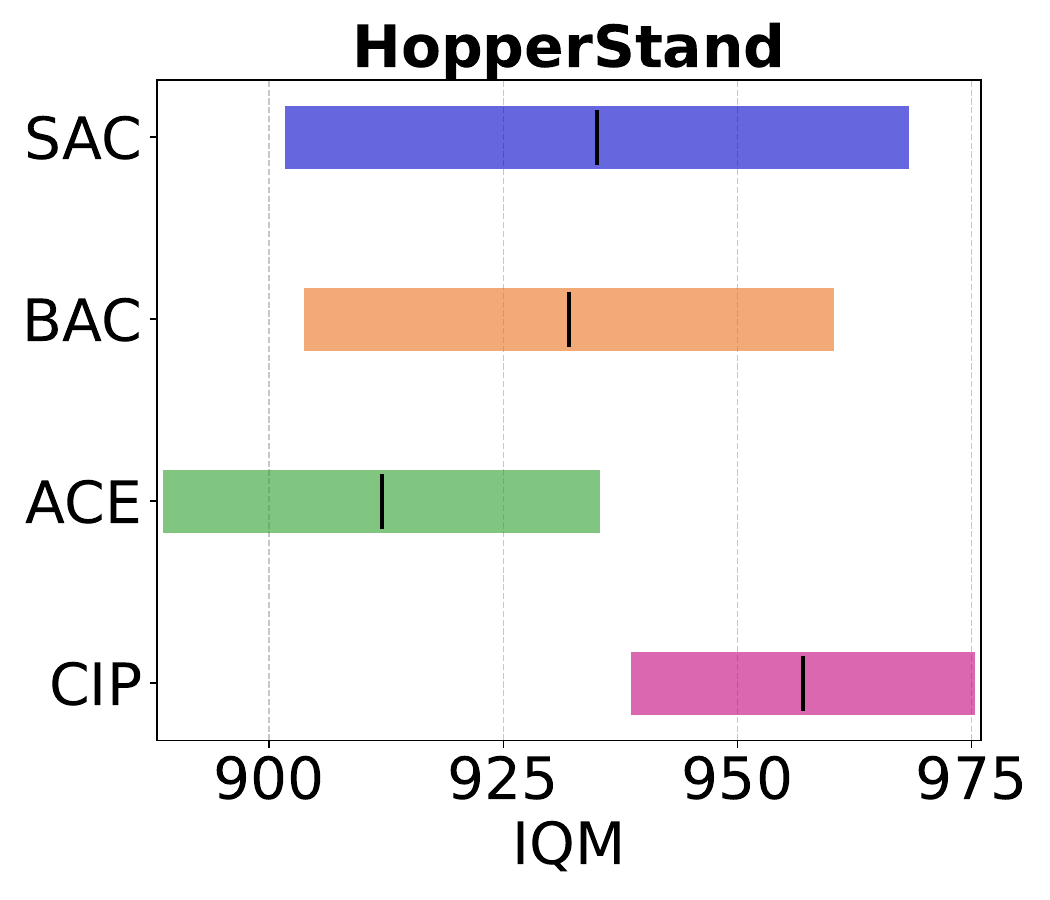}
    \includegraphics[width=0.32\textwidth]{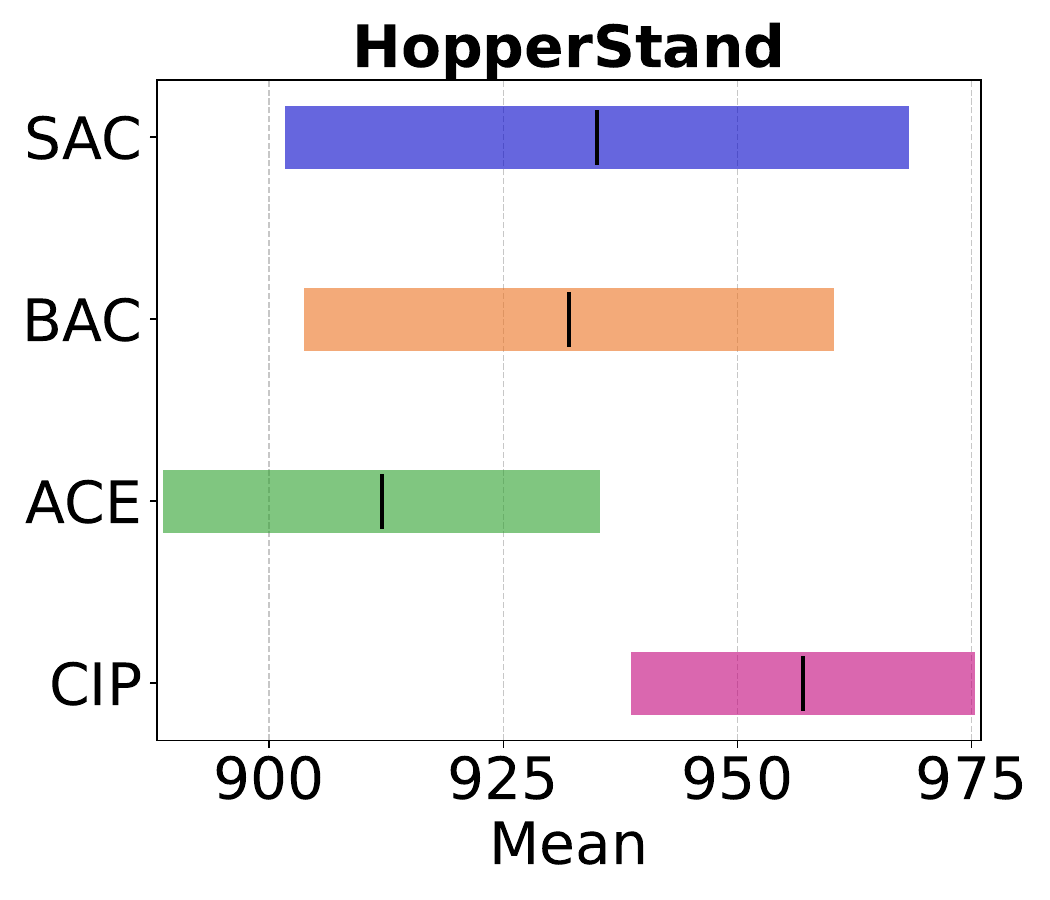}
    \includegraphics[width=0.32\textwidth]{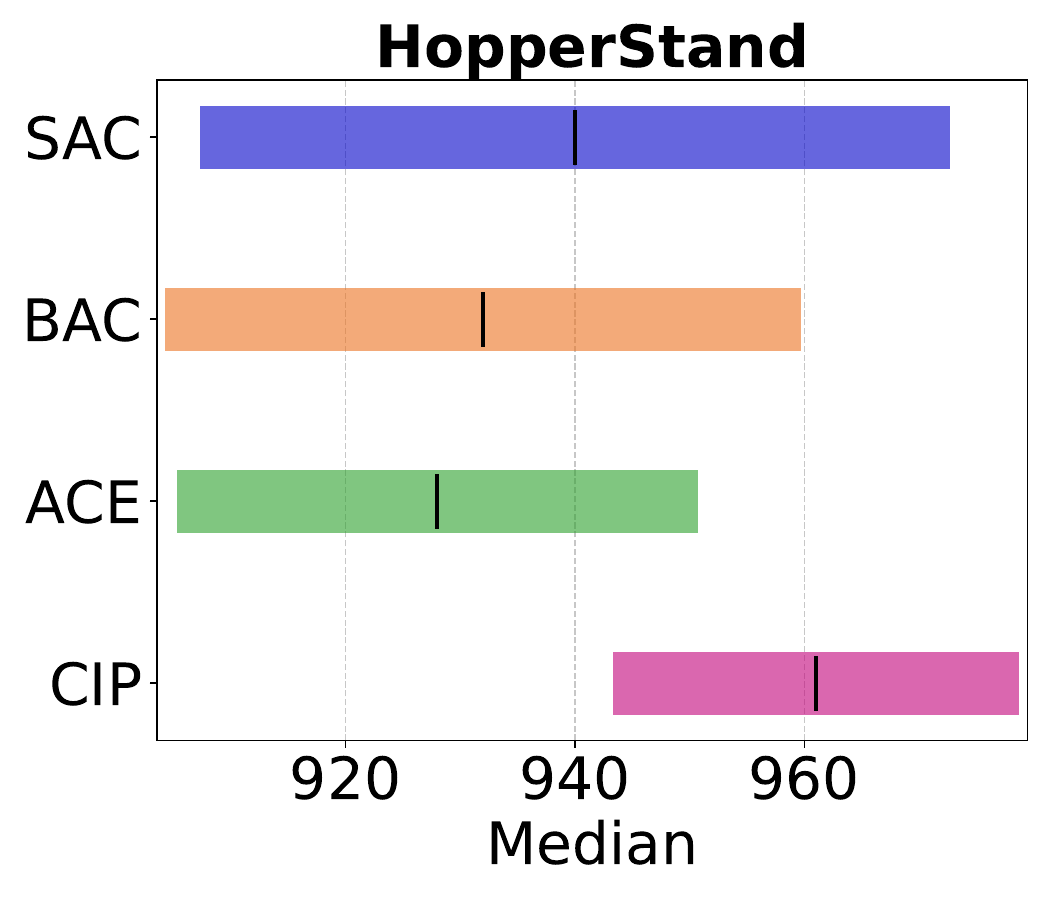}
    \includegraphics[width=0.32\textwidth]{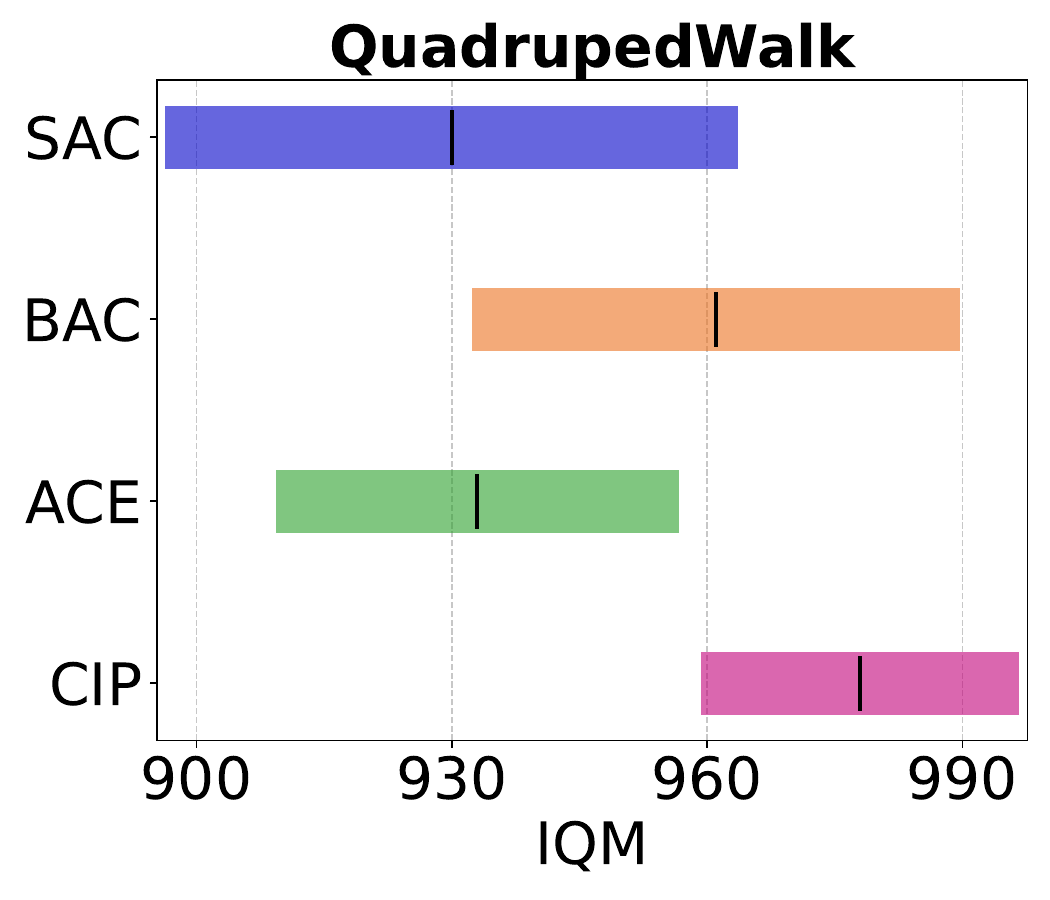}
    \includegraphics[width=0.32\textwidth]{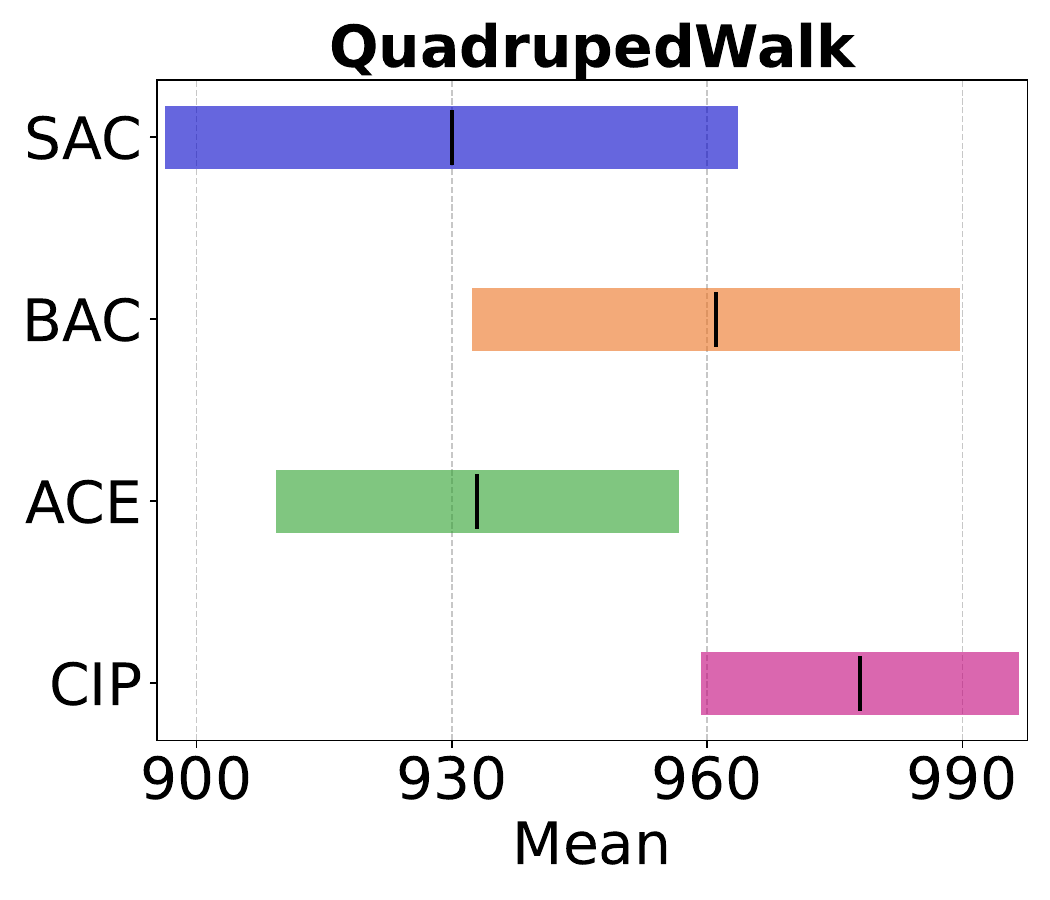}
    \includegraphics[width=0.32\textwidth]{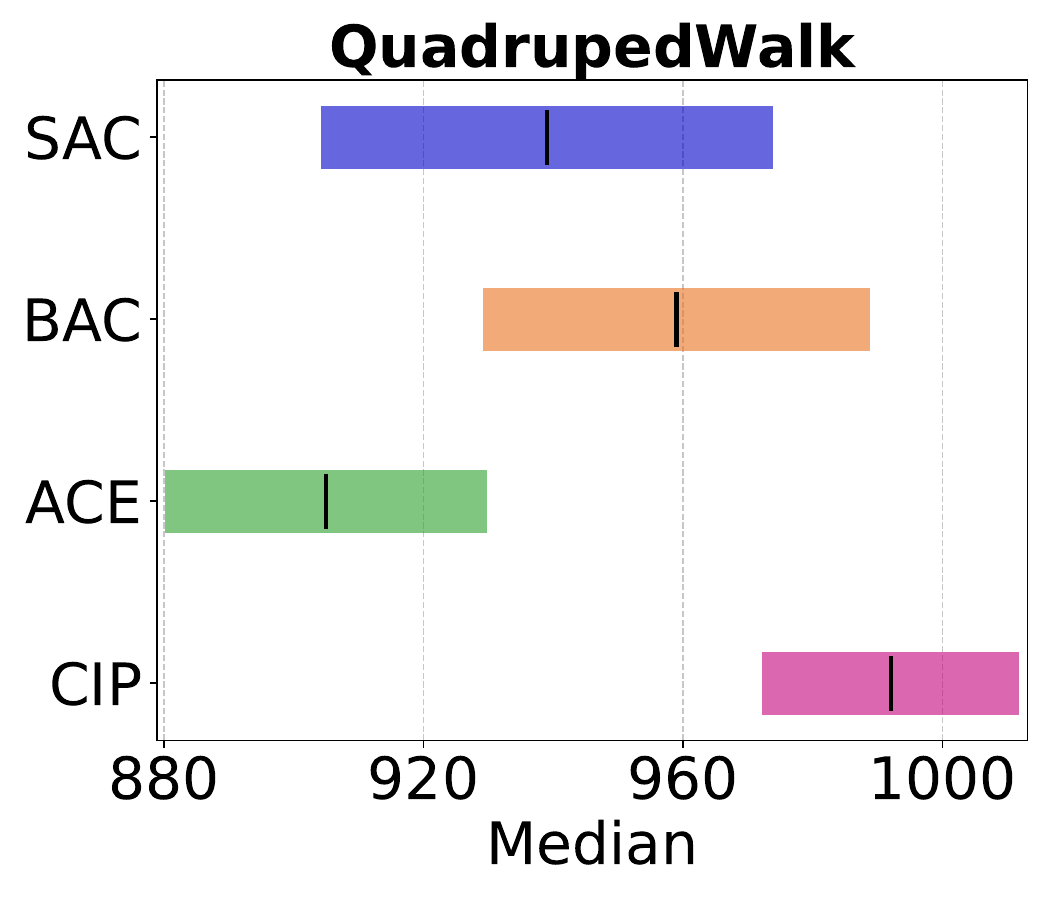}
    \includegraphics[width=0.32\textwidth]{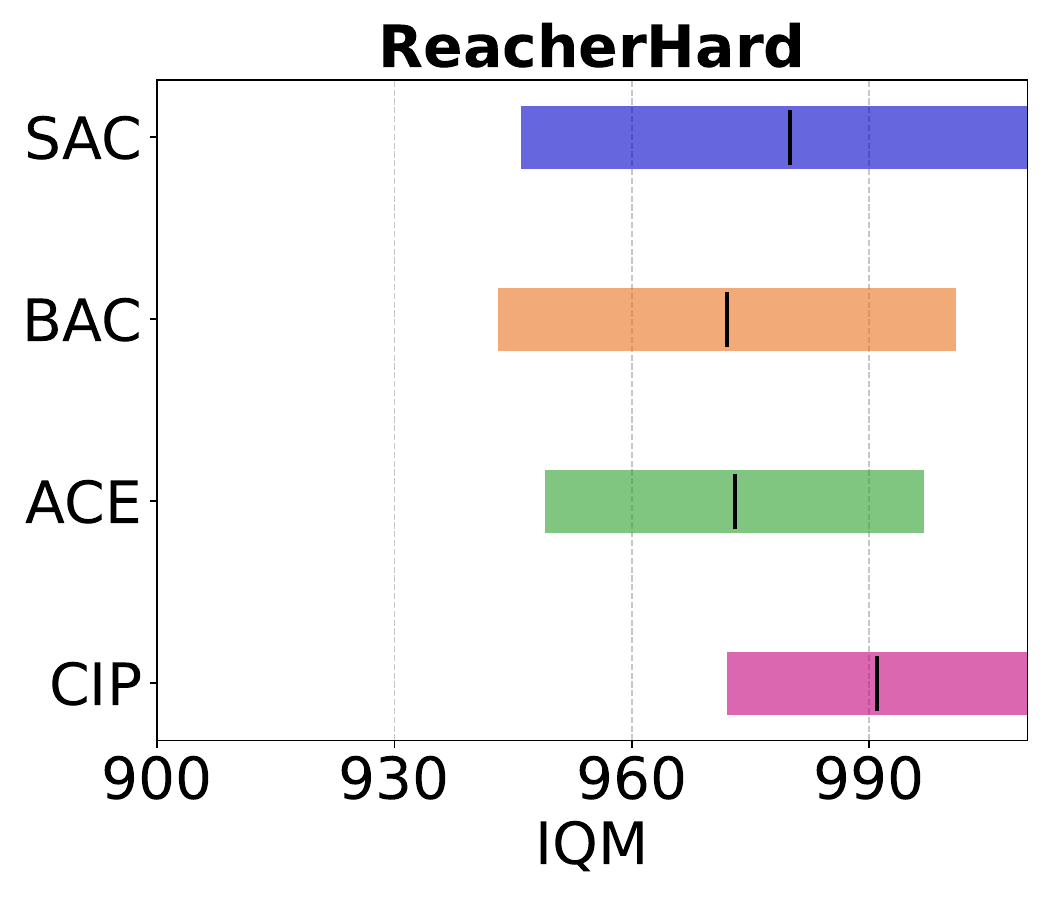}
    \includegraphics[width=0.32\textwidth]{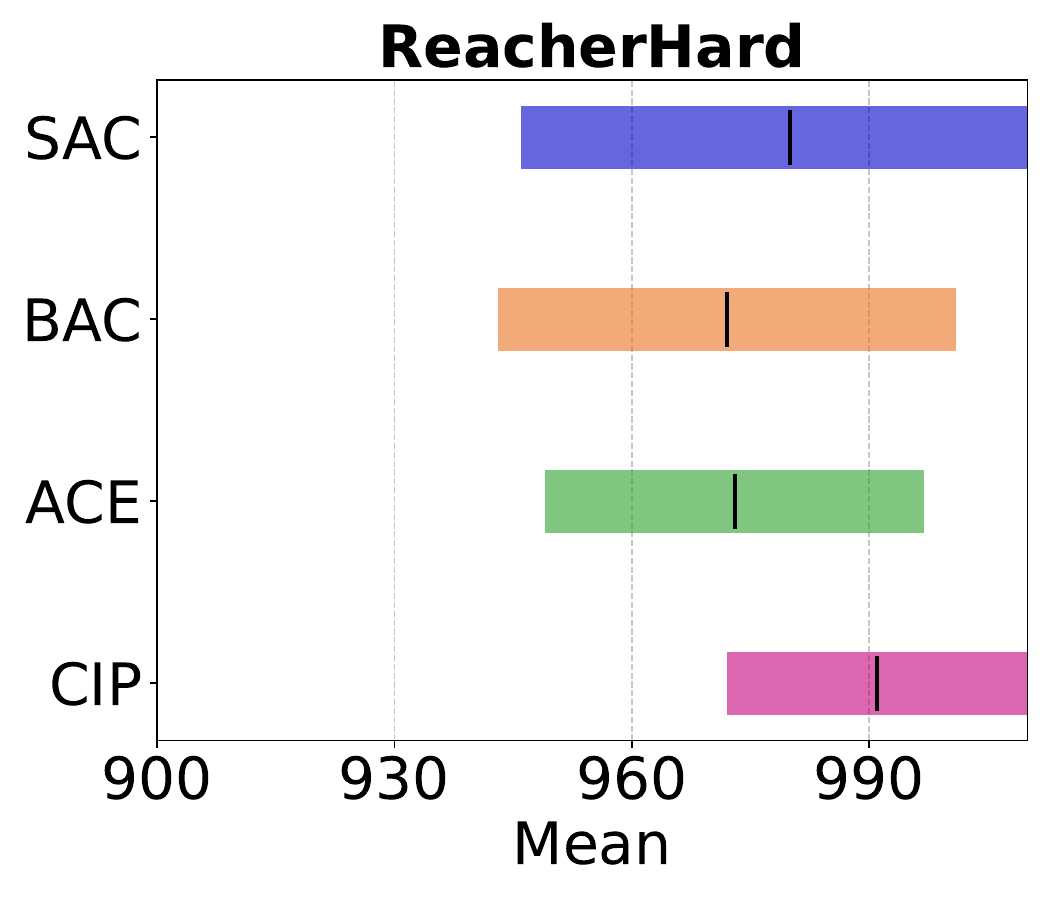}
    \includegraphics[width=0.32\textwidth]{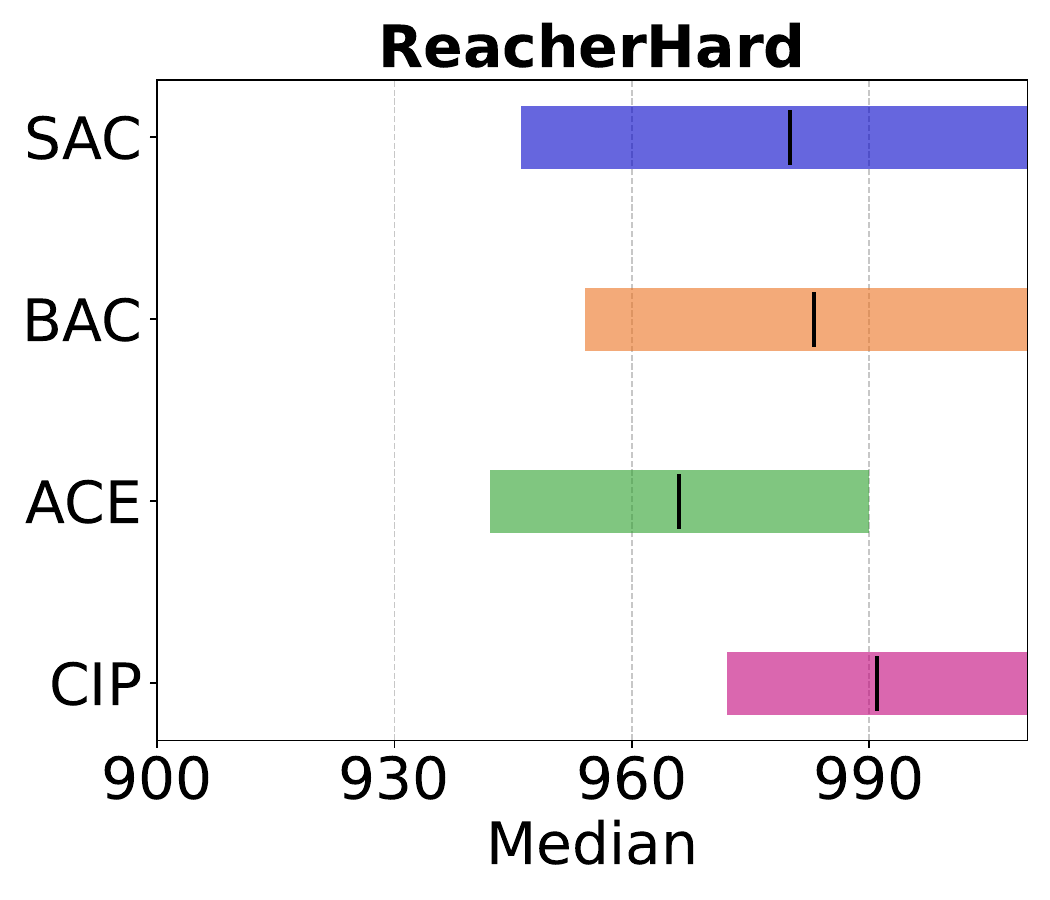}
    \caption{statistical metrics of IQM, Mean, and Median (higher values are better) on 4 DMControl tasks.}
    \label{fig:appendix_IQM_1}
\end{figure}

\begin{figure}[t]
    \centering
    \includegraphics[width=0.32\textwidth]{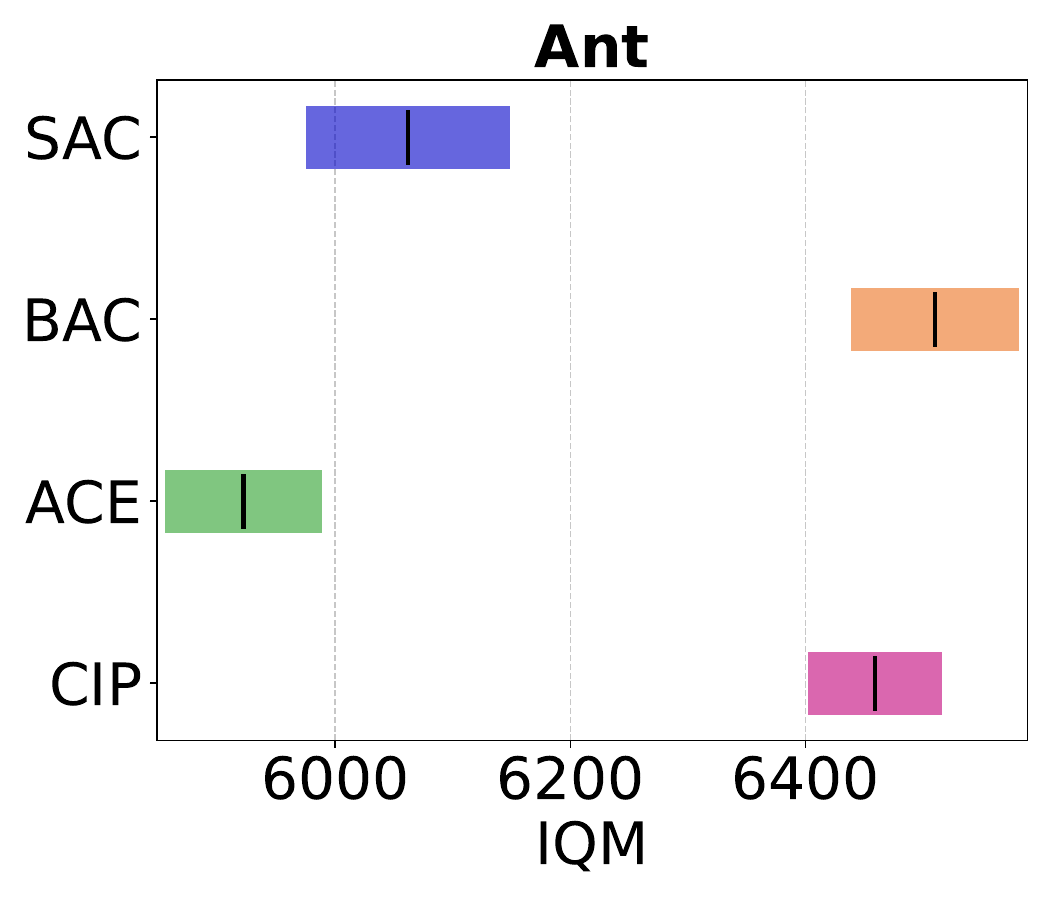}
    \includegraphics[width=0.32\textwidth]{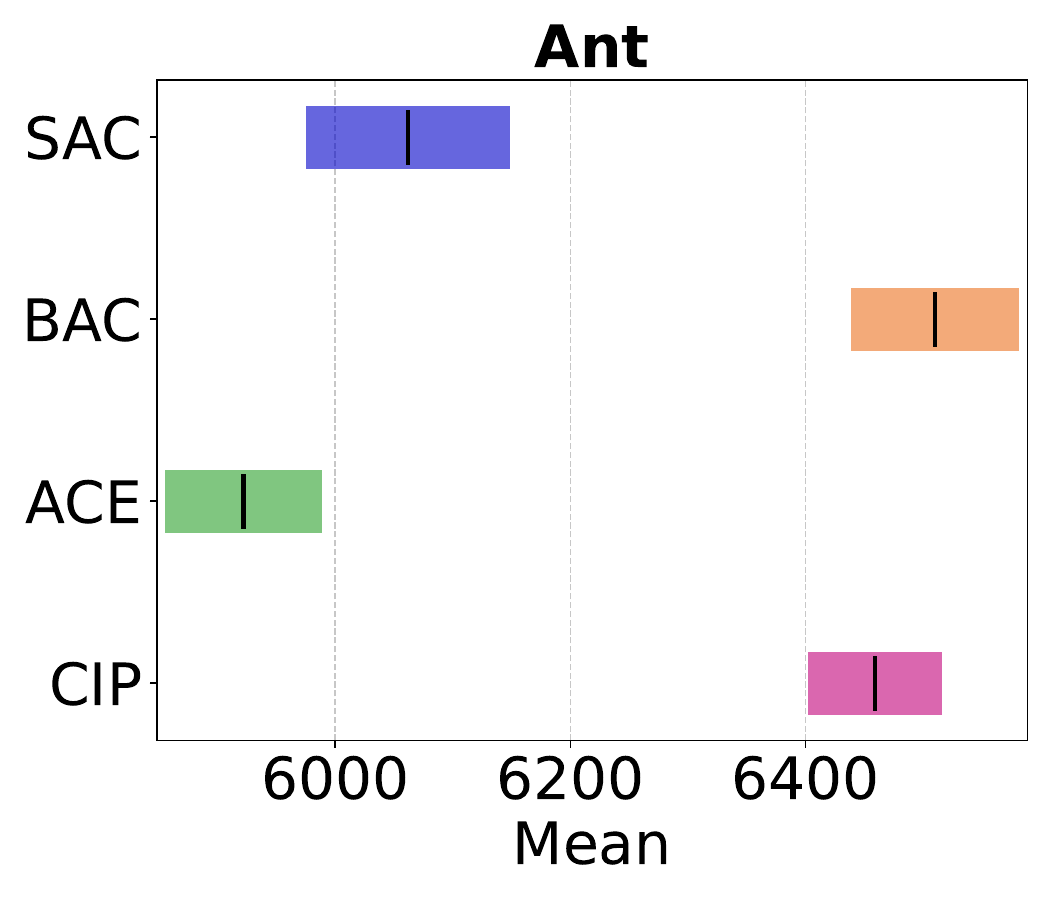}
    \includegraphics[width=0.32\textwidth]{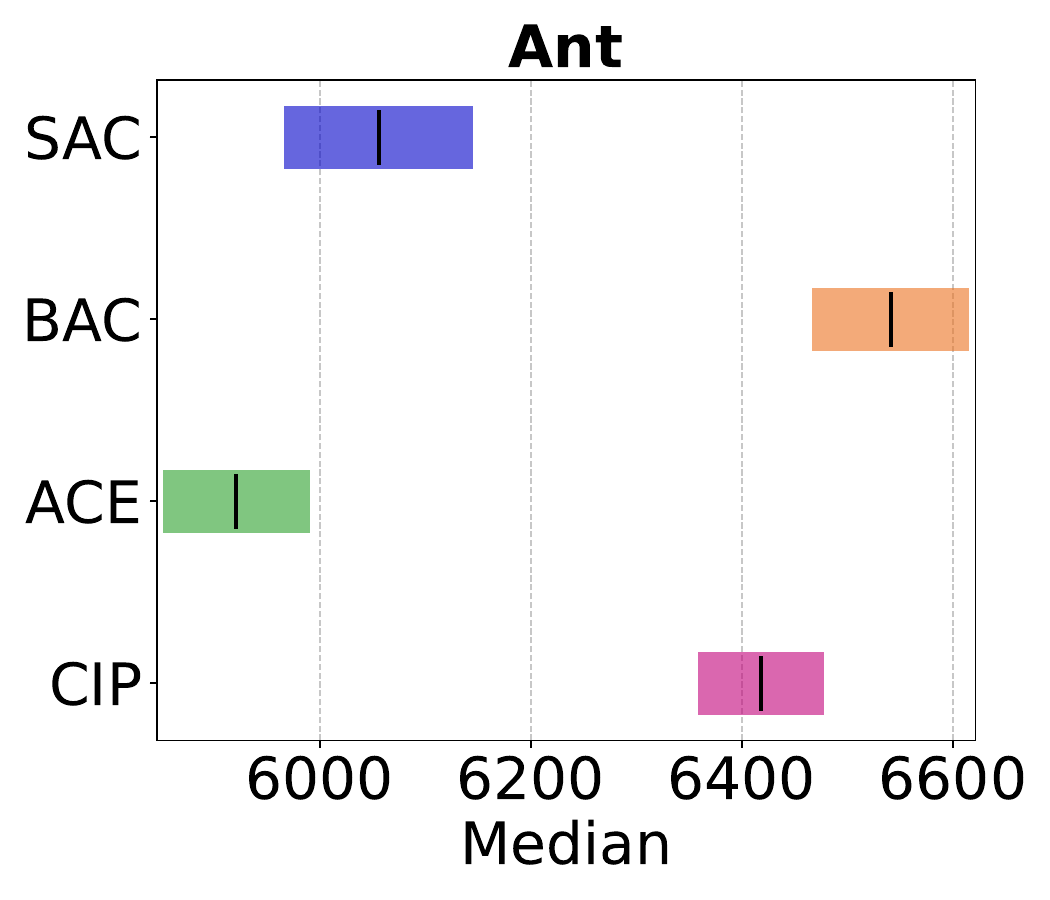}
    \includegraphics[width=0.32\textwidth]{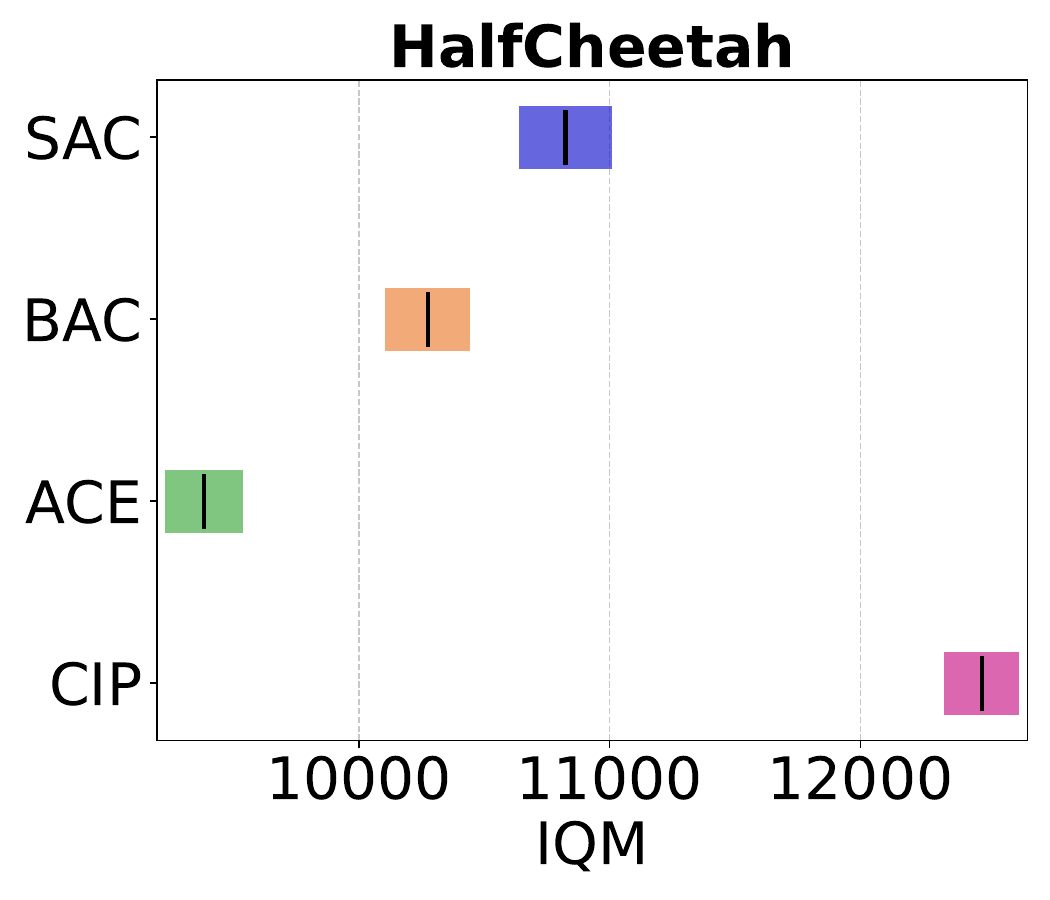}
    \includegraphics[width=0.32\textwidth]{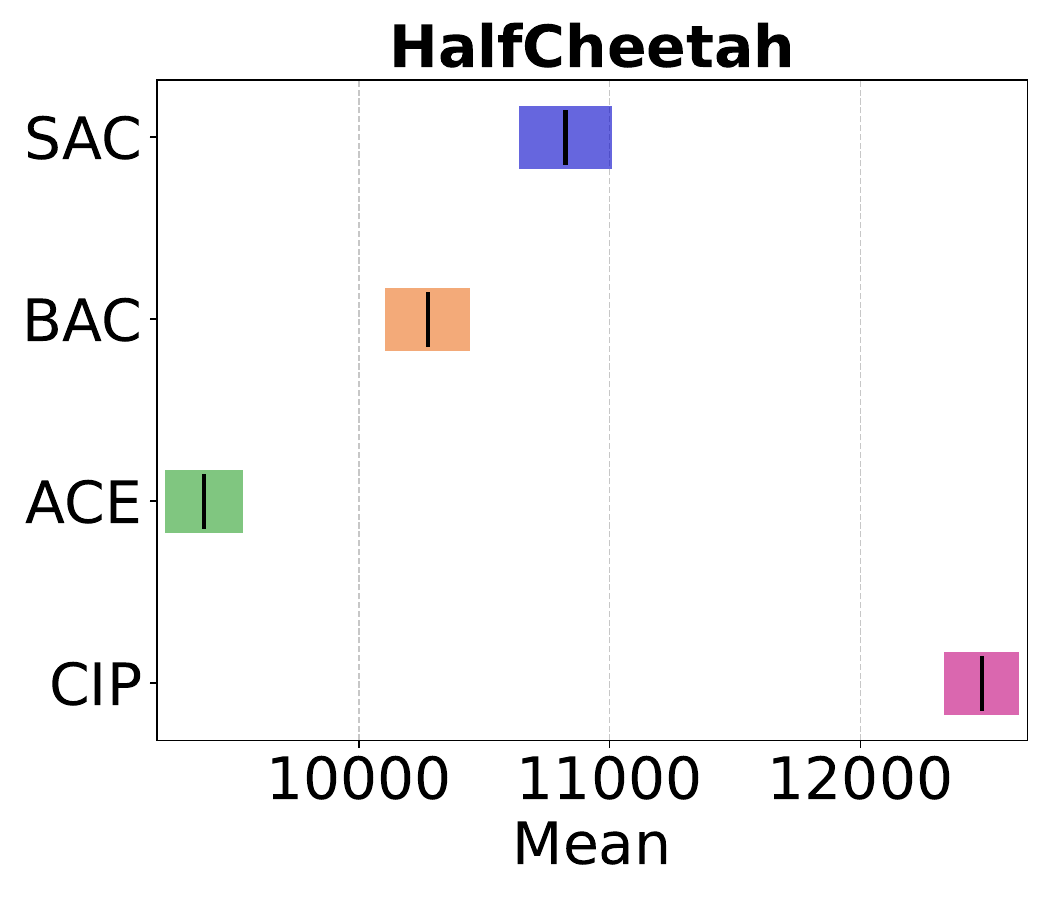}
    \includegraphics[width=0.32\textwidth]{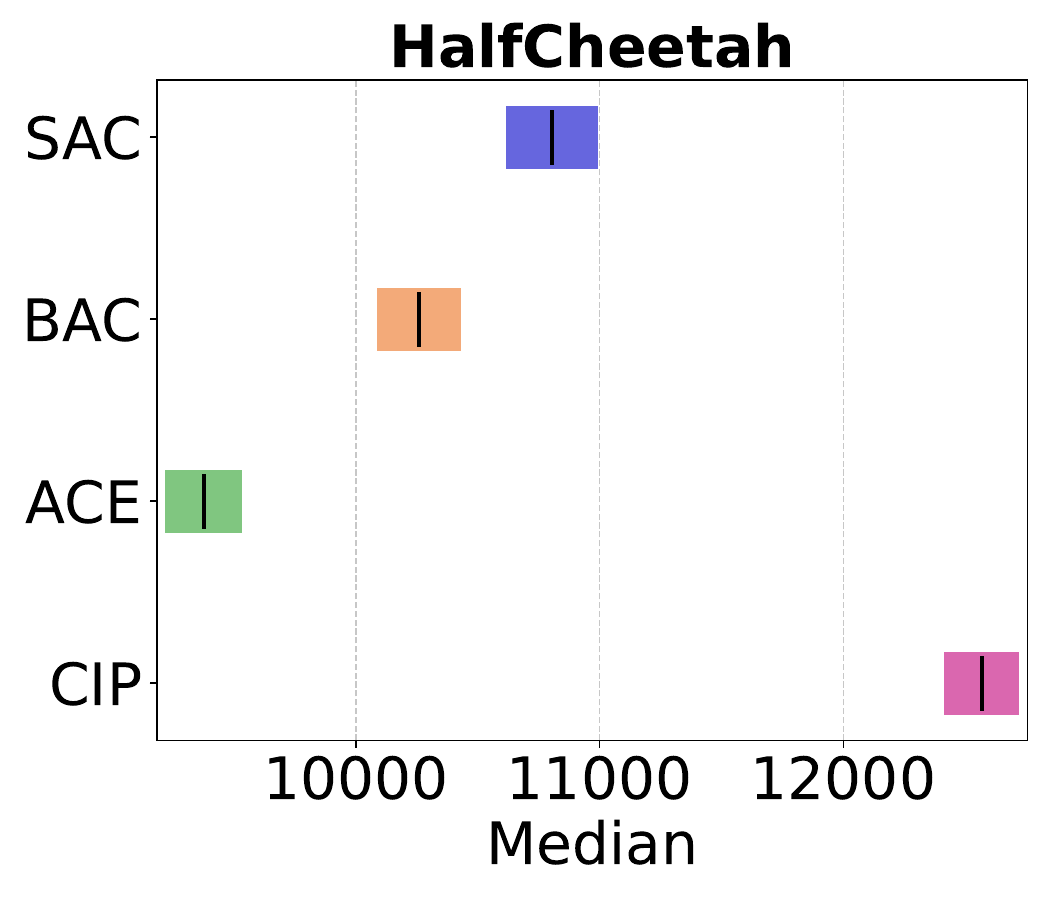}
    \includegraphics[width=0.32\textwidth]{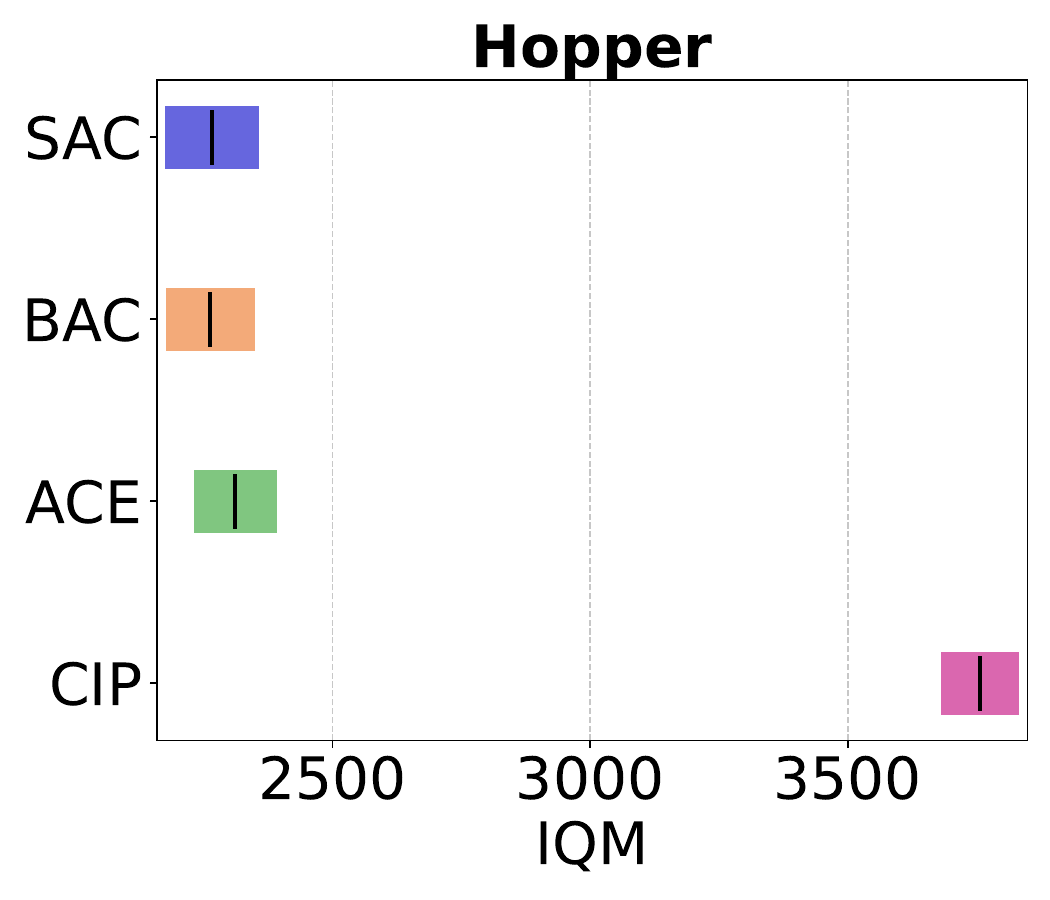}
    \includegraphics[width=0.32\textwidth]{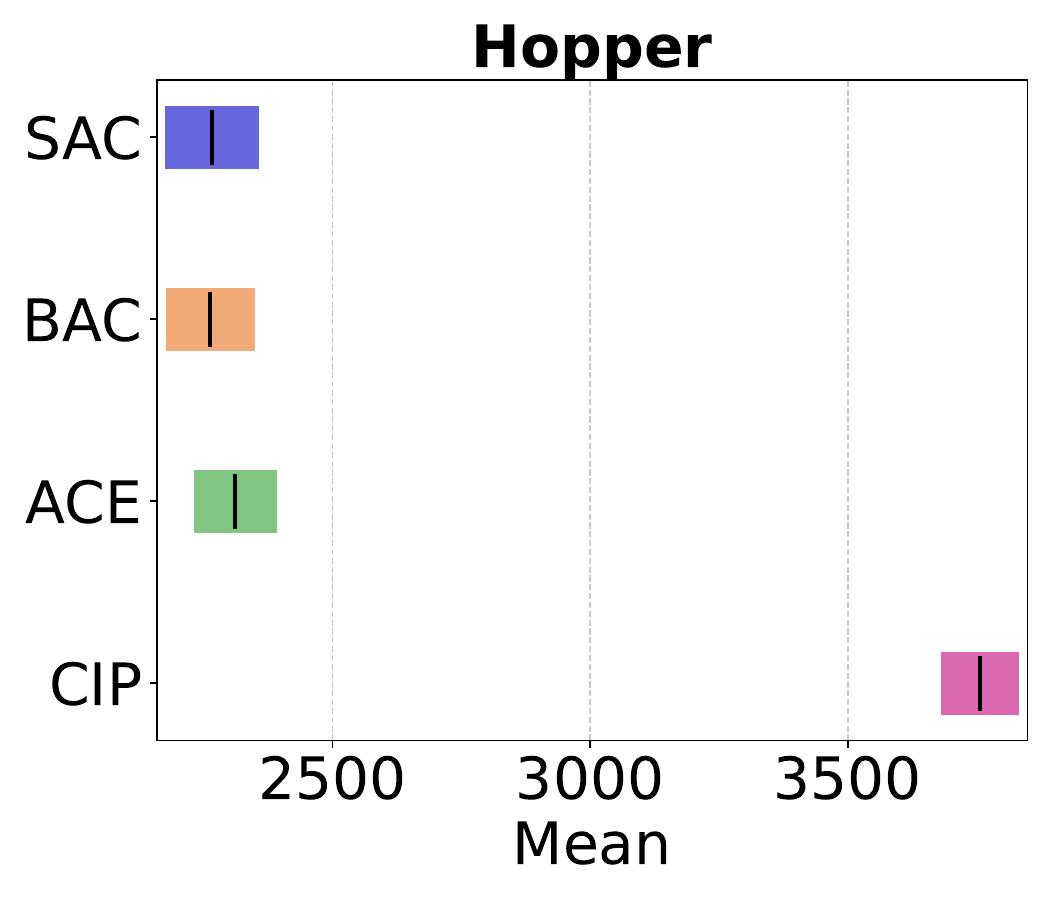}
    \includegraphics[width=0.32\textwidth]{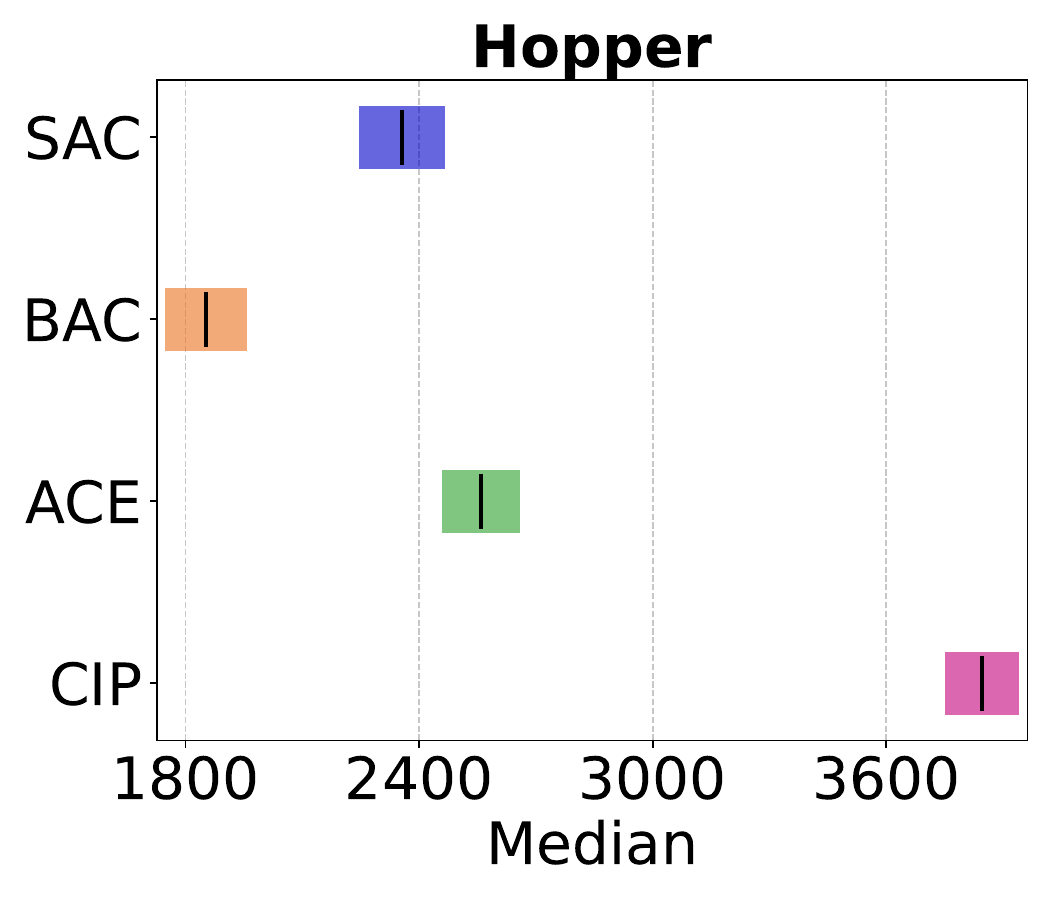}
    \includegraphics[width=0.32\textwidth]{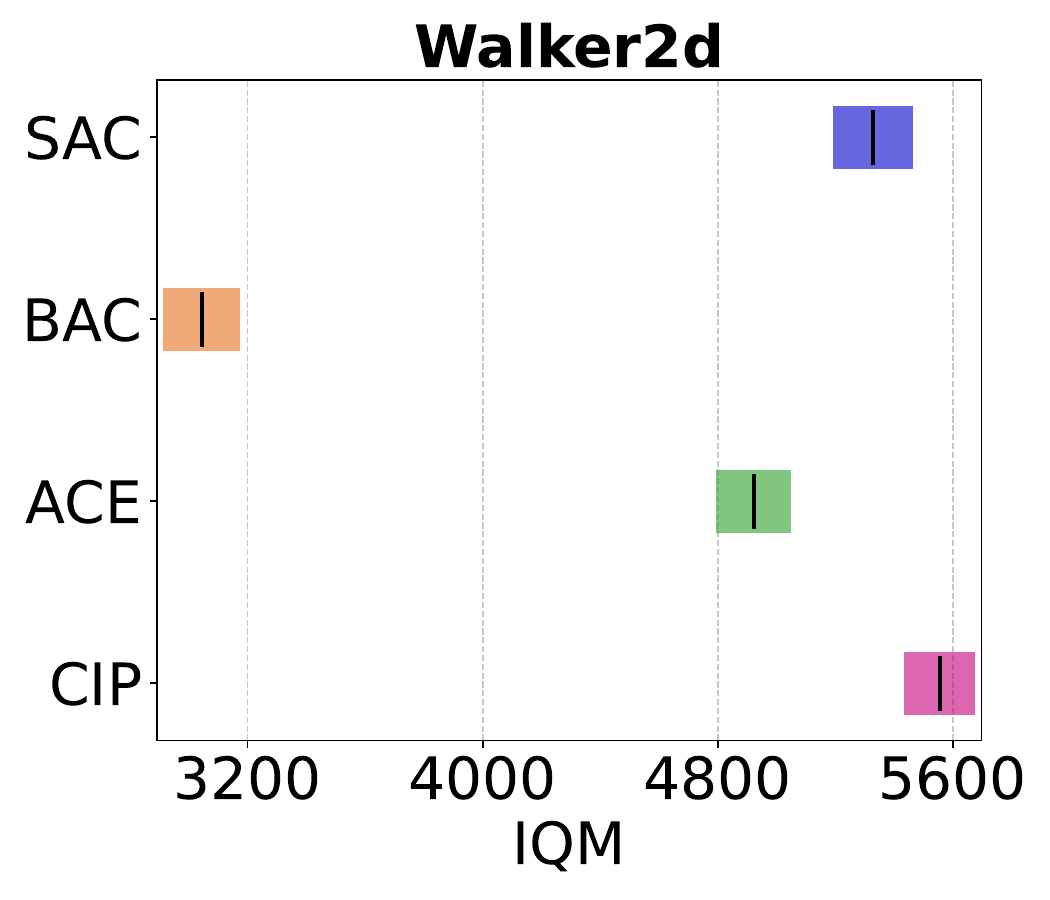}
    \includegraphics[width=0.32\textwidth]{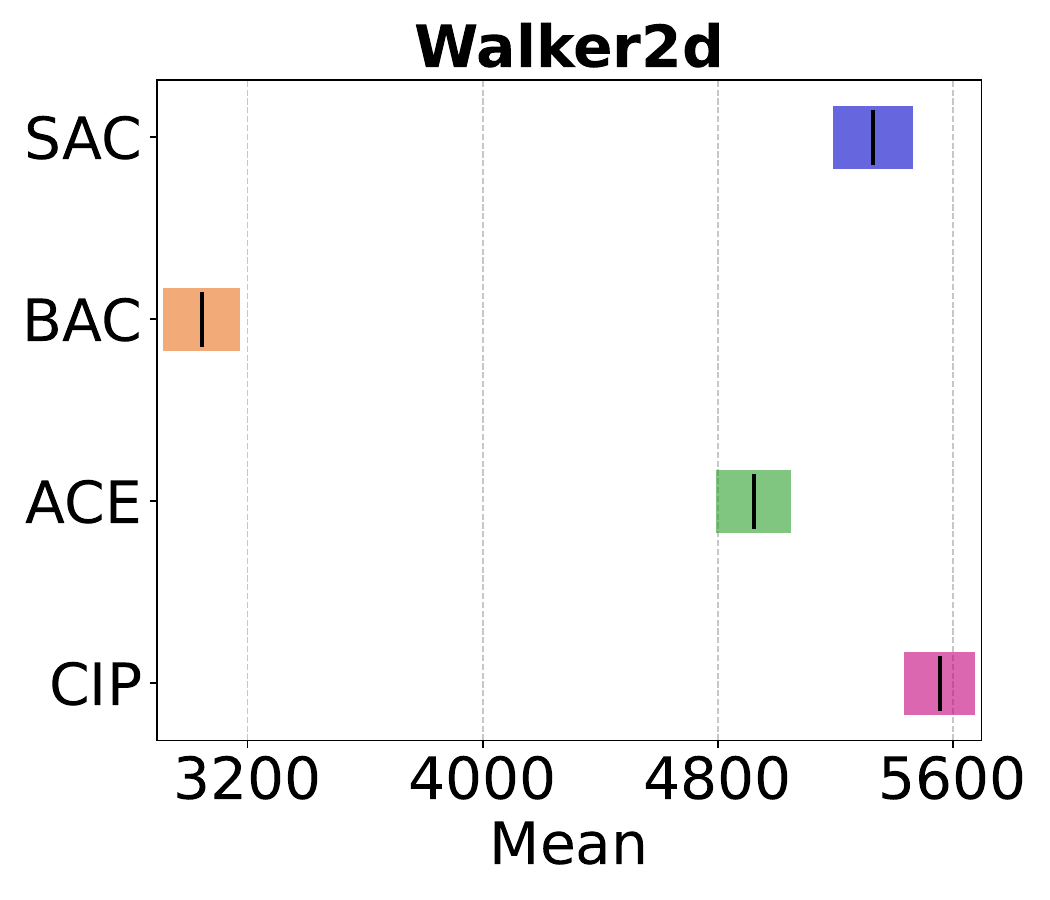}
    \includegraphics[width=0.32\textwidth]{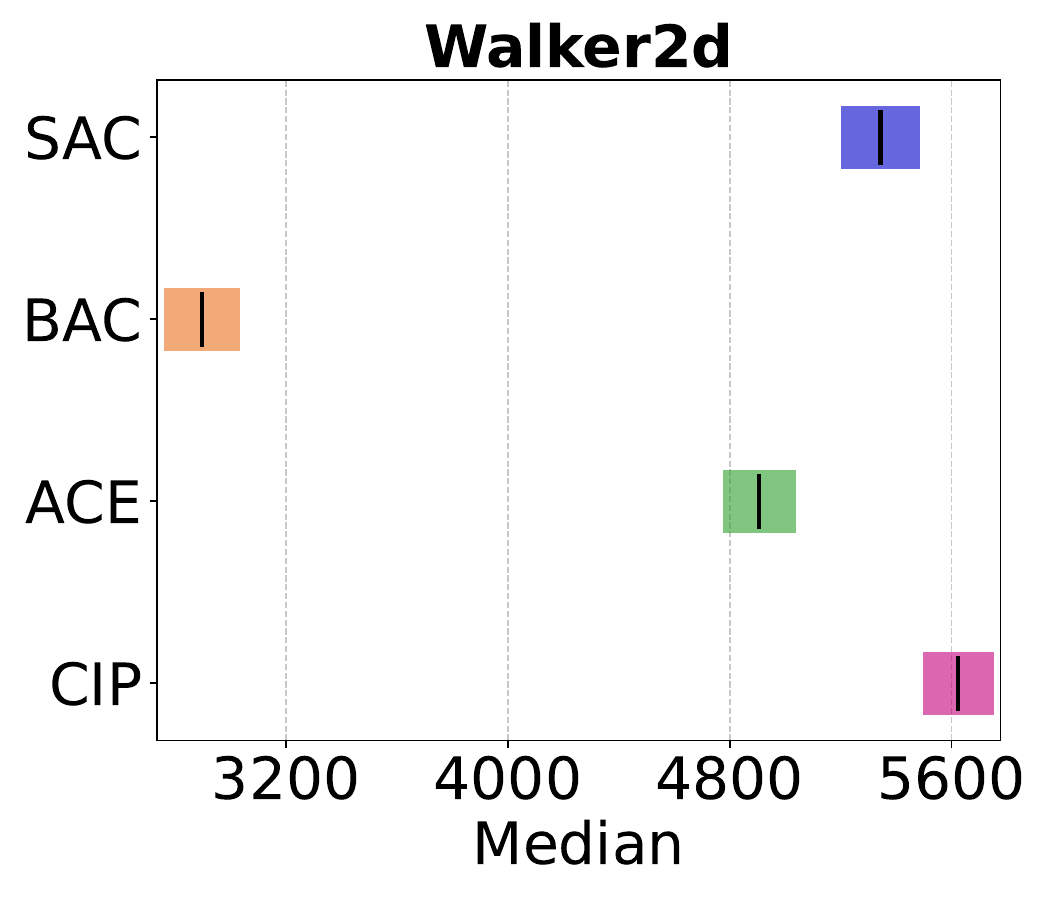}
    \caption{statistical metrics of IQM, Mean, and Median (higher values are better) on 4 MuJoCo tasks.}
    \label{fig:appendix_IQM_2}
\end{figure}

\textcolor{black}{
\subsubsection{Statistical performance analysis}
\label{sec:appendix_statis}
To further validate the statistical significance of the performance, we select $3$ statistical metrics~\citep{agarwal2021deep} - IQM, Mean, and Median - for analysis across $8$ locomotion tasks. The results are shown in Figure~\ref{fig:appendix_IQM_1} and~\ref{fig:appendix_IQM_2}. Our findings indicate that \texttt{\textbf{CIP}} achieves notably superior performance across all tasks, with the sole exception of the ant task where it performs slightly below BAC.
}

\textcolor{black}{
\vspace{-5mm}
\subsubsection{Generalization analysis}
We conduct multi-task experiments in the Meta-World environment~\citep{yu2020meta} to validate the generalizability. We establish MT1 and MT10 tasks for generalization validation:
}

\textcolor{black}{
\textbf{Multi-Task 1 (MT1)}: Learning one multi-task policy that generalizes to $5$ tasks belonging to the same environment. MT1 uses single Meta-World environments, with the training “tasks” corresponding to $5$ random initial object and goal positions. The goal positions are provided in the observation and are a fixed set, as to focus on the ability of algorithms in acquiring a distinct skill across multiple goals, rather than generalization and robustness.
}

\textcolor{black}{
\textbf{Multi-Task 10 (MT10)}: This task involves learning a single multi-task policy that generalizes to 50 tasks across 10 training environments, totaling 500 training tasks. A crucial step towards rapid adaptation to distinctly new tasks is the ability to train a single policy capable of solving multiple distinct training tasks. The multi-task evaluation in Meta-World tests the ability to learn multiple tasks simultaneously, without accounting for generalization to new tasks. The MT10 evaluation encompasses 10 environments: reach, push, pick and place, open door, open drawer, close drawer, press button top-down, insert peg side, open window, and close window.
}

\begin{figure}[t]
    \centering
    \includegraphics[width=0.24\textwidth]{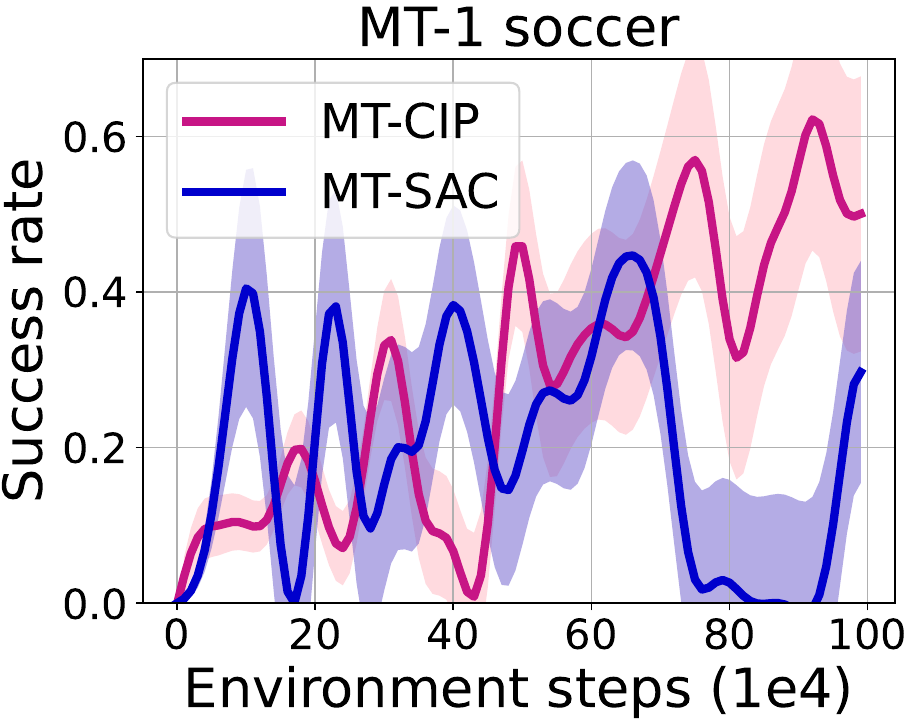}
    \includegraphics[width=0.24\textwidth]{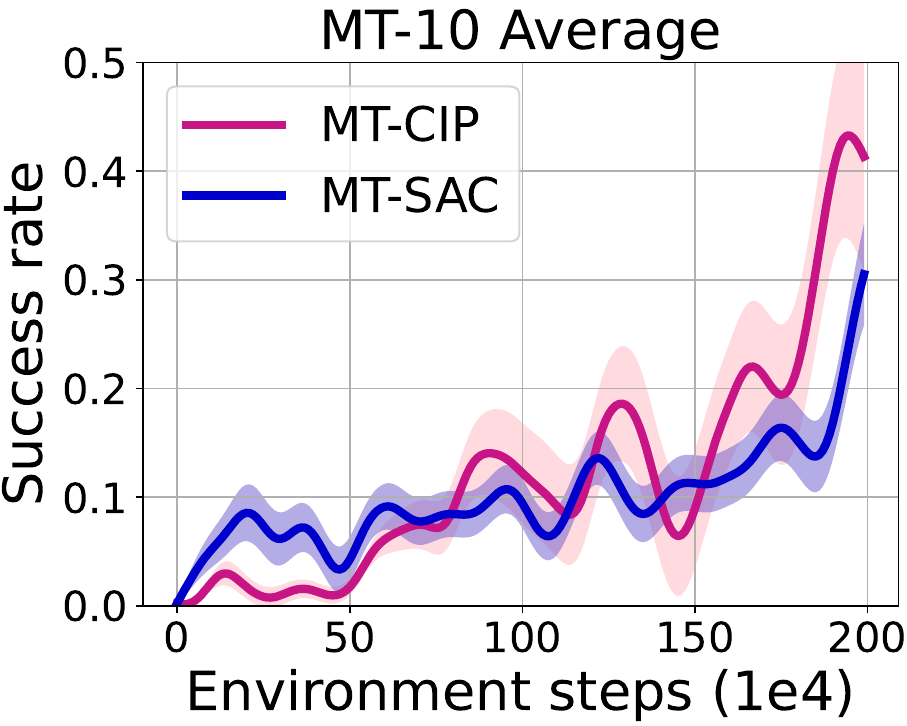}
    \includegraphics[width=0.24\textwidth]{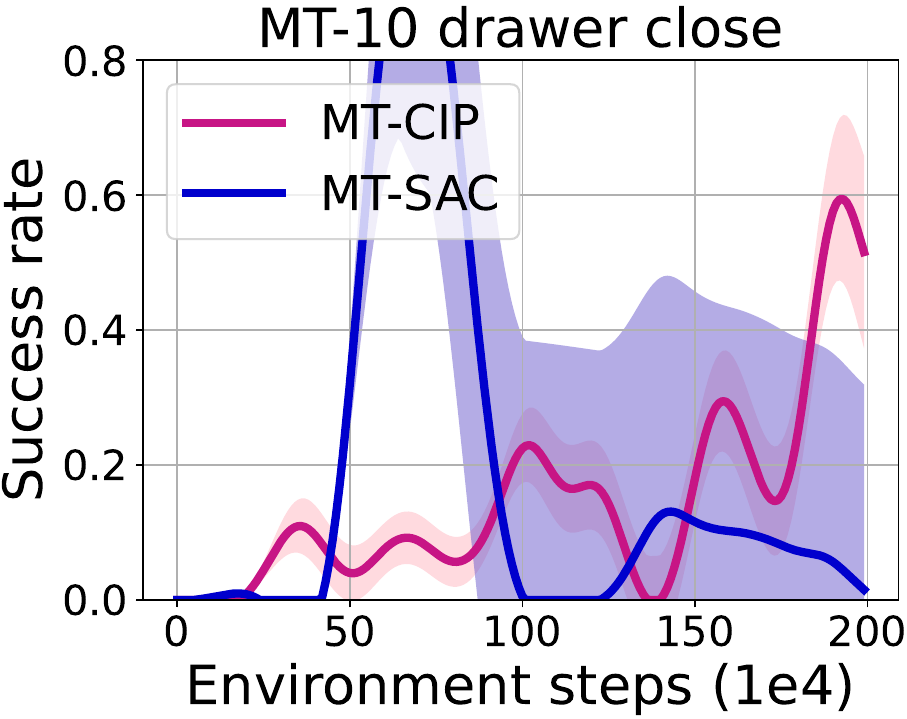}
    \includegraphics[width=0.24\textwidth]{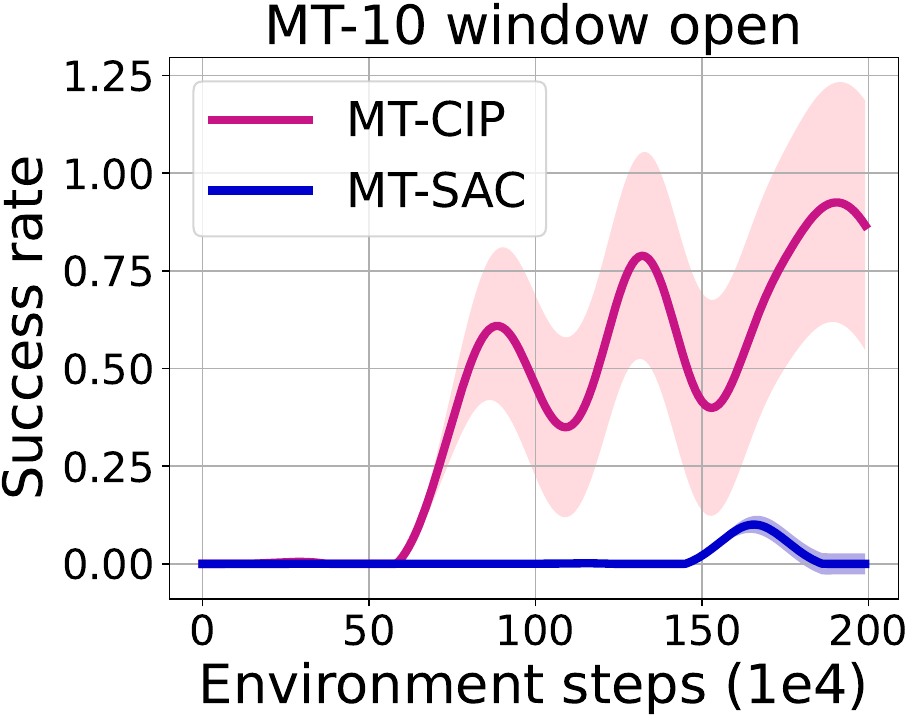}
    \caption{Generalization results in MT1 and MT10 tasks.}
    \label{fig:gen}
\end{figure}

\textcolor{black}{We adapt our proposed $\texttt{\textbf{CIP}}$ to multi-task learning  by incorporating a one-hot task ID as input, comparing $\texttt{\textbf{MT-CIP}}$ with MT-SAC. The results in Figure~\ref{fig:gen} show that $\texttt{\textbf{MT-CIP}}$ outperforms MT-SAC in both MT1 (soccer) and MT10 tasks, achieving average success rates above $50\%$ and $40\%$ respectively. Notably, $\texttt{\textbf{MT-CIP}}$ exhibits strong performance in specific MT10 tasks like drawer close and window open. 
The superior performance of $\texttt{\textbf{MT-CIP}}$ stems from its effective learning of causal information, enabling robust task transfer across diverse domains. While these results are promising, future work will focus on causal state abstraction for enhanced generalization and sample efficiency. All experiments were conducted under the same hyperparameter settings, and the implementation will be made publicly available.}

\begin{figure}[t]
    \centering
    \includegraphics[width=1\textwidth]{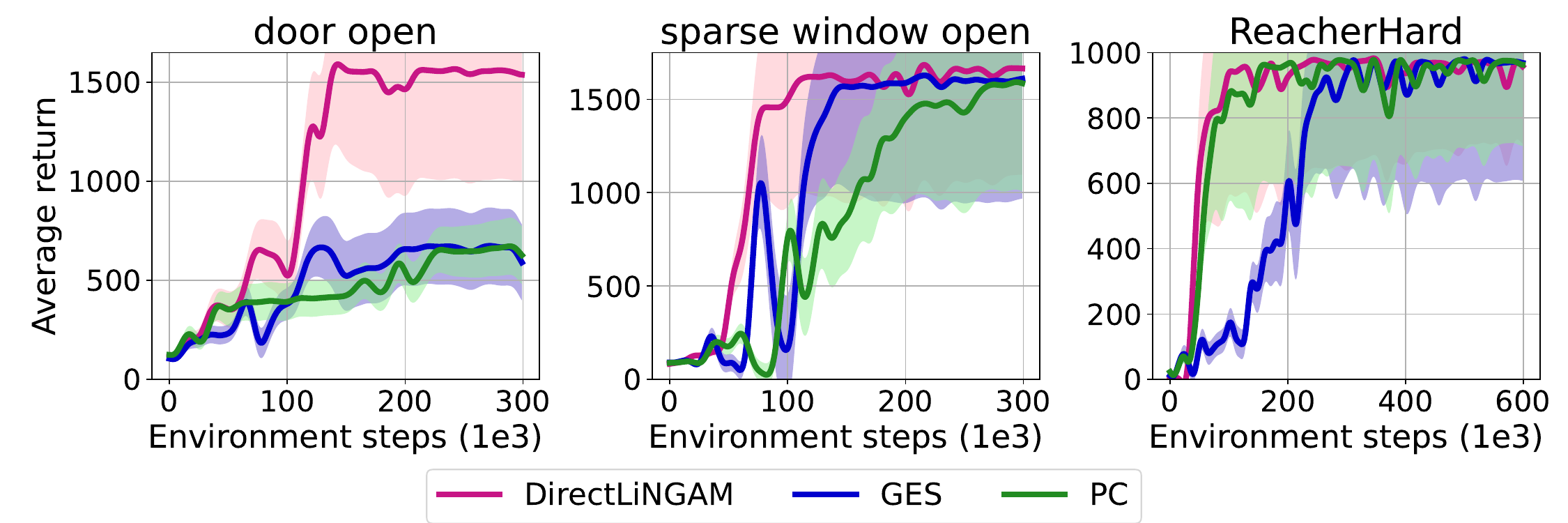}
    \caption{Compared performance with $2$ different causal discovery methods across $3$ task.}
    \label{fig:appendix_causal}
\end{figure}

\textcolor{black}{
\vspace{-5mm}
\subsubsection{Causal discovery analysis}
In $\texttt{\textbf{CIP}}$, we use the linear causal discovery method DirectLiNAM for causal structure learning. To explore alternative approaches, we compare it with two other causal discovery methods: score-based GES~\citep{chickering2002optimal} and constraint-based PC~\citep{spirtes2001causation}. The experimental results in Figure~\ref{fig:appendix_causal} across three tasks demonstrate that our chosen DirectLiNAM method exhibits superior performance compared to both alternatives. 
During experimentation, we also observe that both GES and PC methods incur significant computational overhead and frequently encounter memory constraints. In contrast, our proposed method $\texttt{\textbf{CIP}}$, which is fundamentally reward-guided, efficiently discovers causal relationships between dimensional factors in states and actions with respect to rewards. This approach better aligns with the requirements of policy learning while maintaining minimal computational costs.
}

\clearpage

\section{Details on the Proposed Framework}
\label{Details on the Proposed Framework}

Algorithm~\ref{alg:algorithm1} lists the full pipeline of \texttt{\textbf{CIP}} below. 

\begin{algorithm}[h]
\footnotesize
    \caption{Causal information prioritization for efficient RL}
    \label{alg:algorithm1}
    \textbf{Input}: $Q$ network $Q_{\pi_c}$, policy network $\pi_{c}$, inverse dynamics model $\phi_c$ with $Q$ network $Q_{\phi_c}$, replay buffer $\mathcal{D}$, local causal buffer $\mathcal{D}_c$, causal update interval $I$, causal matrix $M^{a \to s}$ and $M^{a \to r}$. 
    
    \begin{algorithmic}[]
    \FOR{each environment step $t$}
    \STATE Collect data with $\pi_{\theta}$ from real environment
    \STATE Add to replay buffer $\mathcal{D}$ and local buffer $\mathcal{D}_c$ 

    \ENDFOR
    \begin{tcolorbox}[colback=black!0!white,colframe=white!50!black,title=Step 1: Counterfactual data augmentation]
     \IF{every $I$ environment step}
     \STATE Sample transitions $\mathcal{D}_s$ from local buffer $\mathcal{D}_c$
     \STATE Learn causal mask matrix $M^{a \to r}$ with $\left\{(s, a, r, s')\right\}^{|\mathcal{D}_s|}$ for causal state prioritization
     \STATE Compute uncontrollable set $\mathcal{U}_s$ followed by Eq.~\ref{cda}
     \STATE Sample $(s, a, r, s') \in \mathcal{D}_s$
     \FOR{$s^i \in \mathcal{U}_s$}
     \STATE Sample $(\hat{s}, \hat{a}, \hat{r}, \hat{s}') \sim \mathcal{D}_s$
     \IF{state $\hat{s}^i$ $\in$ $\mathcal{U}_{\hat{s}}$ }
     \STATE Construct a counterfactual transition $(\tilde{s}, \tilde{a}, \tilde{r}, \tilde{s}')$ by swapping $(s^i, s'^{i})$ with $(\hat{s}^i, \hat{s}'^{i})$
     \STATE Add $(\tilde{s}, \tilde{a}, \tilde{r}, \tilde{s}')$ to local buffer $\mathcal{D}_c$
     \ENDIF
     \ENDFOR
     \ENDIF
     \end{tcolorbox} 

      \begin{tcolorbox}[colback=orange!0!white,colframe=orange!60!black,title=Step 2: Causal weighted matrix learning]
       \IF{every $I$ environment step}
     \STATE Sample transitions $\mathcal{D}_a$ from local buffer $\mathcal{D}_c$  
     \STATE Learn causal weighted matrix $M^{a \to r}$ with $\left\{(s, a, r, s')\right\}^{|\mathcal{D}_a|}$ for causal action prioritization
     \ENDIF
    \end{tcolorbox}

      \begin{tcolorbox}[colback=green!0!white,colframe=green!50!black,title=Step 3: Policy optimization with causal action empowerment]
     \FOR{each gradient step}
     \STATE Sample $N$ transitions $(s,a,r,s')$ from $\mathcal{D}$
     \STATE Compute causal action empowerment followed by Eq.~\ref{eq:emp_comp}.
     \STATE Calculate the target $Q_{\phi_c}$ value
     \STATE Update $Q_{\phi_c}$ by $\min_{\phi_c}(\mathcal{T}Q_{\phi_c}-Q_{\phi_c})^2$
     \STATE Update $\phi_{c}$ by $\max(Q_{\phi_c}(s,a)$)
     \STATE Calculate the target $Q_{\pi_c}$ value
     \STATE Update $Q_{\pi_c}$ by $\min_{\pi_c}(\mathcal{T}_cQ_{\pi_c}-Q_{\pi_c})^2$
     \STATE Update $\pi_{c}$ by $\max_{c}(Q_{\pi_c}(s,a) + \mathcal{E}_{\pi_c}(s)$)
     \ENDFOR
     \end{tcolorbox}
    \end{algorithmic} 
\end{algorithm}

\section{Experimental Platforms and Licenses}
\label{exp}
\subsection{Experimental platforms}
All experiments of this approach are implemented on 2 Intel(R) Xeon(R) Gold 6430 and 2 NVIDIA Tesla A800 GPUs.

\subsection{Licenses}
In our code, we have utilized the following libraries, each covered by its respective license agreements:
\begin{itemize}
    \item PyTorch (BSD 3-Clause "New" or "Revised" License)
    \item Numpy (BSD 3-Clause "New" or "Revised" License)
    \item Tensorflow (Apache License 2.0)
    \item Meta-World (MIT License)
    \item MuJoCo (Apache License 2.0)
    \item Deep Mind Control (Apache License 2.0)
    \item Adroit Hand (Creative Commons License 3.0)
\end{itemize}

\end{document}